\documentclass{article}
\usepackage{amsfonts}
\usepackage{amsmath}
\usepackage[group-separator={,}]{siunitx}
\usepackage{xfrac}
\usepackage{bm}
\usepackage{mathtools}
\usepackage{mathabx}
\usepackage{nicefrac}
\usepackage{aligned-overset}    
\usepackage{iclr2025_conference,times}
\usepackage[utf8]{inputenc}
\usepackage[T1]{fontenc}
\usepackage{microtype}      
\usepackage{xspace}
\usepackage{placeins}
\usepackage{titletoc}
\usepackage{xcolor}         
\usepackage{graphicx}
\usepackage{subfigure}
\usepackage[labelfont=bf]{caption}
\usepackage{wrapfig}
\usepackage{enumitem}
\usepackage{booktabs}
\usepackage{multirow}
\usepackage{soul}
\usepackage{changepage,threeparttable}
\usepackage{tabularx}
\usepackage{multirow}
\usepackage{etoolbox}
\newrobustcmd{\B}{\bfseries}

\definecolor{mydarkblue}{rgb}{0,0.08,0.45}
\usepackage[
    colorlinks=true,
    linkcolor=mydarkblue,
    citecolor=mydarkblue,
    filecolor=mydarkblue,
    urlcolor=mydarkblue,
]{hyperref}
\usepackage[nameinlink,capitalize,noabbrev]{cleveref}
\usepackage{url}
\usepackage[numbered]{bookmark}

\usepackage[textsize=tiny]{todonotes}
\setuptodonotes{bordercolor=white,color=TUgreen!50}

\DeclareRobustCommand{\colordot}[1]{\begin{tikzpicture}[baseline=(a.south)]
        \node[circle, scale=0.6,color=white, fill=#1] (a) {};
    \end{tikzpicture}}

\DeclareRobustCommand{\colorline}[1]{\begin{tikzpicture}
        \raisebox{1.5pt}{
            \draw[#1,solid,line width=1.5pt] (0,0) -- (1em,0);
        }
    \end{tikzpicture}}

\DeclareRobustCommand{\colorrectangle}[1]{%
  \begin{tikzpicture}[baseline=(a.west)]
      \draw[#1, very thick] (0, 0.02) rectangle (0.3, 0.17);
  \end{tikzpicture}%
}

\newcommand{\colorcross}[1]{\textcolor{#1}{\footnotesize\ding{58}}}

\newcommand{\colordiamond}[1]{\textcolor{#1}{\footnotesize\ding{117}}}

\usepackage{pifont}

\DeclareRobustCommand{\colorgradientbox}[2][white]{\begin{tikzpicture}[baseline=-0.5ex]\node [rectangle, left color=#1, right color=#2, anchor=base, minimum width=1.75em, minimum height=1em, draw=black] (box) at (0,0){};\end{tikzpicture}}

\definecolor{TUred}{RGB}{165,30,55}
\definecolor{TUdark}{RGB}{50,65,75}
\definecolor{TUgold}{RGB}{180,160,105}

\definecolor{TUgray}{RGB}{185,184,188}

\definecolor{TUdarkblue}{RGB}{65,90,140}
\definecolor{TUblue}{RGB}{0,105,170}
\definecolor{TUlightblue}{RGB}{80,170,200}

\definecolor{TUlightgreen}{RGB}{125,165,75}
\definecolor{TUgreen}{RGB}{125,165,75}
\definecolor{TUdarkgreen}{RGB}{50,110,30}

\definecolor{TUlightred}{RGB}{200,80,60}
\definecolor{TUpurple}{RGB}{175,110,150}
\definecolor{TUorange}{RGB}{210,150,0}

\definecolor{colorbiased}{HTML}{dd8452}
\definecolor{colorunbiased}{HTML}{4c72b0}
\definecolor{colorexact}{HTML}{55a868}

\definecolor{colorbiaseddark}{HTML}{6f4229}
\definecolor{colorunbiaseddark}{HTML}{263958}
\definecolor{colorexactdark}{HTML}{2b5434}

\newcommand*{\eg}{e.g.\@\xspace}
\newcommand*{\ie}{i.e.\@\xspace}

\usepackage[super]{nth}

\newcommand{\dataset}{dataset\xspace}
\newcommand{\datasets}{datasets\xspace}

\newcommand{\adam}{\textsc{\mbox{Adam}}\xspace}

\newcommand{\sgd}{\textsc{\mbox{SGD}}\xspace}

\newcommand{\nadamw}{\textsc{\mbox{NadamW}}\xspace}

\newcommand{\kfac}{\textsc{\mbox{K-FAC}}\xspace}
\newcommand{\ekfac}{\textsc{\mbox{EKFAC}}\xspace}
\newcommand{\cg}{\textsc{\mbox{CG}}\xspace}
\newcommand{\lbfgs}{\textsc{\mbox{L-BFGS}}\xspace}

\newcommand{\svd}{\textsc{SVD}\xspace}

\newcommand{\imagenetthou}{\textsc{ImageNet-1k}\xspace}
\newcommand{\imagenet}{\textsc{ImageNet}\xspace}

\newcommand{\cifarten}{\textsc{CIFAR-10}\xspace}
\newcommand{\cifartenc}{\textsc{CIFAR-10-C}\xspace}
\newcommand{\cifarhun}{\textsc{CIFAR-100}\xspace}
\newcommand{\cifarhunc}{\textsc{CIFAR-100-C}\xspace}

\newcommand{\resnetfifty}{\textsc{ResNet-50}\xspace}

\newcommand{\wideresnet}{\textsc{Wide ResNet}\xspace}

\newcommand{\vitfull}{\textsc{Vision Transformer}\xspace}
\newcommand{\vitlittle}{\textsc{ViT Little}\xspace}

\newcommand{\wrn}{\textsc{WideResNet 40-4}\xspace}
\newcommand{\allcnnc}{\textsc{All-CNN-C}\xspace}
\newcommand{\convnet}{\textsc{ConvNet}\xspace}

\newcommand{\batchnorm}{batch normalization\xspace}

\newcommand{\deepobs}{\textsc{DeepOBS}\xspace}
\newcommand{\backpack}{\textsc{BackPACK}\xspace}

\newcommand{\pytorch}{\textsc{PyTorch}\xspace}

\newcommand{\figleft}{{\em (Left)}\xspace}

\newcommand{\figright}{{\em (Right)}\xspace}

\newcommand{\figtop}{{\em (Top)}\xspace}

\newcommand{\figbottom}{{\em (Bottom)}\xspace}
 
\DeclareMathOperator*{\argmax}{arg\,max}
\DeclareMathOperator*{\argmin}{arg\,min}

\DeclareMathOperator{\Diag}{Diag}
\DeclareMathOperator{\diag}{diag}

 \DeclareMathOperator{\vect}{vec}

\newcommand{\Hessian}{\mH}
\newcommand{\GGN}{\mG}
\newcommand{\KFAC}{\mK}

\newcommand{\defeq}{\coloneqq}
\newcommand{\eqdef}{\eqqcolon}
\newcommand{\eqc}{\overset{\text{c}}{=}}

\newcommand{\R}{\mathbb{R}}
\newcommand{\N}{\mathbb{N}}

\newcommand*{\blockdiag}{\operatorname{blockdiag}}

\newcommand{\normltwo}{\ell^2}

\DeclarePairedDelimiterXPP\bigO[1]{\mathcal{O}}{(}{)}{}{#1}
\DeclarePairedDelimiterXPP\smallo[1]{o}{(}{)}{}{#1}
\DeclarePairedDelimiterXPP\bigOmega[1]{\Omega}{(}{)}{}{#1}
\DeclarePairedDelimiterXPP\smallomega[1]{\omega}{(}{)}{}{#1}
\DeclarePairedDelimiterXPP\bigTheta[1]{\Theta}{(}{)}{}{#1}

\newcommand*{\gaussian}[2]{
    {\ensuremath{\operatorname{\mathcal{N}}\mathopen{}\left(#1, #2\right)}}}
\newcommand*{\gaussianpdf}[3]{{\ensuremath{\operatorname{\mathcal{N}}\mathopen{}\left(#1; #2, #3\right)}}}
\newcommand*{\categoricalpdf}[2]{{\ensuremath{\operatorname{Cat}\mathopen{}\left(#1; #2\right)}}}

\newcommand*{\emptyarg}{\,\cdot\,}

\newcommand*{\numclasses}{C}
\newcommand*{\datadim}{D}
\newcommand*{\numparams}{P}
\newcommand*{\inputspace}{\R^\datadim}
\newcommand*{\outputspace}{\R^\numclasses}
\newcommand*{\params}{\vtheta}
\newcommand*{\paramsmap}{\params_\star}
\newcommand*{\Loss}{\mathcal{L}}
\newcommand*{\Lossreg}{\mathcal{L}_{\text{reg}}}
\newcommand*{\loss}{\ell}
\newcommand*{\network}{f}
\newcommand*{\regularizer}{r}
\newcommand*{\softmax}{\text{softmax}}

\newcommand*{\traindata}{\mathcal{D}}
\newcommand*{\testdata}{\mathcal{D}_\text{test}}
\newcommand*{\trainset}{\sD}
\newcommand*{\numtraindata}{N}
\newcommand*{\minibatch}{\mathcal{B}}
\newcommand*{\otherminibatch}{\tilde{\mathcal{B}}}
\newcommand*{\numminibatches}{M}
\newcommand*{\testsymbol}{\diamond}
\newcommand*{\numtestdata}{N_{\testsymbol}}

\newcommand*{\quadratic}{q}
\newcommand*{\grad}{\vg}
\newcommand*{\hessian}{\mH}
\newcommand*{\const}{c}

\newcommand*{\dslope}[1]{\partial_{#1}}
\newcommand*{\dcurv}[1]{\partial^2_{#1}}

\newcommand*{\eigval}{\lambda}
\newcommand*{\eigvals}{\mLambda}
\newcommand*{\eigvec}{\vu} 
\newcommand*{\eigvecs}{\mU}
\newcommand*{\rankbound}{K}      

\def\vzero{{\bm{0}}}

\def\vtheta{{\bm{\theta}}}

\def\vb{{\bm{b}}}

\def\vd{{\bm{d}}}
\def\ve{{\bm{e}}}

\def\vg{{\bm{g}}}

\def\vp{{\bm{p}}}

\def\vr{{\bm{r}}}
\def\vs{{\bm{s}}}
\def\vt{{\bm{t}}}
\def\vu{{\bm{u}}}
\def\vv{{\bm{v}}}
\def\vw{{\bm{w}}}
\def\vx{{\bm{x}}}
\def\vy{{\bm{y}}}
\def\vz{{\bm{z}}}

\def\mA{{\bm{A}}}
\def\mB{{\bm{B}}}
\def\mC{{\bm{C}}}
\def\mD{{\bm{D}}}

\def\mF{{\bm{F}}}
\def\mG{{\bm{G}}}
\def\mH{{\bm{H}}}
\def\mI{{\bm{I}}}

\def\mK{{\bm{K}}}

\def\mP{{\bm{P}}}

\def\mS{{\bm{S}}}

\def\mU{{\bm{U}}}
\def\mV{{\bm{V}}}
\def\mW{{\bm{W}}}
\def\mX{{\bm{X}}}

\def\mLambda{{\bm{\Lambda}}}
\def\mSigma{{\bm{\Sigma}}}

\def\mOmega{{\bm{\Omega}}}

\DeclareMathAlphabet{\mathsfit}{\encodingdefault}{\sfdefault}{m}{sl}
\SetMathAlphabet{\mathsfit}{bold}{\encodingdefault}{\sfdefault}{bx}{n}

\def\sD{{\mathbb{D}}}

\usepackage{amsthm}
\usepackage{thmtools}
\usepackage{thm-restate}

\declaretheoremstyle[
headfont=\normalfont\bfseries,
notefont=\normalfont,
bodyfont=\normalfont,
headpunct={},
postheadspace=\newline,
spaceabove=1.5\parskip,  ]{definitionstyle}

\declaretheoremstyle[
headfont=\normalfont\bfseries,
notefont=\normalfont,
bodyfont=\normalfont\itshape,
headpunct={},
postheadspace=\newline,
spaceabove=1.5\parskip,  ]{lemmastyle}

\declaretheoremstyle[
headfont=\normalfont\bfseries,
notefont=\normalfont,
bodyfont=\normalfont\itshape,
headpunct={},
postheadspace=\newline,
spaceabove=1.5\parskip,  ]{theoremstyle}

\declaretheoremstyle[
headfont=\normalfont\bfseries,
notefont=\normalfont,
bodyfont=\normalfont,
headpunct={},
spaceabove=1.5\parskip,  ]{remarkstyle}

\usepackage{scrhack}
\usepackage{algorithm}
\usepackage[noend]{algpseudocode}

\algrenewcommand\algorithmicindent{2em}
\algrenewcommand\alglinenumber[1]{{\textcolor{darkgray}{#1}}}
\algrenewcommand{\algorithmiccomment}[1]{\hfill{\small \textcolor{darkgray}{#1}}}

\makeatletter
\expandafter\patchcmd\csname\string\algorithmic\endcsname{\itemsep\z@}{\itemsep=0.5ex}{}{}
\newcommand\fs@booktabsruled{\def\@fs@cfont{\bfseries\strut}\let\@fs@capt\floatc@ruled
    \def\@fs@pre{\hrule height\heavyrulewidth depth0pt \kern\belowrulesep}\def\@fs@mid{\kern\aboverulesep\hrule height\lightrulewidth\kern\belowrulesep}\def\@fs@post{\kern\aboverulesep\hrule height\heavyrulewidth\relax}\let\@fs@iftopcapt\iftrue
}
\makeatother
\floatstyle{booktabsruled}
\restylefloat{algorithm}
\allowdisplaybreaks

\title{Debiasing Mini-Batch Quadratics\\for Applications in Deep Learning}

\author{Lukas Tatzel, Bálint Mucsányi, Osane Hackel \& Philipp Hennig \\
Tübingen AI Center\\
University of Tübingen\\
Tübingen, Germany\\
\texttt{\{lukas.tatzel,balint.mucsanyi\}@uni-tuebingen.de} 
}

\iclrfinalcopy

\begin{document}

\maketitle

\begin{abstract}
Quadratic approximations form a fundamental building block of machine learning
  methods. E.g., second-order optimizers try to find the Newton step into the minimum of a
  local quadratic proxy to the objective function; and the second-order approximation of
  a network's loss function can be used to quantify the uncertainty of its outputs via
  the Laplace approximation.
When computations on the entire training set are intractable---typical for deep
  learning---the relevant quantities are computed on mini-batches. This, however,
  distorts and biases the shape of the associated \textit{stochastic} quadratic
  approximations in an intricate way with detrimental effects on applications.
In this paper, we (i) show that this bias introduces a systematic error, (ii) provide
  a theoretical explanation for it, (iii) explain its relevance for second-order
  optimization and uncertainty quantification via the Laplace approximation in deep
  learning, and (iv) develop and evaluate debiasing strategies.
\end{abstract}

\section{Introduction}
\label{sec:introduction}

Quadratic approximations of the loss landscape are increasingly used by algorithms in
deep learning, from pruning methods \citep{Dong2017LearningPrune,Zeng2019MLPrune} and
influence functions \citep{Koh2017Understanding} to second-order optimizers
\citep{Amari1998NaturalGW,martens2010deep,martens2015optimizing,grosse2016kroneckerfactored,botev2017practical,zhang2017blockdiagonal,George2018Fast,martens2018kroneckerfactored,osawa2019large}
and uncertainty quantification via the Laplace approximation
\citep{Ritter2018ScalableLaplace,Ritter2018Online,kristiadi2020being,daxberger2021laplace,Immer2021Improving}.
When the computations are intractable on the entire training set---typical for deep
learning---the quantities of interest are computed on mini-batches subsampled from the
training data.
The goal of this work is to highlight that mini-batching systematically biases the
shape of a quadratic approximation.

\begin{figure}
  \centering
  Top two eigenvectors computed on mini-batch\\[0.25ex]
  \includegraphics[scale=0.99]{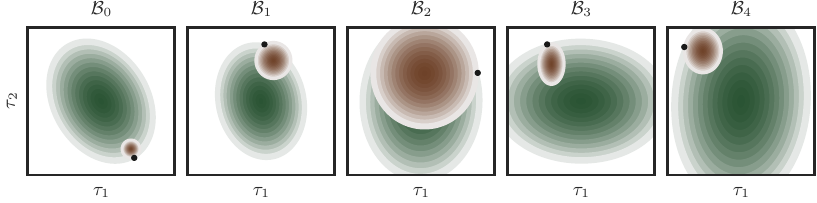}
  \caption{\textbf{A systematic bias?} We compute five mini-batch quadratics
    $\quadratic(\emptyarg; \minibatch_m)$ with batch size $\vert\minibatch_m\vert = 512$
    for the loss landscape of the fully trained \allcnnc model on \cifarhun data around
    $\params_0 \gets \params_\star$ (shown as \colordot{black}).
Each mini-batch quadratic defines a 2D subspace spanned by the top two eigenvectors
    $\eigvec_1, \eigvec_2$ of $\hessian_{\minibatch_m}$, in which we evaluate (i) the
    quadratic $\quadratic(\params_\star + \tau_1 \eigvec_1 + \tau_2 \eigvec_2;
      \minibatch_m)$ itself (shown in \colorgradientbox{colorbiaseddark}) and (ii) the
    full-batch quadratic $\quadratic(\params_\star + \tau_1 \eigvec_1 + \tau_2 \eigvec_2;
      \traindata)$ (shown in \colorgradientbox{colorexactdark}).
In that subspace, the mini-batch quadratic is much ``narrower'' than the full-batch
    version which leads to overly small Newton steps and overconfident uncertainty
    estimates via the Laplace approximation.}
  \label{fig:visual_abstract}
\end{figure}

\textbf{A systematic bias?} \Cref{fig:visual_abstract} illustrates the phenomenon. It
shows five \textit{mini-batch} quadratics in their top-curvature 2D subspace for the
fully trained \allcnnc model on \cifarhun data. For comparison, the full-batch
quadratic, where all quantities are evaluated on the \textit{entire} training set, is
projected into the \textit{same} 2D subspace. Within that subspace, the two quadratics
are quite different: The mini-batch quadratic is much ``narrower'' (exhibits larger
curvature) than the full-batch version. Given that the full-batch quadratic is the
``right'' object to serve as the basis for, e.g., a Newton step or a Laplace
approximation,\footnote{We are \textit{not} concerned with the approximation error
arising from the quadratic approximation of the non-quadratic function
$\Lossreg(\emptyarg; \traindata) \approx \quadratic(\emptyarg; \traindata)$ (see
\cref{eq:fullbatch_quadratic}), but only with the consequences of replacing the
full-batch quantities by their mini-batch counterparts $\quadratic(\emptyarg;
\traindata) \approx \quadratic(\emptyarg; \minibatch)$.} the mini-batch version is
\textit{not} a meaningful surrogate: Its Newton step is overly small and a Laplace
approximation yields an overconfident uncertainty estimate.

\textbf{Contributions.}
To enable stable and efficient stochastic second-order optimizers, as well as reliable
techniques for uncertainty quantification, we analyze this phenomenon and
develop strategies to mitigate it.
More specifically, our contributions are as follows: (i) We study mini-batch quadratics
empirically and show that their geometry is systematically biased; (ii) we provide an
explanation for this phenomenon, explaining the bias as an instance of the classic
regression to the mean (directions of extreme steepness/curvature for one particular
mini-batch are less extreme for other mini-batches), (iii) we explain the relevance of
this bias for second-order optimization and uncertainty quantification via the Laplace
approximation, and (iv) develop and evaluate debiasing strategies.

\section{Notation \& background}

\textbf{The regularized loss.}
We consider a general supervised learning problem, where we try to find the optimal parameters
$\params_\star = \argmin_{\params \in \R^\numparams} \mathcal{L}_{\text{reg}}(\params,
  \traindata)$ for the parameterized function $\network_{\params} : \inputspace
  \rightarrow \outputspace$ by minimizing the regularized loss $\Lossreg$ on $\numtraindata$ training examples $\trainset \defeq \{(\vx_n, \vy_n) \in
  \inputspace \times \outputspace\}_{n \in \traindata}$,
\begin{align}
  \Lossreg(\params; \traindata)
  \defeq
  \Loss(\params; \traindata) + \regularizer(\params)
  \quad \text{with} \quad
  \Loss(\params; \traindata)
  \defeq
  \frac{1}{\vert \traindata \vert}
  \sum_{n \in \traindata}
  \loss(\network_{\params}(\vx_n), \vy_n), \quad \traindata = \{1, \ldots,
  \numtraindata\}.
  \label{eq:Loss_reg}
  \raisetag{2ex}
\end{align}
The empirical risk $\mathcal{L}$ measures the dissimilarity between the model's
predictions $\network_{\params}(\vx_n)$ and the true outputs $\vy_n$ via a loss function
$\loss : \outputspace \times \outputspace \rightarrow \R$. The regularizer $\regularizer
: \R^P \to \R$, $\regularizer(\params) \defeq \nicefrac{\beta}{2} \,
\Vert\params\Vert_2^2$ with $\beta \in \R_{\geq 0}$, penalizes the ``complexity'' of the
model.

\subsection{Local full-batch \& mini-batch quadratic approximations}
\label{sec:background_quadratic_approximations}

\textbf{Full-batch quadratic.}
A local quadratic approximation of the regularized loss around $\params_0 \in
  \R^\numparams$ is given by the second-order Taylor expansion
\begin{equation}
  \Lossreg(\params; \traindata)
  \; \approx \;
  \quadratic(\params; \traindata)
  \defeq
  \frac{1}{2} (\params - \params_0)^\top \hessian_\traindata (\params - \params_0)
  + (\params - \params_0)^\top \grad_\traindata
  + \const_\traindata,
  \label{eq:fullbatch_quadratic}
\end{equation}
where
$
  \const_\traindata
  \defeq
  \Lossreg(\params_0; \traindata)
$,
$
  \grad_\traindata
  \defeq
  \nabla \Lossreg(\params_0; \traindata)
$
and
$
  \hessian_\traindata
  \defeq
  \nabla^2 \Lossreg(\params_0; \traindata)
  =
  \nabla^2 \Loss(\params_0; \traindata) + \beta \mI
$
is (some approximation of) the Hessian at $\params_0$ (all derivatives are with respect
to the parameters $\params$ unless stated otherwise). As all quantities are evaluated on
the \textit{entire} training set $\traindata$, we refer to this as the
\textit{full-batch} quadratic. It holds that
$
  \nabla \quadratic(\params; \traindata)
  =
  \hessian_\traindata (\params - \params_0) + \grad_\traindata
$
and
$
  \nabla^2 \quadratic(\params; \traindata)
  \equiv
  \hessian_\traindata.
$

\textbf{Mini-batch quadratic.}
When the computations are intractable on the entire training set, the quantities in
\cref{eq:fullbatch_quadratic} are typically computed on a \textit{mini-batch}---a
small randomly drawn subset---of the training data $\minibatch \subset \traindata,
\vert\minibatch\vert \ll \numtraindata$, resulting in a \textit{stochastic} quadratic
approximation $\quadratic(\emptyarg; \minibatch) \approx \quadratic(\emptyarg;
\traindata)$. As $\const_\minibatch$, $\grad_\minibatch$ and $\hessian_\minibatch$ are
unbiased estimates of $\const_\traindata$, $\grad_\traindata$ and $\hessian_\traindata$,
this substitution may seem innocent, but, as we will see in
\cref{sec:geometry_minibatch_quadratic}, it affects the geometry of the quadratic
approximation \textit{substantially}.

\textbf{Directional slope and curvature.} Consider a cut $r$ through the quadratic
$\quadratic(\emptyarg; \minibatch)$ from $\params_\bullet \in \R^\numparams$ along the
normalized direction $\vd$, $\Vert\vd\Vert = 1$. It holds that (derivation in
\cref{sec:details_directional_derivatives})
$
  r(\tau)
  \defeq
  \quadratic(\params_\bullet + \tau \vd; \minibatch)
  =
  \nicefrac{1}{2} \tau^2
  \vd^\top \nabla^2 \quadratic(\params_\bullet; \minibatch) \, \vd
  + \tau
  \vd^\top \nabla \quadratic(\params_\bullet; \minibatch)
  + \text{const}.
$
So, as a function of $\tau$, $r : \R \to \R$ is a 1D parabola with derivatives
$
  r'(\tau)
  = \tau \, \vd^\top \nabla^2 \quadratic(\params_\bullet; \minibatch) \, \vd
  + \vd^\top \nabla \quadratic(\params_\bullet; \minibatch)
$ and
$
  r''(\tau) \equiv \vd^\top \nabla^2 \quadratic(\params_\bullet; \minibatch) \, \vd.
$
We denote the \textit{directional} slope and curvature of the quadratic
$\quadratic(\emptyarg, \minibatch)$ at $\params_\bullet$ in direction $\vd$ by
\begin{equation*}
  \dslope{\vd} \, \quadratic(\params_\bullet; \minibatch)
  \defeq r'(0) = \vd^\top \nabla \quadratic(\params_\bullet; \minibatch)
\quad \text{and} \quad
\dcurv{\vd} \, \quadratic(\params_\bullet; \minibatch)
  \defeq r''(0) = \vd^\top \nabla^2 \quadratic(\params_\bullet; \minibatch) \, \vd.
\end{equation*}
The directional slope and curvature at $\params_\bullet$ are thus simply the projections
of the quadratic's gradient 
and Hessian
at that location onto the direction.

\textbf{Eigenvalues as directional curvatures.} The directional curvature of
the quadratic $\quadratic(\emptyarg; \minibatch)$ along one of $\hessian_\minibatch$'s
normalized eigenvectors $\eigvec$ coincides with the corresponding eigenvalue $\eigval$ since
$
  \dcurv{\eigvec}\, \quadratic(\params_\bullet; \minibatch)
  = \eigvec^\top \nabla^2 \quadratic(\params_\bullet; \minibatch) \, \eigvec
  = \eigvec^\top \hessian_\minibatch \, \eigvec
  = \eigval \Vert\eigvec\Vert^2
  = \eigval.
$
Thus, in the context of a quadratic, an
eigenvalue of the Hessian $\hessian_\minibatch$ has a geometric interpretation as the directional curvature
along the respective eigenvector.

\textbf{GGN \& FIM.} Next, we discuss second-order optimization methods and the Laplace
approximation (LA) for neural networks. Both techniques rely on local quadratic
approximations of the regularized loss function and require a \textit{positive definite}
curvature matrix $\hessian_\minibatch$. The empirical risk's Hessian $\nabla^2
\Loss(\params_0; \minibatch)$ can be \textit{indefinite} and is therefore typically
replaced by (an approximation of) the positive semi-definite Generalized Gauss-Newton
matrix (GGN) $\GGN_{\minibatch}$ or Fisher information matrix (FIM) $\mF_{\minibatch}$.
In fact, GGN and FIM are often identical \citep[Sec.~9.2]{martens2020new}. The resulting
$\hessian_\minibatch$ is positive definite if $\beta > 0$, or when a damping term
$\delta \mI$, $\delta \in \R_{> 0}$ is added (\eg,~in trust-region methods).

\subsection{Second-order methods \& conjugate gradients}
\label{sec:background_second_order_methods}

\textbf{The Newton step.} Due to the simple polynomial form of the quadratic
$\quadratic(\emptyarg; \minibatch)$, its minimum can be derived in closed form and is
given by the Newton step
$
  \argmin_{\params} \quadratic(\params; \minibatch)
  =
  \params_0 - \hessian_\minibatch^{-1} \grad_\minibatch.
$ This serves as the basis for second-order optimizers like \lbfgs
\citep{liu1989onthe,nocedal1980updating}, the Hessian-free approach
\citep{martens2010deep} or \kfac (Kronecker-factored approximate curvature)
\citep{martens2015optimizing,grosse2016kroneckerfactored,martens2018kroneckerfactored}.

\textbf{Conjugate gradients.} We focus on the method of conjugate gradients (\cg)
\citep{Hestenes1952}, as it is a powerful method specifically designed to minimize
quadratics with positive definite Hessians effectively (details in
\cref{sec:details_cg}). It is particularly useful in the context of large-scale
optimization because it only requires access to matrix-vector products with the
curvature matrix
$
  \vv
  \mapsto
  \hessian_\minibatch \, \vv
$
that can be computed in a matrix-free manner
\citep{pearlmutter1994fast,schraudolph2002fast}, \ie without ever materializing the full
matrix in memory---and it has been used successfully for training neural networks
\citep{martens2010deep}.
Starting at $\params_0$, \cg creates a sequence of iterates $(\params_0, \ldots,
  \params_\numparams)$. In each iteration $p$, two main steps are performed: (i) Given the
current position $\params_p$ and a normalized search direction $\vd_p$, the algorithm
finds the minimum of the quadratic along $\params_p + \tau \vd_p$, \ie
\begin{equation}
  \params_{p+1} = \params_p + \tau_p \vd_p
  \quad \text{with} \quad
  \tau_p
  \defeq
  \argmin_{\tau \in \R} \quadratic(\params_p + \tau \vd_p; \minibatch)
  =
  -\frac{
    \dslope{\vd_p} \quadratic(\params_p; \minibatch)
  }{
    \dcurv{\vd_p} \quadratic(\params_p; \minibatch)
  }.
  \label{eq:cg_update_with_magnitude}
\end{equation}
In the second step (ii), \cg constructs the next search direction $\vd_{p+1}$ such that
it is \textit{conjugate} to all previous search directions, \ie $\vd_{p+1}^\top
  \hessian_\minibatch \, \vd_i = 0$ for all $i \in \{1, \dots, p\}$.

\subsection{Laplace approximation for neural networks}
\label{sec:laplace_approximation}

\textbf{Laplace approximation.} The Laplace approximation (LA) turns a trained standard
neural network into a Bayesian neural network in a post-hoc manner
\citep{MacKay1991BayesianModel,Ritter2018ScalableLaplace,kristiadi2020being,daxberger2021laplace}
(details in \cref{sec:details_laplace_approximation}). The idea is to reinterpret the
regularized loss $\Lossreg$ as the negative unnormalized log posterior $\log p(\params
  \mid \trainset)$ of a specific Bayesian model.
This interpretation identifies the optimal parameters
$
  \paramsmap
  =
  \argmin_{\params} \Lossreg(\params; \traindata)
  =
  \argmax_{\params} p(\params \mid \trainset)
$
as the mode of the posterior, \ie as the maximum a posteriori (MAP) estimate. A
second-order approximation of the regularized loss around $\params_0 \gets \paramsmap$
then translates into a Gaussian approximation of the posterior---the so-called Laplace
approximation \citep{MacKay1992EvidenceFramework}, \ie
\begin{equation}
  \Lossreg(\params; \traindata)
  \approx
  \quadratic(\params; \traindata)
  =
  \frac{1}{2} (\params - \paramsmap)^\top \hessian_\traindata (\params - \paramsmap) +\text{const.}
  \quad \leadsto \quad
  p(\params \mid \trainset)
  \approx
  \gaussianpdf{\params}{\paramsmap}{\mSigma_\traindata},
  \label{eq:full_batch_quadratic_to_gaussian}
\end{equation}
with $\mSigma_\traindata \defeq \numtraindata^{-1} \hessian_\traindata^{-1}$ (or an
approximation thereof). 
We obtain the predictive uncertainty $p(\vy_\testsymbol \mid \vx_\testsymbol,
\trainset)$ for some test input $\vx_\testsymbol \in \inputspace$ by propagating the
parameter uncertainty $\gaussian{\paramsmap}{\mSigma_\traindata}$ to the model's outputs
via the linearized network \citep{Immer2021Improving,roy2024reparameterization}.

\textbf{Full-batch vs. mini-batch LA.} It is common practice to compute the LA on the
\textit{entire} training set. Depending on the curvature approximation, this comes at
considerable computational costs. For example, a full-batch LA is prohibitive even for
moderately sized models/\datasets when a low-rank approximation of the Hessian is
computed via repeated Hessian-vector products (each of which requires a full pass over
the training data); and it is essentially infeasible for model classes like LLMs, which
are trained on massive datasets. However, what adds most to the costs is that it is
standard to \textit{tune} the prior precision \citep{daxberger2021laplace}. This adds
another outer loop that requires the same procedure to be performed multiple times.
Being able to emulate the behavior of the full-batch quadratic on a mini-batch would
thus be useful. We therefore study the mini-batch setting, \ie we replace
$\quadratic(\emptyarg; \traindata)$ by $\quadratic(\emptyarg; \minibatch)$ in
\cref{eq:full_batch_quadratic_to_gaussian}. We will see in \cref{sec:exp_LA} that, when
mitigating the associated biases, a mini-batch LA can be a good proxy for the full-batch
LA.

\section{The shape of a mini-batch quadratic}
\label{sec:geometry_minibatch_quadratic}

This section studies the ``shape'' of a mini-batch quadratic $\quadratic(\emptyarg;
\!\minibatch)$ and how it differs from $\quadratic(\emptyarg;\!\traindata)$.

\subsection{Empirical study of the directional slopes and curvatures}
\label{sec:empirical_study_dd}

\textbf{The high-curvature subspace is relevant for all common use cases.} In our
empirical study, we focus on the \textit{top}-curvature subspace. This subspace is
particularly relevant for several reasons: 
\begin{enumerate}
  \item By the Eckart-Young-Mirsky Theorem, a truncated \svd is Frobenius norm-optimal,
        \ie a low-rank approximation using the top eigenvectors is ideal from a
        theoretical perspective. As the spectrum of the Hessian typically decays quickly
        \citep{ghorbani2019investigation}, the bulk of the curvature information is
        contained in the top-curvature subspace.

  \item In the context of optimization, it has been observed that the gradient mainly
        lives in the high-curvature space \citep{GurAri2018Gradient, dangel2022vivit}.
        Thus, it makes sense for a second-order method to operate mainly in that space,
        as steps outside of it can not be expected to reduce the objective function
        significantly.

  \item In the context of the LA, directions of large curvature correspond to directions
        in the weight space with \textit{low} variance, \ie these directions capture
        what we \textit{know} about the model's parameters. Consequently,
        \citet[p.~19]{daxberger2021laplace} describe a low-rank approximation of the
        Hessian based on its top eigenvectors.
\end{enumerate}

\textbf{Experimental procedure.} We use the same setting as in
\cref{fig:visual_abstract}: The fully trained \allcnnc model on the \cifarhun dataset.
To isolate the effect of data subsampling, we eliminate \textit{all other} sources of
noise. Thus, we remove the dropout layers from the model. We use the cross-entropy loss,
an $\normltwo$-regularizer, and train the model with \sgd for $350$ epochs, see
\cref{sec:exp_details_training} for details.

\begin{figure}
  \centering
  Eigenvectors computed on mini-batch\\[0.25ex]
  \includegraphics[width=0.99\textwidth]{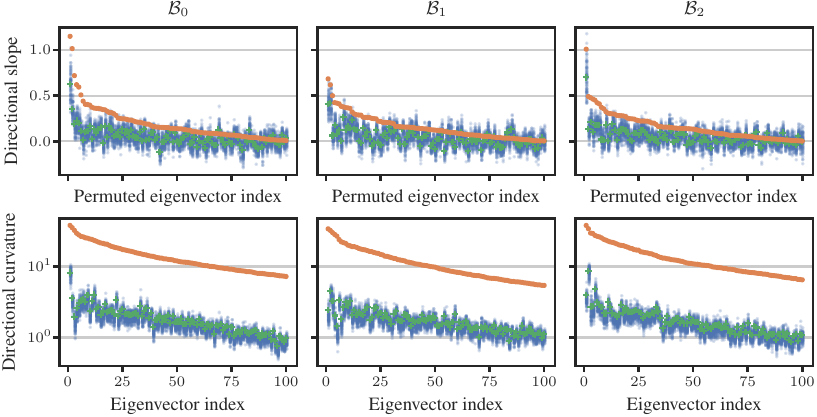}
  \caption{\textbf{Directional slopes and curvatures are biased.}
We use the \cifarhun dataset with the fully trained \allcnnc model and draw three
    mini-batches $\minibatch_m$ of size $\vert\minibatch_m\vert = 512$ to compute the
    top $100$ eigenvectors $\eigvec_1, \ldots, \eigvec_{100}$.
For each mini-batch/column, we show the directional slopes \figtop and curvatures
    \figbottom evaluated on (i) $\quadratic(\params_\star; \minibatch_{m})$ (\ie on the
    \textit{same} mini-batch of data) as \colordot{colorbiased}, (ii)
    $\quadratic(\params_\star; \minibatch_{m'})$ for $m' \neq m$ (\ie for all
    \textit{other} mini-batches) as \colordot{colorunbiased} and (iii) the full-batch
    quadratic (the average of the orange and all blue dots, see
    \cref{eq:average_directional_slopes_curvatures}) as \colorcross{colorexact}.
For the top panel, we switch the order and sign of the eigenvectors such that the
    orange dots are all above zero and in descending order.
There is a strong, systematic bias, particularly in the curvature: Computing the
    eigenvectors and directional curvatures on the same data results in over-estimation
    by roughly one order of magnitude. }
  \label{fig:bias}
\end{figure}

We then pick a mini-batch $\minibatch_m$ of size $512$ and compute the $100$
eigenvectors $\eigvec_1, \ldots, \eigvec_{100}$ to the $100$ \textit{largest}
eigenvalues of $\Hessian_{\minibatch_m} \gets \GGN_{\minibatch_m} + \beta \mI$. That is
the Hessian of the $\normltwo$-regularized mini-batch loss, where we replaced $\nabla^2
\Loss(\params_\star; \minibatch_m)$ with the GGN approximation. Next, we compute the
directional slopes and curvatures for \textit{all} mini-batch quadratics
$\quadratic(\emptyarg; \minibatch_{m'})$, $m' \in \{1, \dots, \numminibatches\}$ along
those $100$ eigenvectors. For a fixed eigenvector $\eigvec_p$, the \textit{average} of
those directional slopes/curvatures over \textit{all} mini-batches coincides with the
directional slope/curvature of the full-batch quadratic,
\ie
\begin{equation}
  \frac{1}{\numminibatches} \sum_{m'=1}^\numminibatches
  \underbracket[0.14ex]{
    \dslope{\eigvec_p} \quadratic(\params_\star; \minibatch_{m'})
  }_{\text{one~} \colordot{colorbiased} \text{~and many~} \colordot{colorunbiased}}
  =
  \underbracket[0.14ex]{
    \dslope{\eigvec_p} \quadratic(\params_\star; \traindata)
  }_{\text{\colorcross{colorexact}}}
\quad \text{and} \quad
\frac{1}{\numminibatches} \sum_{m'=1}^\numminibatches
  \underbracket[0.14ex]{
    \dcurv{\eigvec_p} \quadratic(\params_\star; \minibatch_{m'})
  }_{\text{one~} \colordot{colorbiased} \text{~and many~} \colordot{colorunbiased}}
  =
  \underbracket[0.14ex]{
    \dcurv{\eigvec_p} \quadratic(\params_\star; \traindata)
  }_{\text{\colorcross{colorexact}}}\!,
  \label{eq:average_directional_slopes_curvatures}
\end{equation}
derivation in in \cref{sec:details_directional_derivatives}. The colored markers in
\cref{eq:average_directional_slopes_curvatures} refer to 
\cref{fig:bias}. We repeat this procedure for three mini-batches $\minibatch_m$, $m \in
  \{0, 1, 2\}$.

\textbf{Directional slopes and curvatures are biased.} \cref{fig:bias} reveals a
systematic bias in the directional slopes and, more pronounced, in the directional
curvatures: When eigenvectors and directional derivatives are computed on the
\textit{same} mini-batch, curvature is overestimated by roughly one order of magnitude
compared to the curvature of the full-batch quadratic $\quadratic(\emptyarg;
\traindata)$. Within the space that actually carries curvature information---the
top-eigenspace of the quadratic's Hessian---the curvature magnitude is thus not at all
representative of the true underlying loss landscape. For \textit{other} mini-batches,
the directional slopes and curvatures are similar to the full-batch quadratic. This is
because the averages in \cref{eq:average_directional_slopes_curvatures} are dominated by
these unbiased samples.

We present additional results for batch size $2048$ (instead of $512$), the Hessian
$\nabla^2 \Loss(\params_\star; \minibatch_m)$ (instead of its GGN approximation) and \cg
search directions (instead of eigenvectors) in \cref{sec:exp_details_biases}. The bias
is present in \textit{all} settings, but less pronounced for larger batch sizes. For the
\cg search directions, the bias in directional slope is much more distinct than for the
eigenvectors: The biased slope is always negative, while the full-batch quadratic's
slope is in fact positive for most \cg directions!

\subsection{Where do the biases come from?}
\label{sec:biases_theory}

In the following, we provide an explanation for the biases in the directional slopes and
curvatures (\ie the gap between the orange and a blue dot in \cref{fig:bias}) from a
theoretical perspective.

\subsubsection{Bias in the directional slope}

Here, we focus on the \cg search directions since the bias in the directional slope is
even more pronounced for those directions than for the eigenvectors. This is not
accidental! The \cg directions are constructed from gradients---and gradients are
directions that \textit{maximize} the steepness \colordot{colorbiased} for one
particular mini-batch quadratic. For \textit{other} mini-batch quadratics, the steepness
along those directions (\ie the directional slope \colordot{colorunbiased}) is therefore
\textit{less} extreme. We formalize this intuition in \cref{sec:details_bias_slope}.

\subsubsection{Bias in the directional curvature}

Next, we consider the bias in the directional curvature along the eigenvectors of the
curvature matrix. 

\textbf{Directional curvature along $\hessian_\minibatch$'s eigenvectors.}
Let $\eigvec_1, \ldots, \eigvec_{\numparams}$ and $\tilde{\eigvec}_1, \ldots,
\tilde{\eigvec}_{\numparams}$ denote the eigenvectors of two mini-batch Hessians
$\hessian_\minibatch$ and $\hessian_{\otherminibatch}$, respectively. Assume that the
corresponding eigenvalues are in descending order, \ie $\eigval_1 \geq \ldots \geq
\eigval_{\numparams}$ and $\tilde{\eigval}_1 \geq \ldots \geq
\tilde{\eigval}_{\numparams}$. We can write the directional curvatures on $\minibatch$
and $\otherminibatch$ along an eigenvector $\eigvec_i$ as (details in
\cref{sec:details_bias_curvature})
\begin{equation}
  \underbracket[0.14ex]{
    \dcurv{\eigvec_i} \quadratic(\params_\bullet; \otherminibatch)
  }_{\colordot{colorunbiased}}
  =
  \eigvec_i^\top \hessian_{\otherminibatch} \eigvec_i
  =
  \sum_{p=1}^\numparams
  \tilde{\eigval}_p \, \mOmega_{i, p},
\quad \text{and} \quad
\underbracket[0.14ex]{
    \dcurv{\eigvec_i} \quadratic(\params_\bullet; \minibatch)
  }_{\colordot{colorbiased}}
  =
  \eigvec_i^\top \hessian_{\minibatch} \eigvec_i
  =
  \eigval_i.
  \label{eq:bias_curvature_eigvecs}
\end{equation}
The weights $\{\mOmega_{i, p} \defeq (\eigvec_i^\top
  \tilde{\eigvec}_p)^2\}_{p=1}^\numparams$ are non-negative and sum to one, \ie
  $\sum_{p=1}^\numparams \mOmega_{i, p} = 1$. The bias in the directional curvature thus
  originates from misalignment of the eigenspaces of the two curvature
  matrices---captured by the weights $\mOmega_{i, p}$---and/or a systematic difference
  in their spectra.

\textbf{Curvature bias is not due to differing spectra\dots} Assume that the eigenspaces
are perfectly aligned, \ie $\mOmega_{i, i} = 1$ and $\mOmega_{i, p} = 0 \; \forall p
\neq i$. In this case, we obtain $\dcurv{\eigvec_i} \quadratic(\params_\bullet;
\otherminibatch) = \tilde{\eigval}_i$ from \cref{eq:bias_curvature_eigvecs}, so the bias
originates \textit{exclusively} from the differences in the spectra. \Cref{fig:bias}
shows the eigenvalues (as directional curvatures) for three different \cifarhun
mini-batches. As they are very similar, this cannot serve as the main explanation for
the curvature bias.

\begin{figure}[tb!]
  \begin{minipage}[t]{0.485\textwidth}
    \centering
    \vspace{0pt}  \includegraphics[width=0.99\linewidth]{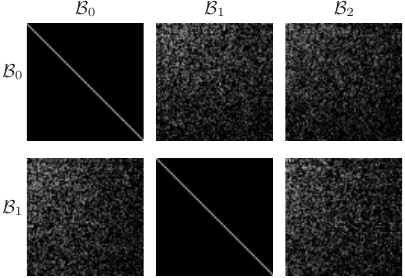}
    \caption{\textbf{In practice, eigenspaces are misaligned.}
We reuse the setting of \cref{fig:bias} and compute the top $100$
    eigenvectors $\eigvecs_m \! \in \R^{\numparams \times 100}$ for
    $\{\minibatch_m\}_{m \in \{0, 1, 2\}}$.
The weights $\mOmega_{i, j}$ are shown as a $100 \times 100$ greyscale image (color
    ranges from black for $\mOmega_{i, j} \leq 10^{-8}$ to white for $\mOmega_{i, j} =
      1$) for $m \in \{0, 1\}$, $m' \in \{0, 1, 2\}$.
Clearly, the eigenspaces for different mini-batches are not perfectly aligned as
    eigenvectors from $\minibatch_m$ overlap with several eigenvectors from
    $\minibatch_{m'}$.}
    \label{fig:overlaps}
  \end{minipage}
  \hfill
  \begin{minipage}[t]{0.485\textwidth}
    \centering
    \vspace{0pt}  \includegraphics[width=0.99\linewidth]{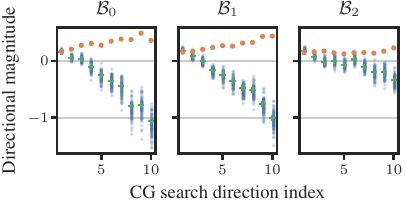}
    \caption{\textbf{\cg update magnitudes are biased.}
Same setting as \cref{fig:bias_ggn_512} \figbottom. We run \cg on
    $\{\minibatch_m\}_{m \in \{0, 1, 2\}}$ and show the directional update magnitudes
    $\tau_1, \ldots, \tau_{10}$ for the first $10$ \cg steps using (i) the same
    mini-batch $\minibatch_m$ (as \colordot{colorbiased}), (ii) all other mini-batches
    (as \colordot{colorunbiased}) and (iii) the entire training set (as
    \colorcross{colorexact}).
The magnitudes are given by the negative ratio of the directional slope and
    curvature (see \cref{eq:cg_update_with_magnitude}) and thus inherit the attached
    biases.
Note that most of the update magnitudes that are based on a single mini-batch of
    data (\colordot{colorbiased}) have the wrong sign resulting in detrimental updates
    in the wrong direction.}
    \label{fig:cg_magnitudes}
  \end{minipage}
\end{figure}

\textbf{\dots but due to misaligned eigenspaces.} Now, assume that the spectra of
$\hessian_\minibatch$ and $\hessian_{\otherminibatch}$ are identical, \ie $\eigval_p =
\tilde{\eigval}_p \, \forall p \in \{1, \ldots, \numparams\}$ which simplifies the
unbiased estimate in \cref{eq:bias_curvature_eigvecs} to
$
  \dcurv{\eigvec_i} \quadratic(\params_\bullet; \otherminibatch)
  =
  \sum_{p=1}^\numparams \eigval_p \, \mOmega_{i, p}
$.
Consider the curvature along $\eigvec_1$ as an example. If $\eigvec_1 =
\tilde{\eigvec}_1$, there is only one non-zero weight $\mOmega_{1, 1} = 1$ and the two
directional curvatures are identical. However, if there is significant overlap with any
other eigenvectors (\ie some weight is distributed also on the \textit{lower}-curvature
directions), the resulting curvature $\dcurv{\eigvec_1} \quadratic(\params_\bullet;
\otherminibatch)$ is \textit{smaller} than $\dcurv{\eigvec_1}
\quadratic(\params_\bullet; \minibatch)$.
Analogously, for $\eigvec_\numparams$, the directional curvature on $\otherminibatch$ is
\textit{larger} than on $\minibatch$ if there is significant overlap between
$\eigvec_\numparams$ and some of $\hessian_{\otherminibatch}$'s
\textit{higher}-curvature eigenvectors.
This can be formalized by the following inequalities (derivation in \cref{sec:details_bias_curvature}):
\begin{equation}
  \underbracket[0.14ex]{
    \dcurv{\eigvec_1} \quadratic(\params_\bullet; \otherminibatch)
  }_{\colordot{colorunbiased}}
  \leq
  \underbracket[0.14ex]{
    \dcurv{\eigvec_1} \quadratic(\params_\bullet; \minibatch)
  }_{\colordot{colorbiased}}
\quad \text{and} \quad
\underbracket[0.14ex]{
    \dcurv{\eigvec_\numparams} \quadratic(\params_\bullet; \otherminibatch)
  }_{\colordot{colorunbiased}}
  \geq
  \underbracket[0.14ex]{
    \dcurv{\eigvec_\numparams} \quadratic(\params_\bullet; \minibatch)
  }_{\colordot{colorbiased}}\!.
  \label{eq:bias_curvature_u1_uP}
\end{equation}
In general, if the eigenspaces are not perfectly aligned such that weight is distributed
on several eigenvectors, this leads to \textit{less extreme} directional curvatures on
mini-batch $\otherminibatch$ on \textit{both} ends of the spectrum.
\Cref{fig:overlaps} shows the weights $\mOmega_{i, p}$ as pixels. The overlap between
the eigenspaces is far from perfect, \ie one eigenvector from $\minibatch$ overlaps with
several eigenvectors from $\otherminibatch$ to some extent.
This explains the curvature overestimation in the top curvature subspace we observe in
\cref{fig:bias} and identifies the misalignment of the eigenspaces as the dominating
factor for the curvature bias.

\subsubsection{Summary of the theoretical findings}

The \cg search directions and the eigenvectors of the curvature matrix are both
\textit{designed} to be ``extreme'' in some sense: The \cg directions are based on
gradients that maximize steepness of the quadratic $\quadratic(\emptyarg; \minibatch)$,
while the top/bottom eigenvectors subsume directions of largest/smallest curvature.
However, these directions are extreme only for one particular mini-batch $\minibatch$.
Another mini-batch $\otherminibatch$ has its own extreme directions that typically
differ from those of $\minibatch$. Thus, \textbf{the extreme directions for
  $\minibatch$ are less extreme for $\otherminibatch$}.
Projecting both quadratics onto $\minibatch$'s directions consequently leads to extreme
steepness and curvatures for $\minibatch$, but less extreme values for
$\otherminibatch$.
This result is an instance of the classic \textbf{regression to the mean}
\citep{galton1886regression}: Using an algorithm to find the directions of most extreme
steepness/curvature in one particular mini-batch, we must expect the steepness/curvature
to be \textit{less} extreme on most \textit{other} batches (and thus also on the entire
dataset). We can expect this phenomenon to occur for other \datasets, models, and curvature
proxies as well, since the underlying mechanism is the stochasticity of the geometric
information.

\section{Debiasing mini-batch quadratics for applications}

We here argue that the biases affect second-order applications, and
develop debiasing strategies.

\subsection{Implications for second-order optimization and the Laplace approximation}
\label{sec:implications_for_applications}

\textbf{Detrimental updates in second-order optimizers.}
In the context of second-order optimization, both the biases in the directional slopes
and curvatures are relevant. From \cref{eq:cg_update_with_magnitude}, the \cg{} update
magnitude $\tau_p$ to reach the minimum of $\quadratic(\params_p + \tau \vd_p;
\minibatch_m)$ is given by the negative ratio of the directional slope and curvature at
the current iterate $\params_p$ in the direction $\vd_p$---that is the 1D Newton step
along that cut.\footnote{The exact same argument can also be made for the eigenvector
directions as the Newton step can be decomposed into 1D Newton steps along the
eigenvectors.} As both these quantities are biased, so is $\tau_p$ as shown in
\cref{fig:cg_magnitudes}: While the update magnitudes for $\minibatch_m$ are
\textit{always positive} (a property of \cg), minimizing the other---\textit{equally
valid}---quadratics would require \textit{negative} update magnitudes for most of the
\cg directions. In this sense, naive \cg on a single mini-batch of data makes updates in
the wrong direction. This can be attributed to the bias in the directional slope since
the slope determines the sign of $\tau_p$. Overestimation of the directional curvature
is ``accidentally beneficial'' in this case as it leads to smaller steps.

\textbf{Unreliable uncertainty quantification with the Laplace approximation.}
For the LA, only the bias in the directional curvature is relevant. After
\cref{sec:laplace_approximation}, the approximate posterior covariance over the
network's parameters is given by $\mSigma_\minibatch = \numtraindata^{-1}
\hessian_\minibatch^{-1}$. Via the eigendecomposition $\hessian_\minibatch =
\sum_{p=1}^\numparams \eigval_p \eigvec_p \eigvec_p^\top$, we obtain $\mSigma_\minibatch
= \numtraindata^{-1} \sum_{p=1}^\numparams \eigval_p^{-1} \eigvec_p \eigvec_p^\top$. Due
to the biases we describe in \cref{sec:geometry_minibatch_quadratic}, the directional
curvatures $\eigval_p$ are not representative of the true underlying curvature,
resulting in a \textit{deformed} posterior covariance $\mSigma_\minibatch$.
Specifically, the overestimation of the curvature in the top-curvature subspace
translates into an \textit{underestimation} of the uncertainty in the posterior
covariance (due to the inversion) with potentially severe consequences in
safety-critical applications.

\subsection{Debiasing strategies}
\label{sec:debiasing_strategies}

We now turn to strategies that mitigate the biases in second-order optimization and the
LA. An empirical evaluation of these methods follows in \cref{sec:exp_cg,sec:exp_LA}.

\textbf{Decoupling directions from magnitudes with a two-batch strategy.} The approaches
we propose are simple yet effective. The idea is to commit to the imperfect directions
from one mini-batch (as before) but use an \textit{additional} mini-batch to estimate
the directional derivatives. With this, we \textit{decouple} the mechanism that
determines the parameter subspace in which the method operates from the slope and
curvature measurements within the space. Effectively, we use the blue dots from
\cref{fig:bias,fig:cg_magnitudes} instead of the orange dots and thus obtain much more
realistic estimates of the actual loss function's geometry (within the subspace defined
by the first mini-batch). Next, we describe in more detail how this strategy can be
applied to debias \cg and the LA.

\subsubsection{Debiased conjugate gradients}
\label{sec:debiased_cg}

\textbf{Debiased approach.} For the debiased \cg method, we need to implement two
processes: (i) The first process applies $\rankbound \leq \numparams$ \cg iterations to
the mini-batch quadratic $\quadratic(\emptyarg; \minibatch)$ and collects the search
directions $(\vd_1, \ldots, \vd_\rankbound)$. As before, this defines the subspace in
which \cg operates. (ii) The second process recomputes the trajectory
$(\tilde{\params}_1, \ldots, \tilde{\params}_\rankbound)$ within that subspace using
debiased update magnitudes that are computed on a different mini-batch
$\otherminibatch$, \ie we use $\tilde{\params}_0 \defeq \params_0$ and 
\begin{equation}
  \tilde{\params}_{p+1} \defeq \tilde{\params}_p + \tilde{\tau}_p \vd_p
  \quad \text{with} \quad
  \underbracket[0.14ex]{\tilde{\tau}_p}_{\colordot{colorunbiased}}
  \defeq 
  -\frac{
    \dslope{\vd_p} \quadratic(\tilde{\params}_p; \otherminibatch)
  }{
    \dcurv{\vd_p} \quadratic(\tilde{\params}_p; \otherminibatch)
  }
  \quad \text{instead of} \quad
  \underbracket[0.14ex]{\tau_p}_{\colordot{colorbiased}}
  =
  -\frac{
    \dslope{\vd_p} \quadratic(\tilde{\params}_p; \minibatch)
  }{
    \dcurv{\vd_p} \quadratic(\tilde{\params}_p; \minibatch)
  }.
  \label{eq:debiased_cg}
\end{equation}
For $\otherminibatch = \minibatch$, $(\tilde{\params}_1, \ldots,
\tilde{\params}_\rankbound)$ is congruent with the original trajectory $(\params_1,
\ldots, \params_\rankbound)$ from the single-batch \cg approach. If the two mini-batches
are different, the debiased trajectory will use updates into the directional minima of
$\quadratic(\emptyarg; \otherminibatch)$ instead of $\quadratic(\emptyarg; \minibatch)$.
Processes (i) and (ii) can either be run side by side in an alternating fashion
(offering more flexibility regarding \eg the termination criterion at the cost of a
small memory overhead, details in \cref{sec:details_cg}) or one after the other.

\subsubsection{Debiased Laplace approximation}
\label{sec:debiased_LA}

\textbf{Laplace approximation with \kfac.} We briefly explore another
popular curvature proxy: Kronecker-Factored Approximate Curvature (\kfac), which is
commonly used both in optimization \citep{martens2015optimizing, martens2018kroneckerfactored,
eschenhagen2023kfac} and uncertainty quantification \citep{Ritter2018ScalableLaplace}.
It is a block-diagonal approximation to the FIM 
$
  \mF_{\minibatch} 
  \approx 
  \mK_{\minibatch} 
  \defeq 
  \blockdiag_{l=1,\ldots, L}(\mK_{\minibatch}^{(l)})
$, 
where each block 
$
  \mK_{\minibatch}^{(l)} 
  \defeq
  \mA^{(l)} \otimes \mB^{(l)}
$ is approximated by the Kronecker product of two smaller
matrices $\mA^{(l)}$ and $\mB^{(l)}$. Sampling from the respective LA with covariance
$\mSigma_\minibatch = \numtraindata^{-1} (\mK_\minibatch + \beta \mI)^{-1}$ can be
performed efficiently due to the specific structure of the \kfac approximation. For
details, see \cref{sec:details_laplace_approximation_sampling}. 

\textbf{Debiased approach.} For the debiased LA, we use two \kfac approximations
$\mK_{\minibatch}$ and $\mK_{\otherminibatch}$ computed on different mini-batches. Via
the eigendecomposition $\mK_{\minibatch} = \eigvecs \eigvals \eigvecs^\top$ with 
$\eigvecs = (\eigvec_1, \ldots, \eigvec_\numparams)$ and $\eigvals = \diag(\eigval_1,
\ldots, \eigval_\numparams)$, we can re-write the covariance matrix as 
$
  \mSigma_\minibatch
  = \numtraindata^{-1} (\mK_{\minibatch} + \beta \mI)^{-1}
  = \numtraindata^{-1} \eigvecs (\eigvals + \beta \mI)^{-1} \eigvecs.
$
For the debiased approach, we keep the eigenspace defined by $\mK_{\minibatch}$ but
recompute the directional curvatures based on $\mK_{\otherminibatch}$, \ie we use
\begin{align}
  \tilde{\mSigma}_\minibatch
  \!=\!
  \frac{1}{\numtraindata} 
  \eigvecs 
  \big(\diag(\tilde{\eigval}_1, \dots, \tilde{\eigval}_\numparams) + \beta \mI\big)^{-1}
  \eigvecs^\top
\;\; \text{with} \;\;
\underbracket[0.14ex]{\tilde{\eigval}_p}_{\colordot{colorunbiased}}
  \!=\! \eigvec_p^\top \mK_{\otherminibatch} \eigvec_p
\;\; \text{instead of} \;\;
\underbracket[0.14ex]{\eigval_p}_{\colordot{colorbiased}}
  \!=\! \eigvec_p^\top \mK_{\minibatch} \eigvec_p.
  \label{eq:debiased_LA}
  \raisetag{2.5ex}  \end{align} 
\vspace{-3ex}

\subsubsection{Computational cost of debiasing}
\label{sec:computational_cost}

Both debiasing techniques roughly double runtime compared to their single-batch
counterparts: Debiased \cg performs one extra matrix-vector product with
$\hessian_{\otherminibatch}$ per \cg iteration to compute the debiased update magnitude
(details in \cref{sec:details_cg}). Debiased LA requires an additional \kfac
approximation $\KFAC_{\otherminibatch}$; all subsequent operations to compute the
debiased \kfac can be carried out efficiently at Kronecker factor level
(details in \cref{sec:details_debiased_kfac}).
In \cref{sec:experiments}, we thus use the debiased versions at \textit{half} the batch size,
for a fair comparison. We will see that, at the resulting \textit{similar} computational cost, the debiased versions clearly outperform the
single-batch alternatives.

\section{Related work}

We here briefly list other, related forms of bias correction that have been suggested
elsewhere.

\textbf{A different two-batch approach.} Other works have proposed to use different (not
necessarily disjoint) mini-batches for the gradient and the Hessian
\citep{martens2010deep,byrd2011use}. 
\citet[Sec. I.5]{benzing2022gradient} mentions the idea of using \textit{independent}
mini-batches for the gradient and the Hessian to obtain an, in some sense, unbiased
estimate
$
- \hessian_\minibatch^{-1} \grad_{\otherminibatch}
$
of the exact Newton step. This does, however, not resolve the biases described in this
paper. Via the eigendecomposition of the Hessian $\hessian_\minibatch = \sum_{p=1}^P
\eigval_p \eigvec_p \eigvec_p^\top$, we obtain 
$
  - \hessian_\minibatch^{-1} \grad_{\otherminibatch}
\!=\!
  - \sum_{p=1}^P 
    \dslope{\eigvec_p} \quadratic(\params_\star; \otherminibatch)
    (\dcurv{\eigvec_p} \quadratic(\params_\star; \minibatch))^{-1} \,
    \eigvec_p.
$
While the numerator yields an unbiased estimate of the directional slope (similar to a
\textit{blue} dot in the upper panel of \Cref{fig:bias}), the bias in the denominator
remains since eigenvectors and directional curvatures are based on the same mini-batch
(as for the \textit{orange} dots in the bottom panel of \Cref{fig:bias}).

\textbf{Another debiasing approach.} \ekfac (Eigenvalue-corrected Kronecker
Factorization) \citep{George2018Fast} corrects the eigenvalues of the \kfac
approximation (similar to \cref{eq:debiased_LA}) which, provably, yields a more accurate
approximation of the FIM than \kfac (in Frobenius norm). This correction, however, is
designed to resolve a different kind of bias that is specific to the \kfac approximation
and does not address the biases described in this work.

\textbf{Running averages.} Other popular deep learning optimizers \textit{aggregate}
curvature information over multiple steps via exponential moving averages (see \eg
\citep{martens2015optimizing}). This emulates larger mini-batch sizes and thus reduces
the curvature biases. However, when curvature evolves rapidly, obsolete curvature
estimates from past steps might slow down training. Aggregating more robust
\textit{debiased} curvature estimates instead might allow for shorter moving average
windows and accelerate the optimization progress. We leave it to future work to explore
these interactions.

\section{Experiments}
\label{sec:experiments}

In this section, we evaluate the effectiveness of the debiasing strategies from
\cref{sec:debiasing_strategies}. In \cref{sec:experimental_details}, we provide the
experimental details as well as additional empirical analyses. For instance, we show how
the curvature biases with \kfac depend on the mini-batch size (see \cref{sec:exp_kfac}),
how the biases evolve over the course of training (see \cref{sec:exp_biases_over_time}),
and that the curvature biases become more pronounced for deeper/wider models (see
\cref{sec:exp_numparams}).

\subsection{Debiased conjugate gradients}
\label{sec:exp_cg}

We compare the standard single-batch \cg method to the debiased version (see
\cref{sec:debiased_cg}) on the fully trained \allcnnc model without curvature damping.
For a fair comparison, the single-batch approach uses one mini-batch of size $1024$
while the debiased approach uses two mini-batches of size $512$, such that a similar
amount of data and runtime budget is used. Details in \cref{sec:exp_details_cg}.

\begin{figure}[tbh]
  \centering
  \includegraphics[width=0.99\textwidth]{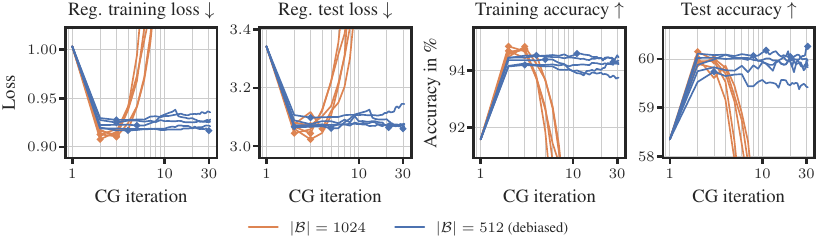}
  \caption{\textbf{Debiased \cg is much more stable than the single-batch approach.} We
  compare \cg runs without curvature damping ($\delta = 0$) with $\rankbound = 30$
  iterations for the fully trained \allcnnc model on the \cifarhun \dataset in terms of
  training/test loss/accuracy at similar computational cost: The single-batch approach
  (shown as \colorline{colorbiased}) uses one mini-batch of size $1024$ while the
  debiased approach (shown as \colorline{colorunbiased}) uses two mini-batches of size
  $512$ each. Both approaches use the GGN curvature proxy and are run $5$ times on
  different mini-batches. The markers \colordiamond{colorbiased} and
  \colordiamond{colorunbiased} are placed at peak performance. 
While the single-batch runs diverge quickly, the debiased \cg runs are stable.}
  \label{fig:cg}
  \end{figure}

  \textbf{Results \& discussion.} 
All \cg runs in \cref{fig:cg} achieve a significant improvement in all four performance
metrics. The most striking difference between the two approaches is their stability: The
single-batch runs quickly reach peak performance and then diverge. In contrast, although
the steps tend to be \textit{larger} (see \cref{fig:cg_magnitudes}), the debiased \cg
runs are much more stable (without using any curvature damping). 
This suggests that the reason for the divergence in the single-batch approach is
\textit{not} the missing damping but the misinformed update magnitudes.
The peak performance is slightly better for the single-batch runs which can be
attributed to its more informative search directions (they were computed using
\textit{double} the data). The peak performance of the debiased runs could likely be
improved (at the same computational cost) by using a larger batch size for the
directions and a smaller one for the update magnitudes (as the former seem harder to
estimate, see \cref{sec:biases_theory}).

\subsection{Debiased Laplace approximation}
\label{sec:exp_LA}

Here, we use a fully trained \allcnnc model on \cifarten data and compare (i) the
vanilla model without LA, (ii) the single-batch \kfac LA approach, (iii) the debiased
version (see \cref{sec:debiased_LA}), and (iv) the full-batch approach (where we compute
\kfac on the entire training set) in terms of accuracy, NLL and ECE. Again, we apply the
debiased approach at half the batch size of the single-batch approach for a fair
comparison. We use prior precisions between $10^{-4}$ and $10$.
\Cref{sec:exp_details_la} contains the experimental details and additional results on
the training and OOD data. 

\begin{figure}[tbh]
  \centering
  \includegraphics[width=0.99\textwidth]{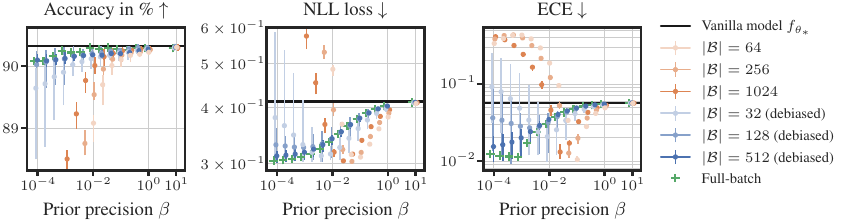}
  \caption{\textbf{Debiased LA mimics the full-batch LA very well.} We compare LAs for
  the fully trained \allcnnc model on \cifarten in terms of accuracy, negative log
  likelihood loss (NLL) and expected calibration error (ECE) on the \cifarten test set.
  For each mini-batch size (lighter color indicating smaller batch size), we draw $5$
  mini-batches and report the mean performance as dot and min/max as vertical line. We
  also report the performance of the vanilla model without LA (shown as
  \colorline{black}) and the full-batch approach based on $\mK_{\traindata}$ (shown as
  \colorcross{colorexact}). 
In contrast to the single-batch approach, the debiased version mimics the behavior of
  the full-batch approach very well over the entire range of prior precisions. }
  \label{fig:la}
\end{figure}

\textbf{Results \& discussion.} \Cref{fig:la} shows the results. If the prior precision
is low (\ie the LA relies mainly on the curvature information without the regularizing
diagonal term), the single-batch version acts erratically due to the deformed curvature
model---its performance drops drastically. The single-batch approach is also more
sensitive to the choice of prior precision (in fact, it suggests a much larger prior
precision than the full-batch approach). In contrast, the debiased approach achieves
good performance over a wider range of prior precisions and mimics the full-batch
approach ($\numtraindata\!=\!\num{40000}$) very well despite using only a tiny fraction
of the data. 

In order to showcase our debiasing strategy's scalability, we provide additional results
on \resnetfifty and \vitlittle on the \imagenet \dataset in \cref{sec:exp_details_la}.

\textbf{Summary of the experimental results.} The use of mini-batch quadratics is a
simple way to keep the costs of second-order optimization and uncertainty quantification
manageable. The resulting biases, however, severely degrade their value, requiring large 
mini-batches or algorithmic add-ons. Our experiments suggest that
even simple debiasing strategies can largely mitigate this issue.

\section{Conclusion}

Our main takeaway is a general principle: \textbf{Quadratic approximations to the
training loss computed on \textit{mini-batches} of the training data provide a
\textit{distorted} representation of the true underlying loss landscape.} In particular,
along the directions of large curvature, the mini-batch quadratics tend to strongly
overestimate the curvature of the true loss.
Our theoretical analysis shows that these biases can be traced back to the misalignment
of the curvature matrices' eigenspaces.
These insights are highly relevant for applications: As we demonstrated empirically, the
biases in the directional slope and curvature lead to severely misinformed updates in
stochastic second-order optimizers, and cause unreliable uncertainty estimates with
Laplace approximations.
We also proposed simple two-batch strategies to mitigate these biases. Our experiments
demonstrate their superiority over the single-batch approaches in terms of stability and
quality at similar computational costs.
Our findings reveal a design prerequisite for building better stochastic curvature-based
methods, which should be generally considered, and further developed, for all methods
using such curvature evaluations.

\subsubsection*{Acknowledgments}

The authors gratefully acknowledge co-funding by the European Union (ERC, ANUBIS,
101123955). Views and opinions expressed are however those of the author(s) only and do
not necessarily reflect those of the European Union or the European Research Council.
Neither the European Union nor the granting authority can be held responsible for them.
Philipp Hennig is a member of the Machine Learning Cluster of Excellence, funded by the
Deutsche Forschungsgemeinschaft (DFG, German Research Foundation) under Germany's
Excellence Strategy - EXC number 2064/1 - Project number 390727645; The authors further
gratefully acknowledge financial support by the DFG through Project HE 7114/5-1 in
SPP2298/1; the German Federal Ministry of Education and Research (BMBF) through the
Tübingen AI Center (FKZ:01IS18039A); and funds from the Ministry of Science, Research
and Arts of the State of Baden-Württemberg. Lukas Tatzel and Bálint Mucsányi are
grateful to the International Max Planck Research School for Intelligent Systems
(IMPRS-IS) for support. Further, the authors thank Felix Dangel and Runa Eschenhagen for
providing feedback to the manuscript.

\bibliography{bibliography}
\bibliographystyle{iclr2025_conference}

\clearpage

\appendix

\section*{Supplementary material}

Below, we provide additional details on the mathematical derivations, describe the
experimental setup and present additional results. 

\startcontents[sections]
\printcontents[sections]{l}{1}{\setcounter{tocdepth}{3}}

\clearpage

\section{Mathematical details}

\subsection{Directional derivatives of a quadratic}
\label{sec:details_directional_derivatives}

We claim in \cref{sec:background_quadratic_approximations} that a cut $r$ through the
quadratic $\quadratic(\emptyarg; \minibatch)$ from $\params_\bullet \in \R^\numparams$
along the normalized direction $\vd$ can be written as
\begin{equation}
  r(\tau) 
  \defeq 
  \quadratic(\params_\bullet + \tau \vd; \minibatch)
  = 
  \frac{1}{2} \tau^2 \vd^\top \nabla^2 \quadratic(\params_\bullet; \minibatch) \, \vd 
  + \tau \vd^\top \nabla \quadratic(\params_\bullet; \minibatch)
  + \text{const}.
  \label{eq:cut_through_quadratic}
\end{equation}

\textbf{Proof for \cref{eq:cut_through_quadratic}.} Here, we provide the derivation for
\cref{eq:cut_through_quadratic}. Let $\params_\bullet \in \R^\numparams$ be a point in
parameter space and $\vd \in \R^\numparams$ a normalized direction, \ie $\Vert\vd\Vert =
1$. We consider the quadratic approximation $\quadratic(\params; \minibatch)$ around
$\params_0$ and evaluate it along the cut $\params_\bullet + \tau \vd$ for $\tau \in
\R$. We assume that the Hessian (or its approximation) $\hessian_\minibatch$ is
symmetric and obtain
\begin{align*}
  r(\tau) 
  \defeq& \;
  \quadratic(\params_\bullet + \tau \vd; \minibatch)\\
=& \;
  \frac{1}{2} (\params_\bullet + \tau \vd - \params_0)^\top \hessian_\minibatch 
  (\params_\bullet + \tau \vd - \params_0) 
  + (\params_\bullet + \tau \vd - \params_0)^\top \grad_\minibatch
  + \const_\minibatch\\
=& \;
  \frac{1}{2} \tau^2 \vd^\top \hessian_\minibatch \vd 
  + \tau \vd^\top \hessian_\minibatch (\params_\bullet - \params_0) 
  + \frac{1}{2} (\params_\bullet - \params_0)^\top 
    \hessian_\minibatch 
    (\params_\bullet - \params_0)\\
  & + \tau \vd^\top \grad_\minibatch
  + (\params_\bullet - \params_0)^\top \grad_\minibatch
  + \const_\minibatch\\
  =& \; 
  \frac{1}{2} \tau^2 \vd^\top \hessian_\minibatch \vd 
  + \tau \vd^\top (\hessian_\minibatch (\params_\bullet - \params_0) + \grad_\minibatch)
  + \text{const}.
\end{align*}
Since 
$
  \nabla \quadratic(\params_\bullet; \minibatch)
  = 
  \hessian_\minibatch (\params_\bullet - \params_0) + \grad_\minibatch
$
and
$
  \nabla^2 \quadratic(\params_\bullet; \minibatch)
  \equiv 
  \hessian_\minibatch,
$
we arrive at \cref{eq:cut_through_quadratic}. \qedsymbol

\textbf{Proof for \cref{eq:average_directional_slopes_curvatures}.} Next, we show that
the average directional slope/curvature over all mini-batches in the training set
coincides with the directional slope/curvature of the full-batch quadratic. Let's assume
that all $\numminibatches$ mini-batches $\{\minibatch_{m'}\}_{m'=1}^\numminibatches$ are
disjoint, have the same size $\vert \minibatch_1 \vert$ and that their union is the
training set, \ie
\begin{equation}
  \vert \minibatch_{m'} \vert 
  = \vert \minibatch_1 \vert \;\; \forall m' \in \{1, \ldots, \numminibatches\}
  \quad \text{and} \quad
  \bigcup_{m'=1}^\numminibatches \minibatch_{m'} = \traindata.
  \label{eq:assumtions_minibatches_trainingdata}
\end{equation}
This implies $\numminibatches \vert \minibatch_1 \vert = \vert \traindata \vert$. It
holds
\begin{align*}
  \frac{1}{\numminibatches} \sum_{m'=1}^\numminibatches
  \underbracket[0.14ex]{
    \dslope{\eigvec_p} \quadratic(\params_\star; \minibatch_{m'}) 
  }_{\text{one~} \colordot{colorbiased} \text{~and many~} \colordot{colorunbiased}}
  &= 
  \frac{1}{\numminibatches} \sum_{m'=1}^\numminibatches
  \eigvec_p^\top 
  \nabla \quadratic(\params_\star; \minibatch_{m'}) 
  \quad \text{
    with 
    $\nabla \quadratic(\params_\star; \minibatch_{m'}) = \grad_{\minibatch_{m'}}$
    since $\params_0 \gets \params_\star$
  } \\
  &= 
  \eigvec_p^\top \frac{1}{\numminibatches} \sum_{m'=1}^\numminibatches
  \nabla \Lossreg(\params_\star; \minibatch_{m'}) \\
  &= 
  \eigvec_p^\top \left(
    \frac{1}{\numminibatches} \numminibatches \, \nabla\regularizer(\params_\star) 
  + \frac{1}{\numminibatches} \sum_{m'=1}^\numminibatches
  \nabla \Loss(\params_\star; \minibatch_{m'})
  \right) \\
  &= 
  \eigvec_p^\top \left(
    \nabla\regularizer(\params_\star) 
  + \frac{1}{\numminibatches} \sum_{m'=1}^\numminibatches
  \frac{1}{\vert \minibatch_{m'} \vert} \sum_{n \in \minibatch_{m'}}
  \nabla \loss(\network_{\params_\star}(\vx_n), \vy_n)
  \right) \\
  &\overset{\text{(\ref{eq:assumtions_minibatches_trainingdata})}}{=}
  \eigvec_p^\top \left(
    \nabla\regularizer(\params_\star) 
  + \frac{1}{\numminibatches \vert \minibatch_{1} \vert} 
  \sum_{m'=1}^\numminibatches \sum_{n \in \minibatch_{m'}}
  \nabla \loss(\network_{\params_\star}(\vx_n), \vy_n)
  \right) \\
  &=
  \eigvec_p^\top \nabla \Lossreg(\params_\star; \traindata) \\
  &=
  \underbracket[0.14ex]{
    \dslope{\eigvec_p} \quadratic(\params_\star; \traindata)
  }_{\text{\colorcross{colorexact}}}\!.
\end{align*}
As the mini-batch curvature $\nabla^2 \Loss(\params_\star; \minibatch_{m'})$ is also an
\textit{average} over the samples in the mini-batch (this applies to both the actual
Hessian and its GGN approximation), the same argument holds for the associated
directional curvatures. For \kfac, the derivation above does \textit{not} hold as $(1/2)
(\mK_{\minibatch} + \mK_{\otherminibatch}) \neq \mK_{\minibatch \cup \otherminibatch}$.
\qedsymbol

\subsection{The method of conjugate gradients (\cg)} 
\label{sec:details_cg}

\textbf{Using \cg for minimizing a quadratic.}
The argmin of a quadratic $\quadratic(\emptyarg; \minibatch)$ is given by $\params_0 -
\hessian_\minibatch^{-1} \grad_\minibatch$ (assuming that $\hessian_\minibatch$ is
symmetric and positive definite), see \cref{sec:background_second_order_methods}. We can
use the method of conjugate gradients (see \cref{alg:cg}) for computing the Newton step
(or an approximation thereof) by setting $\mA \gets \hessian_\minibatch$ and $\vb \gets
-\grad_\minibatch$.

\begin{algorithm}[H]
  \caption{Method of conjugate gradients (\cg), based on \citep[Alg.
  5.2]{nocedal2006numerical}}
  \label{alg:cg}
\textbf{Input.} Access to matrix-vector products $\vv \mapsto \mA \vv$, where $\mA \in
  \R^{\numparams \times \numparams}$ is symmetric and positive definite, right-hand side
  $\vb \in \R^{\numparams}$, convergence tolerance $\epsilon \in \R_{> 0}$, maximum
  number of iterations $\numparams_\text{max} \in \N$, $\numparams_\text{max} \leq
  \numparams$ \\[1ex]
\textbf{Output.} Approximate solution $\vx_p$ to the linear system $\mA \vx = \vb$
  \vspace{1ex}
\begin{algorithmic}[1]
    \State Initialize $\vx_0 = \vzero$, $\vr_0 \gets -\vb$ and $\vs_0 \gets -
    \vr_0$
\For{$p = 0, 1, \ldots, \numparams$}
\If{$p = \numparams_\text{max}$ \textbf{or} $\Vert \vr_p \Vert_2 \leq \epsilon$}
    \Comment{Termination criteria}
    \State \textbf{return} (approximate) solution $\vx_p$
    \EndIf
\State $\displaystyle \vt_p \gets \mA \vs_p$ \Comment{Compute the matrix-vector product only once}
    \State $\displaystyle\alpha_p \gets \frac{\vr_p^\top \vr_p}{\vs_p^\top \vt_p}$
    \Comment{Update magnitude 
    $\displaystyle \alpha_p = \frac{\vr_p^\top \vr_p}{\vs_p^\top \mA \vs_p}$
    }
    \State $\vx_{p+1} \gets \vx_p + \alpha_p \vs_p$
    \Comment{Update along search direction $\vs_p$} 
    \State $\vr_{p+1} \gets \vr_p + \alpha_p \vt_p$
    \Comment{Residual $\vr_p = \mA \vx_p - \vb$ computed via recursion}
    \State $\displaystyle\beta_{p+1} \gets \frac{\vr_{p+1}^\top \vr_{p+1}}{\vr_p^\top \vr_p}$
    \State $\vs_{p+1} \gets - \vr_{p+1} + \beta_{p+1} \vs_p$ 
    \Comment{Construction of new (conjugate) search direction}
    \EndFor
  \end{algorithmic}
\end{algorithm}

\textbf{Properties of \cg \& geometric interpretations.} In the following, we provide
some important properties and a geometric interpretation of the quantities involved in
the \cg method.

\begin{itemize}
  \item \textbf{Residual $\vr_p$.} Let $\vx_p \defeq \params_p - \params_0$. The residual
  $\vr_p$ can be written as 
  $
    \vr_p 
    = 
    \mA \vx_p - b 
    = 
    \hessian_\minibatch (\params_p - \params_0) + \grad_\minibatch
    =
    \nabla \quadratic(\params_p; \minibatch)
  $, \ie it coincides with the gradient of the quadratic at $\params_p$.

  \item \textbf{Update magnitude $\alpha_p$.} Consider a cut through the quadratic from
  $\params_p$ into the direction $\vs_p$, \ie $r(\tau) = \quadratic(\params_p + \tau
  \vs_p; \minibatch)$. Minimizing this 1D quadratic requires its first derivative to
  vanish. It holds (see \cref{eq:cut_through_quadratic})
  \begin{align*}
    r'(\tau) = 0
    \quad \Leftrightarrow \quad
    &\tau \vs_p^\top \nabla^2 \quadratic(\params_p; \minibatch) \, \vs_p 
    + \vs_p^\top \nabla \quadratic(\params_p; \minibatch) = 0 \\
    \Leftrightarrow \quad
    &\tau = -\frac{
      \vs_p^\top \nabla \quadratic(\params_p; \minibatch)
    }{
      \vs_p^\top \nabla^2 \quadratic(\params_p; \minibatch) \, \vd_p
    }
    = -\frac{\vs_p^\top \vr_p}{\vs_p^\top \mA \, \vs_p}
    = \frac{\vr_p^\top \vr_p}{\vs_p^\top \mA \vs_p}
    = \alpha_p,
  \end{align*}
  where the equality $-\vs_p^\top \vr_p = \vr_p^\top \vr_p$ is due to Equation (5.14a)
  and Theorem 5.2 in \citep{nocedal2006numerical}. So, the update magnitude $\alpha_p$
  is chosen such that it minimizes the quadratic along the direction $\vs_p$. Note that
  the update magnitude can also be written as a ratio of the negative directional slope
  and directional curvature at $\params_p$, \ie a 1D directional Newton step. Let $\vd_p
  \defeq \vs_p / \Vert \vs_p \Vert$ denote the \textit{normalized} search direction. It
  holds
  \begin{equation*}
    \vx_{p+1} 
    = \vx_p + \alpha_p \vs_p
    = \vx_p -\frac{\vs_p^\top \vr_p}{\vs_p^\top \mA \, \vs_p} \vs_p
    = \vx_p -\frac{\vd_p^\top \vr_p}{\vd_p^\top \mA \, \vd_p} \vd_p
    = \vx_p 
      \underbracket[0.14ex]{
        -
        \frac{
          \dslope{\vd_p} \quadratic(\params_d; \minibatch)
        }{
          \dcurv{\vd_p} \quadratic(\params_d; \minibatch)
        }
       }_{\displaystyle \eqdef \tau_p}
       \vd_p
  \end{equation*}
  \ie $\tau_p = \alpha_p \Vert \vs_p \Vert$. Via the shift $\params_p = \params_0 +
  \vx_p$, we arrive at \cref{eq:cg_update_with_magnitude}.

  \item \textbf{Search direction $\vs_p$.} \cg constructs the search directions to be
  conjugate, \ie $\vs_{p+1}^\top \mA \, \vs_i = 0$ for all $i \leq p$. Note that this
  property also applies to the normalized search directions. 
\end{itemize}

\textbf{Efficient implementation of the debiased \cg approach.} For the debiased \cg
version, we need to re-evaluate the update magnitudes $\tilde{\tau}_p$ for a given set
of search directions $\{\vd_1, \ldots \vd_\numparams\}$ on a second mini-batch
$\otherminibatch$ (see \cref{eq:debiased_cg}), \ie
\begin{equation*}
  \tilde{\tau}_p
  =
  -\frac{
    \dslope{\vd_p} \quadratic(\tilde{\params}_p; \otherminibatch)
  }{
    \dcurv{\vd_p} \quadratic(\tilde{\params}_p; \otherminibatch)
  }
  =
  -\frac{
    \vd_p^\top \nabla \quadratic(\tilde{\params}_p; \otherminibatch)
  }{
    \vd_p^\top \nabla^2 \quadratic(\tilde{\params}_p; \otherminibatch) \, \vd_p
  }
  =
  -\frac{
    \vd_p^\top (\hessian_{\otherminibatch} (\tilde{\params}_p - \params_0) + \grad_{\otherminibatch})
  }{
    \vd_p^\top \hessian_{\otherminibatch} \, \vd_p
  }
\end{equation*}
Implemented naively, this requires \textit{two} matrix-vector products---one for the 
numerator and one for the denominator. However, we can use a recursive formula for the
numerator:
\begin{align}
  \nabla \quadratic(\tilde{\params}_p; \otherminibatch)
  &=
  \hessian_{\otherminibatch} 
  (\tilde{\params}_p - \params_0) 
  + \grad_{\otherminibatch}\nonumber\\
  &=
  \hessian_{\otherminibatch} 
  (\tilde{\params}_{p - 1} + \tilde{\tau}_{p - 1} \vd_{p - 1} - \params_0) 
  + \grad_{\otherminibatch}\nonumber\\
  &= 
  \hessian_{\otherminibatch} 
  (\tilde{\params}_{p - 1} - \params_0)
  + \grad_{\otherminibatch} 
  + \tilde{\tau}_{p - 1} \hessian_{\otherminibatch} \, \vd_{p - 1}\nonumber\\
  &=
  \nabla \quadratic(\tilde{\params}_{p - 1}; \otherminibatch) 
  + \tilde{\tau}_{p - 1} \hessian_{\otherminibatch} \, \vd_{p - 1}.
  \label{eq:cg_recursive_gradient}
\end{align}
So, if we store the gradient from the previous iteration $\nabla
\quadratic(\tilde{\params}_{p - 1}; \otherminibatch)$ and the Hessian vector product
with the previous direction $\hessian_{\otherminibatch} \, \vd_{p - 1}$, the current
gradient can be computed \textit{without} an additional matrix-vector product. At
iteration $p$, we thus (i) compute the gradient $\nabla \quadratic(\tilde{\params}_p;
\otherminibatch)$ recursively from the cached vectors $\nabla
\quadratic(\tilde{\params}_{p - 1}; \otherminibatch)$ and $\hessian_{\otherminibatch} \,
\vd_{p - 1}$ via \cref{eq:cg_recursive_gradient} (for $p=0$, we have $\nabla
\quadratic(\tilde{\params}_0; \otherminibatch) = \grad_{\otherminibatch}$), (ii) compute
the Hessian-vector product with the current direction $\hessian_{\otherminibatch} \,
\vd_{p}$, (iii) store both these vectors and (iv) compute the update magnitude
$\tilde{\tau}_p$ (both the numerator and the denominator only require a simple dot
product of two pre-computed vectors).

\subsection{Laplace approximation for neural networks}
\label{sec:details_laplace_approximation}

\subsubsection{Derivation of the Laplace approximation for neural networks}

\textbf{Preliminaries.} 
The softmax function $\softmax : \R^\numclasses \to \R^\numclasses$ is defined as 
\begin{equation*}
    \softmax(\vz) 
    \defeq 
    \left(
        \frac{\exp(z_1)}{\sum_{c'=1}^{\numclasses} \exp(z_{c'})},
        \ldots,
        \frac{\exp(z_{\numclasses})}{\sum_{c'=1}^{\numclasses} \exp(z_{c'})}
    \right).
\end{equation*}
It maps an arbitrary vector $\vz \in \R^{\numclasses}$ to a vector whose entries are
non-negative and sum up to one. So, the output of the softmax function can be
interpreted as a probability distribution over $\numclasses$ classes. With this, the
cross-entropy loss for a single datum $(\vx, \vy)$ is given by
\begin{equation}
    \loss(\network_{\params}(\vx), \vy)
    \defeq 
    - \sum_{c=1}^\numclasses \vy_c \cdot \log(\softmax(\network_{\params}(\vx))_c).
  \label{eq:cross_entropy}
\end{equation}
Here, we assume that $\vy \in \{0, 1\}^\numclasses$ is a one-hot encoded vector
representing the true class label. So, if $c_\star \in \{1, \ldots, \numclasses\}$ is the
correct class (\ie $\vy_{c_\star} = 1$), the cross-entropy loss is given by the negative
logarithm of the probability the network assigns to this class
$
    \loss(\network_{\params}(\vx), \vy) 
    = 
    - \log(\softmax(\network_{\params}(\vx))_{c_\star}).
$
Finally, let 
\begin{equation}
  \categoricalpdf{\vy}{\vp} 
  \defeq 
  \prod_{c=1}^{\numclasses} \vp_c^{\vy_c}
  \label{eq:categorical_pdf}
\end{equation}
denote the probability mass function of the categorical distribution, where $\vp \in
\R^\numclasses$ is a vector of probabilities (\ie $\vp_c \geq 0$ and $\sum_{c} \vp_c =
1$), and $\vy \in \{0, 1\}^\numclasses$ is a one-hot encoded vector representing a class
label.

\textbf{Probabilistic interpretation of the regularized loss.} Recall from
\cref{eq:Loss_reg} that the regularized loss function is given by
\begin{equation*}
    \Lossreg(\params; \traindata) 
    = 
    \Loss(\params; \traindata) + \regularizer(\params)
    \quad \text{with} \quad
    \Loss(\params; \traindata)
    =
    \frac{1}{\numtraindata} 
    \sum_{n \in \traindata}
    \loss(\network_{\params}(\vx_n), \vy_n).
\end{equation*}
We assume a classification problem that uses the cross-entropy loss (see
\cref{eq:cross_entropy}) and an $\normltwo$ regularizer $\regularizer(\params) =
\nicefrac{\beta}{2} \, \Vert\params\Vert_2^2$ with parameter $\beta \in \R_{> 0}$. We
use one-hot encoded labels $\vy_n \in \{0, 1\}^\numclasses$ (with $\vy_{nc}$, we denote
the $c$-th entry of $\vy_n$) and obtain
\begin{align*}
    \numtraindata \cdot \Lossreg(\params; \traindata) 
    &= 
    \sum_{n \in \traindata} 
    \loss(\network_{\params}(\vx_n), \vy_n)
    + \frac{\numtraindata \beta}{2} \Vert\params\Vert_2^2\\
\overset{\text{(\ref{eq:cross_entropy})}}&{=}
    - \sum_{n \in \traindata} 
    \sum_{c=1}^\numclasses \vy_{nc} \cdot \log(\softmax(\network_{\params}(\vx_n))_c)
    + \frac{\numtraindata \beta}{2} \Vert\params\Vert_2^2\\
\overset{\text{(\ref{eq:categorical_pdf})}}&{=}
    - \sum_{n \in \traindata}
    \log
    \big( 
      \categoricalpdf{\vy_n}{\softmax(\network_{\params}(\vx_n))}
    \big)
    - 
    \left(
      -\frac{1}{2} \params^\top (\numtraindata \beta \cdot \mI) \, \params
    \right)\\
    &= 
    - \log
    \Bigg(
      \underbrace{
        \prod_{n \in \traindata} 
        \categoricalpdf{\vy_n}{\softmax(\network_{\params}(\vx_n))}
      }_{\displaystyle \text{likelihood} \; p(\trainset \mid \params)}
    \Bigg)
    - \log
    \Bigg(
      \underbrace{
        \gaussianpdf{\theta}{\vzero}{\frac{1}{\numtraindata \beta} \mI}
      }_{\displaystyle \text{prior} \; p(\params)}
    \Bigg)
    - Z,
\end{align*}
where
$
  Z \defeq \nicefrac{\numparams}{2} \log(\nicefrac{2 \pi}{\numtraindata \beta})
$
absorbs the normalization constant of the Gaussian prior. The cross entropy loss $\Loss$
for the training set is thus connected to the negative log categorical likelihood and
the regularizer can be seen as a negative log Gaussian prior over the parameters. Note
that a similar derivation is possible for other loss functions as well, \eg the MSE loss
(which is equivalent to a negative log Gaussian likelihood). 

The derivation above shows that the (rescaled) regularized loss function can be
interpreted as the negative unnormalized log posterior of a Bayesian model
\begin{equation}
    \numtraindata \cdot \Lossreg(\params; \traindata) 
    \eqc 
    - \log p(\trainset \mid \params) - \log p(\params)
    \eqc
    - \log p(\params \mid \trainset)
  \label{eq:Lossreg_as_log_posterior}
\end{equation}
with Gaussian prior and categorical likelihood. With $\eqc$, we denote equality up to an
additive constant.

\textbf{Training as MAP estimation.} \Cref{eq:Lossreg_as_log_posterior} allows
re-interpreting the training procedure of a neural network as a maximum a posteriori
(MAP) estimation problem since minimizing the regularized loss 
\begin{equation*}
  \underbrace{
    \argmin_{\params \in \R^\numparams} 
    \mathcal{L}_{\text{reg}}(\params, \traindata) 
  }_{\displaystyle \phantom{\params_\star} = \params_\star}
  = \argmin_{\params \in \R^\numparams} 
  \numtraindata \cdot \mathcal{L}_{\text{reg}}(\params, \traindata)
  \overset{\text{(\ref{eq:Lossreg_as_log_posterior})}}{=} 
  \argmax_{\params \in \R^\numparams}
  \log p(\params \mid \trainset) 
  = \underbrace{
    \argmax_{\params \in \R^\numparams}
    p(\params \mid \trainset) 
  }_{\displaystyle \phantom{\paramsmap} \eqdef \paramsmap}
\end{equation*}
is equivalent to maximizing the posterior $p(\params \mid \trainset)$.

\textbf{Laplace approximation.} The idea of the Laplace approximation is to approximate
the posterior distribution $p(\params \mid \trainset) \approx
\gaussianpdf{\params}{\paramsmap}{\mSigma_\traindata}$ with a Gaussian at the mode
$\paramsmap$ of the posterior, \ie after training the network. For this, we approximate
the log posterior by a second-order Taylor expansion around the mode $\params_0 \gets
\paramsmap$, \ie
\begin{equation}
  \log p(\params \mid \trainset)
  \eqc
  - \numtraindata \cdot \Lossreg(\params; \traindata)
\overset{(\ref{eq:fullbatch_quadratic})}{\approx}
  - \numtraindata \cdot \quadratic(\params; \traindata)
  \eqc
  - \frac{1}{2} 
  (\params - \paramsmap)^\top 
  (\numtraindata \cdot \hessian_\traindata) 
  (\params - \paramsmap),
  \label{eq:laplace_fullbatch_quadratic}
\end{equation}
where we assumed that $\paramsmap$ is a (local) minimum of the regularized loss function
(\ie $\grad_\traindata = \nabla \Lossreg(\paramsmap; \traindata) = 0$) such that the
linear term $(\params - \params_0)^\top \grad_\traindata$ in the quadratic approximation
vanishes. The additive constants we did not state explicitly in
\cref{eq:laplace_fullbatch_quadratic} turn into some multiplicative factor (denoted by
$Z^{-1}$) when taking the exponential
\begin{equation*}
  p(\params \mid \trainset)
  \approx
  \frac{1}{Z}
  \exp\left(
    - \frac{1}{2} 
    (\params - \paramsmap)^\top 
    (\numtraindata \cdot \hessian_\traindata) 
    (\params - \paramsmap)
  \right).
\end{equation*}
This immediately identifies $Z$ as the normalization constant of the Gaussian and we 
obtain the Laplace approximation 
\begin{equation}
  p(\params \mid \trainset)
  \approx
  \gaussianpdf{\params}{\paramsmap}{\mSigma_\traindata}
  \quad
  \text{with}
  \quad
  \mSigma_\traindata 
  \defeq (\numtraindata \cdot \hessian_\traindata)^{-1}
  = \frac{1}{\numtraindata} \hessian_\traindata^{-1}.
  \label{eq:laplace_approximation}
\end{equation}

\textbf{Mini-batch version of the Laplace approximation.} In order to reduce the
computational cost of the Laplace approximation, we can replace the full-batch quadratic
$\quadratic(\params; \traindata)$ in \cref{eq:laplace_fullbatch_quadratic} by a
mini-batch quadratic $\quadratic(\params; \minibatch)$, \ie we obtain
$\mSigma_\minibatch = \numtraindata^{-1} \hessian_\minibatch^{-1}$.

\textbf{Predictive uncertainty via the Laplace approximation.} Ultimately, we want to
use the Laplace approximation to equip the prediction for an unknown test input
$\vx_\testsymbol$ with uncertainty. Ideally, we would compute the expectation of the
model likelihood under the approximate posterior, \ie
\begin{align*}
  p(\vy_\testsymbol \mid \vx_\testsymbol, \trainset)
  &= 
  \int 
  p(\vy_\testsymbol \mid \vx_\testsymbol, \params) 
  \, p(\params \mid \trainset) 
  \, d\params\\
  \overset{\text{(\ref{eq:laplace_approximation})}}&{\approx}
  \int 
  \categoricalpdf{\vy_\testsymbol}{\softmax(\network_{\params}(\vx_\testsymbol))}
  \, \gaussianpdf{\params}{\paramsmap}{\mSigma_\minibatch} 
  \, d\params.
\end{align*}
One way to approximate this integral is via Monte Carlo sampling, \ie
\begin{equation}
  p(\vy_\testsymbol \mid \vx_\testsymbol, \trainset)
  \approx
  \frac{1}{S} \sum_{s=1}^S 
  \categoricalpdf{\vy_\testsymbol}{\softmax(\network_{\params^{(s)}}^{\text{lin}}(\vx_\testsymbol))}
  \quad \text{where} \quad
  \params^{(s)} \sim \gaussian{\paramsmap}{\mSigma_\minibatch}.
  \label{eq:laplace_monte_carlo}
\end{equation}
As suggested by \citet{Immer2021Improving,roy2024reparameterization}, we use the
linearized network
$
  \network_{\params}^{\text{lin}}(\vx) 
  \defeq 
  \network_{\paramsmap}(\vx)
  + \nabla \network_{\paramsmap}(\vx) (\params - \paramsmap)  
$, where $\nabla \network_{\paramsmap}(\vx) \in \R^{\numclasses \times \numparams}$ is 
the network's Jacobian at $\paramsmap$.

\subsubsection{Sampling from the \kfac Laplace approximation}
\label{sec:details_laplace_approximation_sampling}

The Monte Carlo (MC) approach from \Cref{eq:laplace_monte_carlo} requires samples from
the Gaussian $\gaussian{\paramsmap}{\mSigma_{\minibatch}}$ (the following derivations
work exactly the same when the full-batch LA based on the entire training set is used).
However, $\mSigma_{\minibatch} \in \R^{\numparams \times \numparams}$ is often too large
to be built explicitly in memory and it requires inverting the $\numparams \times
\numparams$ Hessian $\hessian_\minibatch$ of the regularized loss function. One approach
for that issue is to use the \kfac curvature approximation $\hessian_{\minibatch} =
\nabla^2 \Loss(\params; \minibatch) + \beta \mI \approx \mK_{\minibatch} + \beta \mI$,
\ie $\mSigma_\minibatch \approx \numtraindata^{-1} (\mK_{\minibatch} + \beta \mI)^{-1}$.
As we will see in the following, the \kfac approximation enables us to sample
efficiently from the corresponding LA.

\textbf{Leveraging \kfac's block-diagonal structure.} \kfac is a block-diagonal
curvature approximation, \ie $\mK_{\minibatch} = \blockdiag_{l = 1, \ldots,
L}(\mK_{\minibatch}^{(l)})$ and $\mK_{\minibatch} + \beta \mI$ inherits this structure.
The inverse of a block-diagonal matrix is also block-diagonal with the block inverses on
the diagonal,
\ie
\begin{align*}
  \mSigma_\minibatch
  &\approx 
  \frac{1}{\numtraindata} 
  (\mK_{\minibatch} + \beta \mI)^{-1}\\
  &=
  \frac{1}{\numtraindata}
  \big(
    \blockdiag_{l = 1, \ldots, L}(\mK_{\minibatch}^{(l)} + \beta \mI)
  \big)^{-1}\\
  &=
  \blockdiag_{l = 1, \ldots, L}\big(
    \underbracket[0.14ex]{      
    \frac{1}{\numtraindata}(\mK_{\minibatch}^{(l)} + \beta \mI)^{-1}
    }_{\displaystyle \eqdef \mSigma^{(l)}}
  \big) \\
  &=
  \blockdiag_{l = 1, \ldots, L}(\mSigma^{(l)}).
\end{align*}
To draw a sample $\vv \in \R^\numparams$ from
$\gaussian{\paramsmap}{\mSigma_\minibatch}$, we can thus simply sample from the block
covariances $\vv^{(l)} \sim \gaussian{\vzero}{\mSigma^{(l)}}$, stack these samples and
add the mean, \ie $\vv = \paramsmap + (\vv^{(1)}, \dots, \vv^{(L)})$.

\textbf{Leveraging the blocks' Kronecker structure.} To sample from
$\gaussian{\vzero}{\mSigma^{(l)}}$, we can exploit the Kronecker structure of the blocks
$\mK_{\minibatch}^{(l)} = \mA \otimes \mB$ (we omit the layer index $l$ for $\mA$ and
$\mB$ for brevity). First, we compute the eigendecompositions of the Kronecker factors,
\ie $\mA = \mU_{\mA} \mS_{\mA} \mU_{\mA}^\top$ and $\mB = \mU_{\mB} \mS_{\mB}
\mU_{\mB}^\top$. It follows
\begin{align*}
  \mK_{\minibatch}^{(l)} + \beta \mI
  &=
  \mA \otimes \mB + \beta \mI \\
  &=
  \mU_{\mA} \mS_{\mA} \mU_{\mA}^\top 
  \otimes 
  \mU_{\mB} \mS_{\mB} \mU_{\mB}^\top 
  + \beta \mI \\
  &=
  \underbracket[0.14ex]{(\mU_{\mA} \otimes \mU_{\mB})}_{\displaystyle \eqdef \mU}
  \underbracket[0.14ex]{(\mS_{\mA} \otimes \mS_{\mB})}_{\displaystyle \eqdef \mS}
  (\mU_{\mA} \otimes \mU_{\mB})^\top
  + \beta \mI\\
  &= 
  \mU \mS \mU^\top + \beta \mI\\
  &=
  \mU (\mS + \beta \mI) \mU^\top.
\end{align*}
$\mU = \mU_{\mA} \otimes \mU_{\mB}$ forms an orthogonal eigenbasis of the block and the
diagonal matrix $\mS + \beta \mI = \mS_{\mA} \otimes \mS_{\mB} + \beta \mI$ contains the
corresponding eigenvalues (see \eg \citep{George2018Fast}). It follows
\begin{align*}
  \mSigma^{(l)} 
  &= 
  \frac{1}{\numtraindata} (\mK_{\minibatch}^{(l)} + \beta \mI)^{-1} \\
  &= 
  \frac{1}{\numtraindata} \mU (\mS + \beta \mI)^{-1} \mU^\top\\
  &= 
  \underbracket[0.14ex]{
    \frac{1}{\sqrt{\numtraindata}} \mU (\mS + \beta \mI)^{-1/2}
  }_{
    \displaystyle \eqdef \mV}
  \frac{1}{\sqrt{\numtraindata}} (\mS + \beta \mI)^{-1/2} \mU^\top\\
  &= 
  \mV \mV^\top.
\end{align*}
So, in order to draw a sample $\vv^{(l)} \sim \gaussian{\vzero}{\mSigma^{(l)}}$, we
first draw from a standard Gaussian $\vw \sim \gaussian{\vzero}{\mI}$ and then transform
the sample via $\vv^{(l)} = \mV \vw$. The resulting vector $\vv^{(l)}$ has mean $\vzero$
and covariance $\mV \mI \mV^\top = \mSigma^{(l)}$. As Gaussians are closed under affine
linear transformations, $\vv^{(l)}$ is indeed distributed according to
$\gaussian{\vzero}{\mSigma^{(l)}}$.

The matrix-vector product $\vw \mapsto \mV \vw$ can be computed efficiently without
actually forming $\mV$ in memory. The first step is to multiply with $(\mS + \beta
\mI)^{-1/2}$. Since this is a diagonal matrix, we can simply multiply the vector $\vw$
element-wise with the inverse square root of the diagonal entries of $\mS + \beta \mI$.
The second step is to multiply with $\mU = \mU_{\mA} \otimes \mU_{\mB}$, which can be
implemented efficiently by using the property $(\mU_{\mA} \otimes \mU_{\mB}) \vect(\mW)
= \vect(\mU_{\mB} \mW \mU_{\mA}^\top)$. Finally, we scale by $\numtraindata^{-1/2}$.

\textbf{Summary.} Drawing a sample from the \kfac LA
$\gaussian{\paramsmap}{\numtraindata^{-1} (\mK_{\minibatch} + \beta \mI)^{-1}}$ can be
done without ever forming the blocks of the covariance matrix explicitly. Using
properties of the Kronecker product, we can efficiently transform a sample from the
standard Gaussian to a sample $\vv^{(l)} \sim \gaussian{\vzero}{\mSigma^{(l)}}$.
Stacking the samples from all blocks and adding the mean $\paramsmap$ yields a sample
from the LA due to the block-diagonal structure of the covariance matrix.

\subsubsection{Debiased \kfac Laplace approximation}
\label{sec:details_debiased_kfac}

The idea of the debiased \kfac Laplace approximation is to construct one
Kronecker-factored curvature matrix from two mini-batches $\minibatch$ and
$\otherminibatch$. 

\textbf{Eigenbasis of a \kfac block.} First, recall that the eigendecomposition of a
block $\mK_{\minibatch}^{(l)}$ from a \kfac matrix can be constructed from the
eigendecompositions of the Kronecker factors (see \eg \citep{George2018Fast}). Consider
the $l$-th block from both \kfac approximations $\mK_{\minibatch}^{(l)}\!=\!\mA \otimes
\mB$ and $\mK_{\otherminibatch}^{(l)}\!=\!\mC \otimes \mD$ (we omit the layer index $l$
for $\mA$, $\mB$, $\mC$, and $\mD$ for brevity). Again, let $\mA\!=\!\mU_{\mA} \mS_{\mA}
\mU_{\mA}^\top$ and $\mB\!=\!\mU_{\mB} \mS_{\mB} \mU_{\mB}^\top$ denote the
eigendecompositions of the Kronecker factors. It holds
\begin{align*}
  \mK_{\minibatch}^{(l)} 
  &= \mA \otimes \mB\\
  &= \mU_{\mA} \mS_{\mA} \mU_{\mA}^\top \otimes \mU_{\mB} \mS_{\mB} \mU_{\mB}^\top\\
  &= \underbracket[0.14ex]{(\mU_{\mA} \otimes \mU_{\mB})}_{\displaystyle \eqdef \mU}
  \underbracket[0.14ex]{(\mS_{\mA} \otimes \mS_{\mB})}_{\displaystyle \eqdef \mS}
  (\mU_{\mA} \otimes \mU_{\mB})^\top\\
  &= \mU \mS \, \mU^\top
\end{align*}
$\mU = \mU_{\mA} \otimes \mU_{\mB}$ forms an orthogonal eigenbasis of the block and the
diagonal matrix $\mS= \mS_{\mA} \otimes \mS_{\mB}$ contains the eigenvalues.

\textbf{Re-evaluation of the directional curvatures.} For the debiased approach, we keep
the block's eigenbasis $\mU$, but instead of using the directional curvatures $\mS$, we
re-evaluate these measurements on the second mini-batch $\otherminibatch$. First,
consider the projection of the block $\mK_{\otherminibatch}^{(l)}$ onto the eigenvectors
$\mU$, \ie
\begin{align*}
  \mU^\top \mK_{\otherminibatch}^{(l)} \mU
  =
  (\mU_{\mA} \otimes \mU_{\mB})^\top (\mC \otimes \mD) (\mU_{\mA} \otimes \mU_{\mB})
  =
  \mU_{\mA}^\top \mC \, \mU_{\mA} \otimes \mU_{\mB}^\top \mD \, \mU_{\mB}.
\end{align*}
The debiased directional curvatures are on the diagonal of $\mU^\top \mK_{\otherminibatch}^{(l)} \mU$.
For a square matrix $\mX \in \R^{n \times n}$, let $\Diag(\mX)$ denote the operator that maps the
matrix onto its diagonal, \ie $\Diag \, \colon \, \R^{n \times n} \to \R^n$ with
$\Diag(\mX)_i := X_{ii}$ for $i \in \{1, \dots, n\}$. It holds:
\begin{align*}
  \Diag(\mU^\top \mK_{\otherminibatch}^{(l)} \mU)
  &=
  \Diag(\mU_{\mA}^\top \mC \, \mU_{\mA} \otimes \mU_{\mB}^\top \mD \, \mU_{\mB}) \\
  &= 
  \underbracket[0.14ex]{
    \Diag(\mU_{\mA}^\top \mC \, \mU_{\mA})
  }_{\displaystyle \eqdef \tilde{\vs}_{\mA}}
  \otimes 
  \underbracket[0.14ex]{  
    \Diag(\mU_{\mB}^\top \mD \, \mU_{\mB})
  }_{\displaystyle \eqdef \tilde{\vs}_{\mB}} \\
  &=
  \tilde{\vs}_{\mA} \otimes \tilde{\vs}_{\mB} \\
  &\eqdef 
  \tilde{\vs}.
\end{align*}

\textbf{Construction of a debiased block.} Now that we have the eigenbasis $\mU$ and the
debiased directional curvatures $\tilde{\vs}$, we construct the debiased block. Let
$\tilde{\mS}_{\mA} \defeq \diag(\tilde{\vs}_{\mA})$, $\tilde{\mS}_{\mB} \defeq
\diag(\tilde{\vs}_{\mB})$ and 
\begin{equation*}
  \tilde{\mS} 
  \defeq \diag(\tilde{\vs})
  = \diag(\tilde{\vs}_{\mA} \otimes \tilde{\vs}_{\mB}) 
  = \diag(\tilde{\vs}_{\mA}) \otimes \diag(\tilde{\vs}_{\mB}) 
  = \tilde{\mS}_{\mA} \otimes \tilde{\mS}_{\mB}.
\end{equation*}
The debiased block is given by $\mU \tilde{\mS} \mU^\top$. It can be written as the
Kronecker product of two matrices $\tilde{\mA}$ and $\tilde{\mB}$ since 
\begin{align*}
  \mU \tilde{\mS} \mU^\top
  =
  (\mU_{\mA} \otimes \mU_{\mB}) 
  (\tilde{\mS}_{\mA} \otimes \tilde{\mS}_{\mB}) 
  (\mU_{\mA} \otimes \mU_{\mB})^\top
  =
  \underbracket[0.14ex]{
  (\mU_{\mA} \tilde{\mS}_{\mA} \mU_{\mA}^\top)
  }_{\displaystyle \eqdef \tilde{\mA}}
  \otimes 
  \underbracket[0.14ex]{
  (\mU_{\mB} \tilde{\mS}_{\mB} \mU_{\mB}^\top)
  }_{\displaystyle \eqdef \tilde{\mB}}.
\end{align*}
This is important as it allows for efficient sampling as described in
\Cref{sec:details_laplace_approximation_sampling}.

\textbf{Computational cost.} Since we need the eigendecompositions of the Kronecker
factors $\mA = \mU_{\mA} \mS_{\mA} \, \mU_{\mA}^\top$ and $\mB = \mU_{\mB} \mS_{\mB} \,
\mU_{\mB}^\top$ for sampling in any case (see
\cref{sec:details_laplace_approximation_sampling}), the computational overhead for the
debiased \kfac approximation $\tilde{\mA} \otimes \tilde{\mB}$ consists of computing a
\kfac approximation $\mC \otimes \mD$ on another mini-batch, re-evaluating the
directional curvatures $\tilde{\vs}_{\mA} = \Diag(\mU_{\mA}^\top \mC \, \mU_{\mA})$ and
$\tilde{\vs}_{\mB} = \Diag(\mU_{\mB}^\top \mD \, \mU_{\mB})$ and finally computing
$\tilde{\mA} = \mU_{\mA} \diag(\tilde{\vs}_{\mA}) \, \mU_{\mA}^\top$ and $\tilde{\mB} =
\mU_{\mB} \diag(\tilde{\vs}_{\mB}) \, \mU_{\mB}^\top$. 

\textbf{From block-level to full matrix.} So far, we have only considered the debiasing
of a single block. However, correcting the blocks's eigenvalues is sufficient because
they coincide with the eigenvalues of the full matrix (due to the block-diagonal
structure). Let $\vu^{(l)}$ denote an eigenvector of the $l$-th block $\mX^{(l)}$ of
some block-diagonal matrix $\mX = \blockdiag_{l=1, \dots, L}(\mX^{(l)})$ corresponding
to the eigenvalue $\eigval$. Then, $\vu^\top \defeq (\vzero^\top, \dots,
{\vu^{(l)}}^\top, \dots, \vzero^\top)$ is an eigenvector of $\mX$ corresponding to the
same eigenvalue $\eigval$, because 
\begin{equation*}
  \mX \cdot \vu 
  = 
  \blockdiag_{l = 1, \ldots, L}(\mX^{(l)}) \cdot \vu
  = 
  \begin{pmatrix}
    \mX^{(1)} \cdot \vzero\\
    \vdots\\
    \mX^{(l)} \cdot \vu^{(l)}\\
    \vdots\\
    \mX^{(L)} \cdot \vzero
  \end{pmatrix}
  =
  \begin{pmatrix}
    \vzero\\
    \vdots\\
    \eigval \, \vu^{(l)}\\
    \vdots\\
    \vzero
  \end{pmatrix}
  =
  \eigval \vu.
\end{equation*}
The eigenvalues of $\mX$ thus coincide with the eigenvalues of its blocks; and $\mX$'s
eigenvectors can be constructed from the the eigenvectors of the blocks by filling them
up with zeros. 

\textbf{Connection to \cref{eq:debiased_LA}:} The equivalence between
\cref{eq:debiased_LA} and the approach we describe above may not be obvious. We thus
show here that the directional curvature of the debiased matrix $\hat{\mK}$ along an
eigenvector $\vu$ of $\mK_{\minibatch}$ indeed coincides with the directional curvature
$\vu^\top \mK_{\otherminibatch} \, \vu$ on $\otherminibatch$.

More concretely, let $\hat{\mK} = \blockdiag_{l=1, \dots, L}(\hat{\mK}^{(l)})$ denote
the debiased \kfac approximation constructed from $\mK_{\minibatch}$ and
$\mK_{\otherminibatch}$ as described above. Also, let $\vu^\top \defeq (\vzero^\top,
\dots, {\vu^{(l)}}^\top, \dots, \vzero^\top)$ denote an eigenvector of
$\mK_{\minibatch}$, where $\vu^{(l)}$ is the $i$-th eigenvector of
$\mK_\minibatch^{(l)}$ (\ie the $i$-th column of $\mU$). 
It holds
\begin{equation*}
  \vu^\top \hat{\mK} \vu
  =
  {\vu^{(l)}}^\top \hat{\mK}^{(l)} \, \vu^{(l)}
  = 
  {\vu^{(l)}}^\top \mU \tilde{\mS} \, \mU^\top \, \vu^{(l)}
  = 
  \ve_i^\top \tilde{\mS} \, \ve_i
  = 
  \tilde{\vs}_i
  =
  {\vu^{(l)}}^\top \mK_{\otherminibatch}^{(l)} \, \vu^{(l)}
  =
  \vu^\top \mK_{\otherminibatch} \, \vu,
\end{equation*}
where $\ve_i$ denotes the $i$-th eigenvector.

\subsection{Bias in the directional slope}
\label{sec:details_bias_slope}

\textbf{Biased directional slopes along negative gradient directions.}
Assume that we are given a quadratic $\quadratic(\emptyarg; \minibatch)$ around the
current parameters $\params_0$. We consider the negative normalized gradient direction
$
  \vd
  = -\nabla \quadratic(\params_\bullet; \minibatch) \cdot
  \Vert \nabla \quadratic(\params_\bullet; \minibatch) \Vert^{-1}
$
at some location $\params_\bullet$ evaluated on mini-batch $\minibatch$. We have
\begin{equation}
  \underbracket[0.14ex]{
    \dslope{\vd} \, \quadratic(\params_\bullet; \otherminibatch)
  }_{\colordot{colorunbiased}}
  =
  \underbracket[0.14ex]{
    \dslope{\vd} \, \quadratic(\params_\bullet; \minibatch)
  }_{\colordot{colorbiased}}
  +
  \,\Vert \nabla \quadratic(\params_\bullet; \minibatch) \Vert (1 - \cos(\alpha))
  \geq
  \underbracket[0.14ex]{
    \dslope{\vd} \, \quadratic(\params_\bullet; \minibatch)
  }_{\colordot{colorbiased}},
  \label{eq:bias_slope_cg}
\end{equation}
where $\alpha \defeq \angle(\nabla \quadratic(\params_\bullet; \minibatch), \nabla
  \quadratic(\params_\bullet; \otherminibatch))$ and we assumed $\Vert\nabla
  \quadratic(\params_\bullet; \otherminibatch)\Vert = \Vert\nabla
  \quadratic(\params_\bullet; \minibatch)\Vert$ which is true at least in expectation.
Projecting a gradient onto its negative direction---the right-hand side of the
inequality---will \textit{always} result in a directional slope that is $\leq 0$ since
\begin{equation*}
  \dslope{\vd} \, \quadratic(\params_\bullet; \minibatch)
  = \vd^\top \nabla \quadratic(\params_\bullet; \minibatch)
  = - \frac{\nabla \quadratic(\params_\bullet; \minibatch)^\top}
           {\Vert \nabla \quadratic(\params_\bullet; \minibatch) \Vert}
    \nabla \quadratic(\params_\bullet; \minibatch)
  = - \Vert \nabla \quadratic(\params_\bullet; \minibatch) \Vert
  \leq 0
\end{equation*}
and $= 0$ only if $\nabla \quadratic(\params_\bullet; \minibatch) = \vzero$. Projecting
a \textit{different} gradient (of equal length) onto that direction---that is the
left-hand side of the inequality---will result in a larger (possibly even positive)
directional slope.

\textbf{Proof of \cref{eq:bias_slope_cg}.} \Cref{eq:bias_slope_cg} quantifies the bias
in the directional slope along $\vd$ as a function of the alignment of the two
mini-batch gradients. It holds
\begin{align*}
  \underbracket[0.14ex]{
    \dslope{\vd} \, \quadratic(\params_\bullet; \minibatch)
  }_{\colordot{colorbiased}}
  -
  \underbracket[0.14ex]{
    \dslope{\vd} \, \quadratic(\params_\bullet; \otherminibatch)
  }_{\colordot{colorunbiased}}
  &=
  \vd^\top \nabla \quadratic(\params_\bullet; \minibatch) 
  - \vd^\top \nabla \quadratic(\params_\bullet; \otherminibatch) \\[-2ex]
  &=
  \underbracket[0.140ex]{\Vert \vd \Vert}_{=1} \, 
  \Vert \nabla \quadratic(\params_\bullet; \minibatch) \Vert 
  \underbracket[0.140ex]{\cos(\pi)}_{=-1}
  -
  \underbracket[0.140ex]{\Vert \vd \Vert}_{=1} \, 
  \Vert \nabla \quadratic(\params_\bullet; \otherminibatch) \Vert
  \cos(\gamma)
  \\
  &= 
  \Vert \nabla \quadratic(\params_\bullet; \minibatch) \Vert (-1 - \cos(\gamma)),
\end{align*}
where $\gamma = \angle(\vd, \nabla \quadratic(\params_\bullet; \otherminibatch))$ and we
assumed that $\Vert \nabla \quadratic(\params_\bullet; \minibatch) \Vert = \Vert \nabla
\quadratic(\params_\bullet; \otherminibatch) \Vert$ in the last step (which is true, at
least, in expectation). Next, we re-write $\gamma$ as
\begin{equation*}
  \gamma 
  = \angle(\vd, \nabla \quadratic(\params_\bullet; \otherminibatch))
  = \angle(-\nabla \quadratic(\params_\bullet; \minibatch), 
            \nabla \quadratic(\params_\bullet; \otherminibatch))
  = \pi - \underbracket[0.140ex]{
    \angle(\nabla \quadratic(\params_\bullet; \minibatch), 
           \nabla \quadratic(\params_\bullet; \otherminibatch))
  }_{\eqdef \alpha}.
\end{equation*}
It follows 
$
  -1 - \cos(\gamma)
  =
  -1 - \cos(\pi - \alpha)
  =
  \cos(\alpha) - 1.
$
Substituting this into the expression for the bias, we arrive at
\begin{equation*}
  \underbracket[0.14ex]{
    \dslope{\vd} \, \quadratic(\params_\bullet; \minibatch)
  }_{\colordot{colorbiased}}
  -
  \underbracket[0.14ex]{
    \dslope{\vd} \, \quadratic(\params_\bullet; \otherminibatch)
  }_{\colordot{colorunbiased}}
  =
  \Vert \nabla \quadratic(\params_\bullet; \minibatch) \Vert (-1 - \cos(\gamma))
  =
  \Vert \nabla \quadratic(\params_\bullet; \minibatch) \Vert (\cos(\alpha) - 1),
\end{equation*}
from which \cref{eq:bias_slope_cg} follows. \qedsymbol

\textbf{The first \cg search direction.} Assume that the current parameters are
$\params_0$ and we apply \cg to the local quadratic approximation $\quadratic(\emptyarg;
\minibatch)$. The very first search direction is the quadratic's normalized negative
gradient $\vd_0  = -\nabla \quadratic(\params_0; \minibatch) \cdot \Vert \nabla
\quadratic(\params_0; \minibatch) \Vert^{-1}$ at $\params_0$. This is exactly the
situation we describe above, where $\params_\bullet$ is set to $\params_0$.
\Cref{eq:bias_slope_cg} thus explains the bias for the first \cg direction.

\textbf{The subsequent \cg search directions.} For the subsequent \cg search directions
($\vd_p$ with $p \geq 1$), the situation is more complex: Each direction $\vd_p$ is a
linear combination of the current normalized negative residual $-\vr_p$ (note that
$\vr_p$ coincides with the gradient $\nabla \quadratic(\params_p; \minibatch)$, see
\cref{sec:details_cg}) and the previous search direction $\vd_{p - 1}$ (see
\cref{sec:details_cg}) and the previous search direction $\vd_{p - 1}$ (see
\cref{alg:cg}), \ie
\begin{equation*}
  \vd_{p}
  = 
  \frac{\vs_p}{\Vert \vs_p \Vert}
  =
  \frac{1}{\Vert \vs_p \Vert} (- \vr_p)
  + 
  \frac{\beta_p}{\Vert \vs_p \Vert} \vs_{p-1}
  =
  \underbracket[0.14ex]{
    \frac{\Vert \vr_p \Vert}{\Vert \vs_p \Vert}
  }_{\eqdef \eta_1 \geq 0} 
  \frac{- \vr_p}{\Vert \vr_p \Vert}
  +
  \underbracket[0.14ex]{
    \frac{\beta_p \Vert \vs_{p-1} \Vert}{\Vert \vs_p \Vert} 
  }_{\eqdef \eta_2 \geq 0} 
  \vd_{p - 1}
  =
  \eta_1 \frac{- \vr_p}{\Vert \vr_p \Vert} + \eta_2\, \vd_{p - 1}.
\end{equation*}
The additional correction term ensures conjugacy. The
directional slope along $\vd_p$ thus also splits into two corresponding terms: The slope
along the negative gradient direction and the slope along the previous search direction.
It holds
\begin{align*}
  \dslope{\vd_p} \quadratic(\params_p; \minibatch)
  &=
  \vd_p^\top \nabla \quadratic(\params_p; \minibatch)\\
  &=
  \eta_1 \frac{- \vr_p}{\Vert \vr_p \Vert}^\top \nabla \quadratic(\params_p; \minibatch)
  +
  \eta_2 \, \vd_{p - 1}^\top \nabla \quadratic(\params_p; \minibatch)\\
  &=
  \eta_1 
  \dslope{\nicefrac{- \vr_p}{\Vert \vr_p \Vert}} \quadratic(\params_p; \minibatch)
  +
  \eta_2 \, \dslope{\vd_{p - 1}} \quadratic(\params_p; \minibatch).
\end{align*}
The bias in the first term---as explained by \cref{eq:bias_slope_cg}---also introduces a
bias along the search direction $\vd_p$.

\subsection{Bias in the directional curvature}
\label{sec:details_bias_curvature}

\textbf{Derivation for \cref{eq:bias_curvature_eigvecs}.}
Let $\hessian_\minibatch = \eigvecs \eigvals \eigvecs^\top$ and
$\hessian_{\otherminibatch} = \tilde{\eigvecs} \tilde{\eigvals} \tilde{\eigvecs}^\top$
denote the eigendecompositions of the mini-batch Hessians, where $\eigvecs = (\eigvec_1,
\ldots, \eigvec_{\numparams}), \tilde{\eigvecs} = (\tilde{\eigvec}_1, \ldots,
\tilde{\eigvec}_{\numparams}) \in \R^{\numparams \times \numparams}$ contain the
orthonormal eigenvectors and $\eigvals = \diag(\eigval_1, \ldots,
\eigval_{\numparams})$, $\tilde{\eigvals} = \diag(\tilde{\eigval}_1, \ldots,
\tilde{\eigval}_{\numparams})$ the respective eigenvalues in descending order, \ie
$\eigval_1 \geq \ldots \geq \eigval_{\numparams}$ and $\tilde{\eigval}_1 \geq \ldots
\geq \tilde{\eigval}_{\numparams}$.
The directional curvature along one of $\hessian_\minibatch$'s eigenvectors $\eigvec_i$
on mini-batch $\minibatch$ is given by the corresponding eigenvalue since
\begin{equation*}
  \underbracket[0.14ex]{
    \dcurv{\eigvec_i} \quadratic(\params_\bullet; \minibatch)
  }_{\colordot{colorbiased}}
  = 
  \eigvec_i^\top \hessian_{\minibatch} \eigvec_i
  = 
  \eigvec_i^\top \eigval_i \eigvec_i
  =
  \eigval_i \Vert \eigvec_i \Vert_2^2
  = 
  \eigval_i.
\end{equation*}
For $\otherminibatch$, we obtain
\begin{equation*}
  \underbracket[0.14ex]{
    \dcurv{\eigvec_i} \quadratic(\params_\bullet; \otherminibatch)
  }_{\colordot{colorunbiased}}
  = 
  \eigvec_i^\top \hessian_{\otherminibatch} \eigvec_i
  = 
  \eigvec_i^\top
  \left( 
    \sum_{p=1}^\numparams \tilde{\eigval}_p \tilde{\eigvec}_p \tilde{\eigvec}_p^\top
  \right)
  \eigvec_i
  = 
  \sum_{p=1}^\numparams
  \tilde{\eigval}_p 
  \underbracket[0.14ex]{
    (\tilde{\eigvec}_p^\top \eigvec_i)^2
  }_{\eqdef \mOmega_{i, p}}
  = 
  \sum_{p=1}^\numparams
  \tilde{\eigval}_p \, \mOmega_{i, p}.
\end{equation*}

\textbf{Weights sum up to one.} The weights $\mOmega_{i, p}$ are non-negative and sum up
to one, \ie $\sum_{p=1}^\numparams \mOmega_{i, p} = 1$. This is because
\begin{equation}
  \sum_{p=1}^\numparams \mOmega_{i, p} 
  = 
  \sum_{p=1}^\numparams (\tilde{\eigvec}_p^\top \eigvec_i)^2
  =
  \Vert \tilde{\eigvecs}^\top \eigvec_i \Vert_2^2
  = 
  \eigvec_i^\top 
  \underbracket[0.14ex]{\tilde{\eigvecs} \tilde{\eigvecs}^\top}_{= \mI} 
  \eigvec_i
  =
  \Vert \eigvec_i \Vert_2^2
  =
  1
  \label{eq:weights_sum_to_one}
\end{equation}
where we used that $\tilde{\eigvecs}$ is an orthogonal matrix and that $\eigvec_i$ is
normalized.

\textbf{Proof of \cref{eq:bias_curvature_u1_uP}.} Here, we assume the simplified case
where the spectra of $\hessian_\minibatch$ and $\hessian_{\otherminibatch}$ are
identical, \ie $\eigval_p = \tilde{\eigval}_p \, \forall p \in \{1, \ldots,
\numparams\}$. 

First, consider $\eigvec_1$ and assume that $\eigvec_1$ and $\tilde{\eigvec}_1$ are
\textit{not} perfectly aligned, \ie $\mOmega_{1, 1} = (\eigvec_1^\top
\tilde{\eigvec}_1)^2 < 1$. It holds
\begin{equation*}
  \underbracket[0.14ex]{
    \dcurv{\eigvec_1} \quadratic(\params_0, \otherminibatch)
  }_{\colordot{colorunbiased}}
\overset{\text{(\ref{eq:bias_curvature_eigvecs})}}{=}
  \sum_{p=1}^\numparams \eigval_p \mOmega_{1, p}
  =
  \eigval_1 \mOmega_{1, 1}
  +
  \sum_{p=2}^\numparams \eigval_p \mOmega_{1, p}.
\end{equation*}
The sum can be bounded from above by putting all remaining weight $1 - \mOmega_{1, 1}$
(see \cref{eq:weights_sum_to_one}) on the second-largest eigenvalue $\eigval_2$, \ie 
\begin{equation*}
  \underbracket[0.14ex]{
    \dcurv{\eigvec_1} \quadratic(\params_0, \otherminibatch)
  }_{\colordot{colorunbiased}}
  \leq
  \eigval_1 \mOmega_{1, 1} + \eigval_2 (1 - \mOmega_{1, 1})
  \overset{(*)}{\leq}
  \eigval_1 \mOmega_{1, 1} + \eigval_1 (1 - \mOmega_{1, 1})
  = 
  \eigval_1
  =
  \underbracket[0.14ex]{
    \dcurv{\eigvec_1} \quadratic(\params_0, \minibatch)
  }_{\colordot{colorbiased}}\!.
\end{equation*}
The inequality $(*)$ turns into $<$ if the top two eigenvalues are separated, \ie
$\eigval_1 > \eigval_2$.
The proof for $\eigvec_\numparams$ is similar. We assume that $\eigvec_\numparams$ and
$\tilde{\eigvec}_\numparams$ are \textit{not} perfectly aligned, \ie
$\mOmega_{\numparams, \numparams} = (\eigvec_P^\top \tilde{\eigvec}_\numparams)^2 < 1$.
It follows 
\begin{equation*}
  \underbracket[0.14ex]{
    \dcurv{\eigvec_\numparams} \quadratic(\params_0, \otherminibatch)
  }_{\colordot{colorunbiased}}
\overset{\text{(\ref{eq:bias_curvature_eigvecs})}}{=}
  \sum_{p=1}^\numparams \eigval_p \mOmega_{\numparams, p}
  =
  \sum_{p=1}^{\numparams - 1} \eigval_p \mOmega_{\numparams, p}
  + 
  \eigval_\numparams \mOmega_{\numparams, \numparams}.
\end{equation*}
This time, we bound the sum from below by putting all remaining weight $1 - \mOmega_{\numparams, \numparams}$ (see \cref{eq:weights_sum_to_one}) on the second-smallest eigenvalue $\eigval_{\numparams - 1}$, \ie
\begin{equation*}
  \underbracket[0.14ex]{
    \dcurv{\eigvec_\numparams} \quadratic(\params_0, \otherminibatch)
  }_{\colordot{colorunbiased}}
  \geq
  \eigval_{\numparams - 1} (1 - \mOmega_{\numparams, \numparams})
  + 
  \eigval_\numparams \mOmega_{\numparams, \numparams}
  \overset{(*)}{\geq}
  \eigval_\numparams (1 - \mOmega_{\numparams, \numparams})
  + 
  \eigval_\numparams \mOmega_{\numparams, \numparams}
  = 
  \eigval_\numparams
  =
  \underbracket[0.14ex]{
    \dcurv{\eigvec_\numparams} \quadratic(\params_0, \minibatch)
  }_{\colordot{colorbiased}}\!.
\end{equation*}
Again, the inequality $(*)$ turns into $>$ if the bottom two eigenvalues are separated,
\ie $\eigval_{\numparams - 1} > \eigval_\numparams$. This concludes the proof of
\cref{eq:bias_curvature_u1_uP}. \qedsymbol

\section{Experimental details}
\label{sec:experimental_details}

In the following, we provide information on the experimental setup and give detailed
instructions on how to replicate all empirical results.

\subsection{Test problems and training procedures}
\label{sec:exp_details_training}

Throughout the paper, we use a series of test problems with different models, \datasets,
and training procedures which we describe in more detail in the following. We use
\deepobs \citep{schneider2019deepobs} on top of \pytorch \citep{Paszke2019PyTorch} as
our general benchmarking framework as it provides easy access to a variety of \datasets
and model architectures.

\textbf{Data.} We use the \datasets \cifarten (with $\numclasses = 10$ classes) and
\cifarhun (with $\numclasses = 100$ classes) \citep{krizhevsky2009learning}. Each
\dataset contains \num{60000} data points that are split into \num{40000} training
samples, \num{10000} validation samples and \num{10000} test samples. For the
experiments on out-of-distribution (OOD) data, we create the \datasets \cifartenc and
\cifarhunc, each containing \num{10000} images, as described in
\citep{hendrycks2018benchmarking}. In these \datasets, each image is corrupted using one
out of $15$ different corruptions (chosen uniformly at random) at a specific severity
level (a number between $1$ and $5$). We also use the \imagenet \dataset
\citep{Deng2009ImageNet} which contains images from $\numclasses = 1000$ different
classes. 

\begin{figure}[p]
  \centering

  \begin{minipage}[c]{0.05\textwidth}
    \ref{tp:allcnnc_cifar100}
  \end{minipage}\hfill
  \begin{minipage}[c]{0.93\textwidth}
    \includegraphics[scale=1]{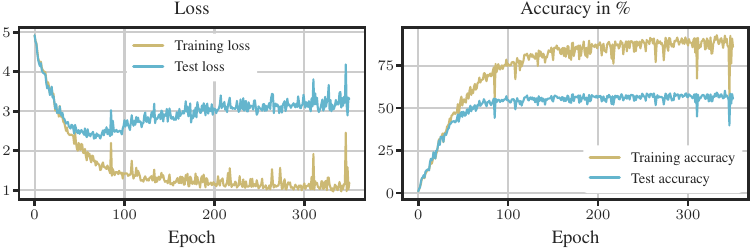}
  \end{minipage}

  \vspace{2ex}

  \begin{minipage}[c]{0.05\textwidth}
    \ref{tp:allcnnc_cifar10}
  \end{minipage}\hfill
  \begin{minipage}[c]{0.93\textwidth}
    \includegraphics[scale=1]{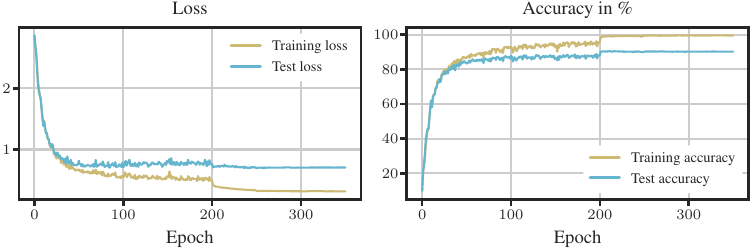}
  \end{minipage}

  \vspace{2ex}

  \begin{minipage}[c]{0.05\textwidth}
    \ref{tp:wideresnet_cifar100}
  \end{minipage}\hfill
  \begin{minipage}[c]{0.93\textwidth}
    \includegraphics[scale=1]{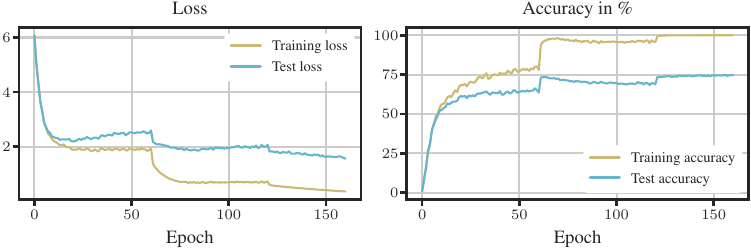}
  \end{minipage}

  \vspace{2ex}

  \begin{minipage}[c]{0.05\textwidth}
    \vspace{-8ex}
    \ref{tp:convnet_cifar10}
  \end{minipage}\hfill
  \begin{minipage}[c]{0.93\textwidth}
    \includegraphics[scale=1]{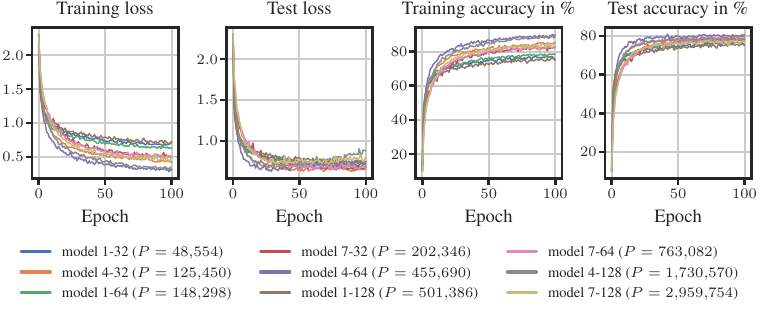}
  \end{minipage}

  \caption{\textbf{Training metrics for all test problems.} The panels show the
  training/test loss/accuracy during training for all test problems
  \ref{tp:allcnnc_cifar100} (\allcnnc on \cifarhun), \ref{tp:allcnnc_cifar10} (\allcnnc
  on \cifarten), \ref{tp:wideresnet_cifar100} (\wrn on \cifarhun) and
  \ref{tp:convnet_cifar10} (\convnet on \cifarten). Test problem
  \ref{tp:resnet50_imagenet} (\resnetfifty on \imagenet) is not shown since a
  pre-trained model is used here.}
  \label{fig:training_metrics}
\end{figure}

\textbf{Test problems.} All of the following test problems use the cross-entropy loss
function.
\begin{enumerate}[label={(\Alph*)}]
  \item \label{tp:allcnnc_cifar100} \textbf{\allcnnc on \cifarhun.} This test problem
  uses the \allcnnc model architecture \citep{springenberg2015striving} (where we
  removed the dropout layers as explained in \cref{sec:empirical_study_dd}) and
  \cifarhun data. The training hyperparameters are taken from an existing benchmark
  \citep{schmidt2021descending}: We train the model with \sgd (learning rate $0.171234$)
  with batch size $256$ for $350$ epochs. Weight decay $\beta = 0.0005$ is used on the
  weights but not the biases of the model.

  \item \label{tp:allcnnc_cifar10} \textbf{\allcnnc on \cifarten.} This test problem is
  similar to \ref{tp:allcnnc_cifar100} but uses the \cifarten \dataset. The model is
  trained with \sgd (learning rate $0.025$, momentum $0.9$) with batch size $256$ for
  $350$ epochs. The learning rate is reduced by a factor of $10$ at epochs $200$, $250$
  and $300$, as suggested in \citep{springenberg2015striving}. Weight decay $\beta =
  0.001$ is used on the weights but not the biases of the model.

  \item \label{tp:wideresnet_cifar100} \textbf{\wrn on \cifarhun.} This is a test problem
  from \deepobs. For details on the architecture, see \citep{zagoruyko2017wide}. We
  train this model with \sgd (learning rate $0.1$, momentum $0.9$) with batch size $128$
  for $160$ epochs. The learning rate is reduced by a factor of $5$ after $60$ and $120$
  epochs. Weight decay $\beta = 0.0005$ is used on the non-bias weights of the model.

  \item \label{tp:convnet_cifar10} \textbf{\convnet on \cifarten.} This test problem uses
  a simple convolutional neural network with variable depth $d$ and width $w$ (denoted
  by ``model $d$-$w$'' in \cref{fig:training_metrics,fig:bias_numparams}). The
  \textit{first} block of the model consists of a convolutional layer (kernel size $5$,
  padding $2$) with $3$ input and $w$ output channels, a ReLU activation function, and a
  max-pooling layer (kernel size $2$). The \textit{last} block consists of a max-pooling
  layer (kernel size $2$), a flatten layer, and a dense linear layer. In between those
  blocks, there are $d$ \textit{hidden} blocks each consisting of a convolutional layer
  (kernel size $5$, padding $2$) with $w$ input and $w$ output channels followed by a
  ReLU activation function. So, the depth $d$ determines the number of hidden blocks,
  whereas $w$ controls the number of parameters in the layers and is thus referred to as
  the width.
We use $d \in \{1, 4, 7\}$ and $w \in \{32, 64, 128\}$ (resulting in $9$ models) and
  train each model for $100$ epochs on the \cifarten \dataset using \adam with standard
  hyperparameters (learning rate $0.001$, $\beta_1 = 0.9$, $\beta_2 = 0.999$, $\epsilon
  = 10^{-8}$) with batch size $256$. No weight decay is used for this test problem. 

  \item \label{tp:resnet50_imagenet} \textbf{\resnetfifty on \imagenet.} This
  test problem uses the \resnetfifty model architecture \citep{He2016ResNet} and the
  \imagenet \dataset. We use the \href{https://download.pytorch.org/models/resnet50-19c8e357.pth}{\textsc{IMAGENET1K\_V1}}
  weights pre-trained on \imagenetthou using \sgd (initial learning rate $0.1$, momentum $0.9$)
  with batch size $256$ for $90$ epochs. For the pre-training, the learning rate was set by a multistep scheduler
  which multiplied the learning rate by a factor of $0.1$ after
  every $30$ epochs. Additionally, a weight decay of $\beta = 0.0001$ was used on all weights apart from
  biases and the learned \batchnorm weights.

  \item \label{tp:vit_little_imagenet} \textbf{\vitlittle on \imagenet.} This
  test problem uses the \vitlittle architecture of the \vitfull model family \citep{dosovitskiy2021an} on the
  \imagenet \dataset. We use the \href{https://huggingface.co/timm/vit_little_patch16_reg4_gap_256.sbb_in1k}{\textsc{vit\_little\_patch16\_reg4\_gap\_256.sbb\_in1k}}
  weights pre-trained on \imagenetthou using \nadamw. For the exact hyperparameters used
  for the weights, refer to the \href{https://huggingface.co/timm/vit_little_patch16_reg4_gap_256.sbb_in1k/blob/main/train_hparams.yaml}{HuggingFace Model Card}.
\end{enumerate}

During training, we store the model's parameters at $10$ checkpoints spaced
log-equidistantly between the first and last epoch. The training metrics for all
test problems are shown in \cref{fig:training_metrics}.

\subsection{Matrix-vector products with the curvature matrix}
\label{sec:exp_details_curvature_matrices}

\textbf{Access to curvature matrices via \backpack.} In order to
compute an
eigendecomposition of the curvature matrix or apply the \cg method to a quadratic, we
need access to matrix-vector products with the curvature matrix $\vv \mapsto
\hessian_{\minibatch} \cdot \vv$. When products with the full-batch curvature matrix are
required, we accumulate the mini-batch quantities over manageable chunks of the
training set.
Our implementation uses \backpack \citep{dangel2020backpack} that provides access to
products with the Hessian $\vv \mapsto \nabla^2 \Loss(\params; \minibatch) \cdot \vv$
and GGN $\vv \mapsto \GGN_\minibatch \cdot \vv$ of the empirical risk $\Loss$ as well as
the \kfac curvature approximation $\mK_\minibatch$. 

\textbf{Eigenvectors of the curvature matrix.} Given access to matrix-vector products
with the curvature matrix, we can construct an instance of a
\texttt{scipy.sparse.linalg.LinearOperator} that can be used in
\texttt{scipy.sparse.linalg.eigsh}.

\subsection{\cref{sec:introduction} (\nameref{sec:introduction}): \cref{fig:visual_abstract}}
\label{sec:exp_details_visual_abstract}

For the visual abstract in \cref{fig:visual_abstract}, we use the fully trained \allcnnc
model from test problem \ref{tp:allcnnc_cifar100} (see \cref{sec:exp_details_training}).
The experimental procedure below is repeated for $5$ different mini-batches
$\minibatch_m$, $m \in \{0, 1, 2, 3, 4\}$ of size $\vert\minibatch_m\vert = 512$ (we
omit the index $m$ in the following).

\textbf{Experimental procedure.} We use the GGN curvature proxy $\hessian_\minibatch
\gets \GGN_\minibatch + \beta \mI$ and compute its top two eigenvectors $\eigvec_1$ and
$\eigvec_2$ (see \cref{sec:exp_details_curvature_matrices}). In order to evaluate the quadratic $\quadratic(\paramsmap; \minibatch)$ in the 2D space
spanned by those eigenvectors efficiently, we use the following equation:
\begin{align*}
  \quadratic(\paramsmap + \tau_1 \eigvec_1 + \tau_2 \eigvec_2; \minibatch) 
  &= \frac{1}{2} \tau_1^2 [\eigvec_1^\top \hessian_\minibatch\, \eigvec_1]
  + \tau_1 \tau_2 [\eigvec_1^\top \hessian_\minibatch\, \eigvec_2]
  + \frac{1}{2} \tau_2^2 [\eigvec_2^\top \hessian_\minibatch\, \eigvec_2]\\
  &\;\;\;\;+ \tau_1 [\eigvec_1^\top \grad_\minibatch]
  + \tau_2 [\eigvec_2^\top \grad_\minibatch]
  + [\const_\minibatch].
\end{align*}
We can compute all the terms on the right-hand side in brackets \textit{once}, store
them, and then evaluate the quadratic for arbitrary values of $\tau_1$ and $\tau_2$ at
basically no cost. The derivation for the full-batch quadratic $\quadratic(\paramsmap +
\tau_1 \eigvec_1 + \tau_2 \eigvec_2; \traindata)$ is analogous.

\subsection{\cref{sec:geometry_minibatch_quadratic} (\nameref{sec:geometry_minibatch_quadratic}): \cref{fig:bias,fig:overlaps,fig:cg_magnitudes}}
\label{sec:exp_details_biases}

For the evaluation of the directional derivatives, we use the fully trained \allcnnc
model from test problem \ref{tp:allcnnc_cifar100} (see \cref{sec:exp_details_training}). 

\textbf{Three settings.} Throughout this section, we consider three
different settings: (i)~Quadratics that use the Hessian of the empirical risk, \ie
$\hessian_\minibatch \gets \nabla^2 \Loss(\paramsmap; \minibatch) + \beta \mI$ at batch
size $512$, (ii)~quadratics that use the Hessian's GGN approximation, \ie
$\hessian_\minibatch \gets \GGN_{\minibatch} + \beta \mI$ at batch size $512$ and
(iii)~at batch size $2048$. For each setting, the experimental procedure below is
repeated for $3$ mini-batches $\minibatch_m$, $m \in \{0, 1, 2\}$. The prior precision
$\beta$ is set to $0.0005$ (the same $\beta$ was used for training, see
\cref{sec:exp_details_training}) and only acts on the weights of the model but not its
biases.

\textbf{Experimental procedure (biases).} The experimental procedure consists of two
steps: The computations of the directions on mini-batch $\minibatch_m$ and the
evaluation of the directional derivatives on \textit{all} mini-batches (of the same
size) in the training \dataset.
\begin{enumerate}
\item \textbf{Directions based on $\minibatch_m$.} First, we use the quadratic
  $\quadratic(\paramsmap; \minibatch_m)$ with the given curvature proxy and batch size
  to compute a set of $100$ directions. This is either the top $100$ eigenvectors of
  $\hessian_\minibatch$ computed via \texttt{scipy.sparse.linalg.eigsh} (see
  \cref{sec:exp_details_curvature_matrices}) or the first $100$ \cg search directions
  (see \cref{sec:exp_details_cg}). In the case of \cg, we also store the trajectory of
  the iterates $\paramsmap = \params_0, \params_1, \ldots, \params_{100}$.

\item \textbf{Directional derivatives on all mini-batches.} Finally, we evaluate the
  directional slope and curvature for all directions from step 1 on \textit{all}
  mini-batches $\minibatch_{m'}$, $m' \in \{0, \dots, \numminibatches - 1\}$ in the
  training \dataset (where $\numminibatches$ denotes the number of mini-batches in the
  training data). 
  \begin{itemize}
    \item \textbf{Directional derivatives along eigenvectors.} For an eigenvector
    $\eigvec$ the directional slope and curvature are given by $\dslope{\eigvec} \,
    \quadratic(\paramsmap; \minibatch_{m'})$ and $\dcurv{\eigvec} \,
    \quadratic(\paramsmap; \minibatch_{m'})$, respectively.
\item \textbf{Directional derivatives along \cg search directions.} For \cg, we
    evaluate the directional slope and curvature along each search direction $\vd_p$ at
    the respective iterate $\params_p$ of the trajectory,     
    \ie $\dslope{\vd_p} \, \quadratic(\params_p; \minibatch_{m'})$ for the slope and
    $\dcurv{\vd_p} \, \quadratic(\params_p; \minibatch_{m'})$ for the curvature. As the
    curvature $\nabla^2 \quadratic(\params; \minibatch_{m'}) \equiv
    \hessian_{\minibatch_{m'}}$ is independent of $\params$, this small distinction is
    irrelevant for the directional curvature. For the directional slope, however, it
    does make a difference.
  \end{itemize}
\end{enumerate}

\begin{figure}[p]
  \centering
  \textbf{Biases: Additional results for $\hessian_\minibatch \gets \nabla^2 \Loss(\paramsmap;
    \minibatch) + \beta \mI$ (batch size $512$)}

  \vspace{4ex}

  Eigenvectors computed on mini-batch\\[0.25ex]
  \includegraphics[scale=0.99]{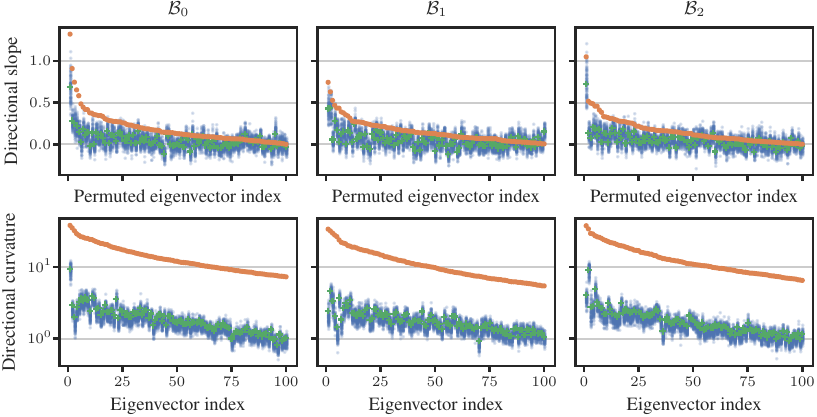}

  \vspace{4ex}

  \cg trajectory and search directions computed on mini-batch\\[0.25ex]
  \includegraphics[scale=0.99]{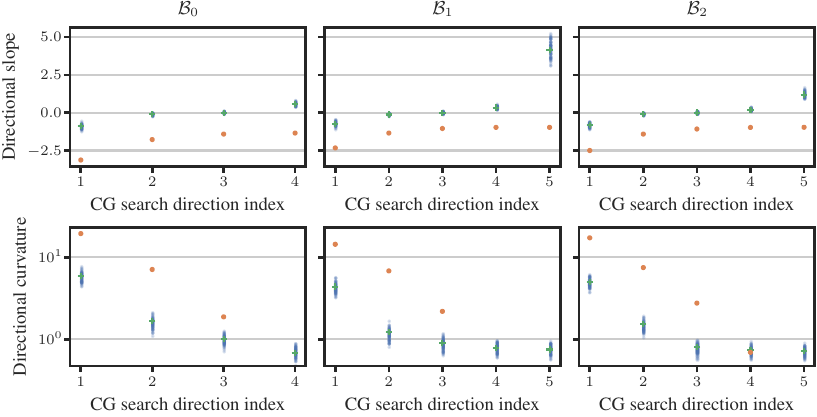}

  \caption{\textbf{Directional slopes and curvatures are biased.}
The experimental setting is similar to \cref{fig:bias} but uses the Hessian of the
    empirical risk, \ie $\hessian_\minibatch \gets \nabla^2 \Loss(\paramsmap;
    \minibatch) + \beta \mI$ at batch size $512$.
The upper plot \figtop shows the directional derivatives along the top $100$
    eigenvectors of $\hessian_\minibatch$, the lower plot \figbottom shows the
    directional derivatives along the first $100$ \cg search directions.
For the top panel of the upper plot, we switch the order and sign of the
    eigenvectors such that the orange dots are all above zero and in descending order.
There are strong systematic biases in the directional slopes and curvatures. The
    curvature biases are more pronounced along the eigenvectors, whereas the biases in
    the slope are larger along the \cg directions. }
  \label{fig:bias_hessian_512}
\end{figure}

\begin{figure}[p]
  \centering
  \textbf{Biases: Additional results for $\hessian_\minibatch \gets \GGN_\minibatch + \beta \mI$ (batch size $512$)}

  \vspace{4ex}

  Eigenvectors computed on mini-batch\\[0.25ex]
  \includegraphics[scale=0.99]{exp09_dd1_and_dd2/cifar100_allcnnc_wo_dropout_SGD_curv_opt_ggn_batch_size_512_direction_opt_eigvecs_dtype_torch.float32_ylog.pdf}

  \vspace{4ex}

  \cg trajectory and search directions computed on mini-batch\\[0.25ex]
  \includegraphics[scale=0.99]{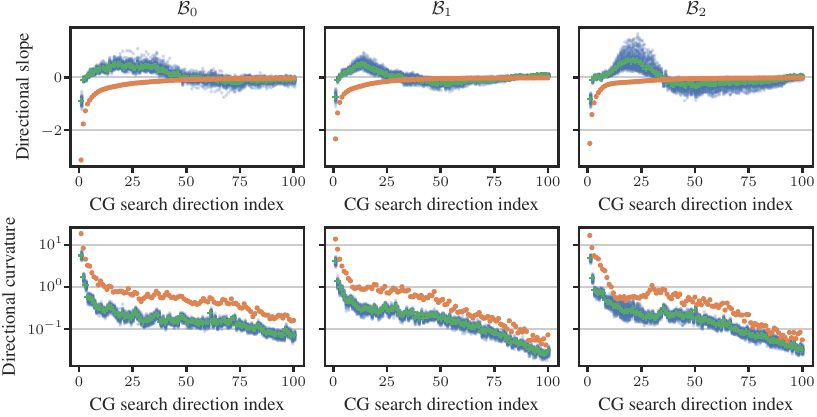}

  \caption{\textbf{Directional slopes and curvatures are biased.}
The experimental setting is the same as in \cref{fig:bias}: We use the GGN curvature
    proxy $\hessian_\minibatch \gets \GGN_\minibatch + \beta \mI$ at batch size $512$.
The upper plot \figtop shows the directional derivatives along the top $100$
    eigenvectors of $\hessian_\minibatch$, the lower plot \figbottom shows the
    directional derivatives along the first $100$ \cg search directions.
For the top panel of the upper plot, we switch the order and sign of the
    eigenvectors such that the orange dots are all above zero and in descending order.
There are strong systematic biases in the directional slopes and curvatures. The
    curvature biases are more pronounced along the eigenvectors, whereas the biases in
    the slope are larger along the \cg directions. }
  \label{fig:bias_ggn_512}
\end{figure}

\begin{figure}[p]
  \centering
  \textbf{Biases: Additional results for $\hessian_\minibatch \gets \GGN_\minibatch + \beta \mI$ (batch size $2048$)}

  \vspace{4ex}

  Eigenvectors computed on mini-batch\\[0.25ex]
  \includegraphics[scale=0.99]{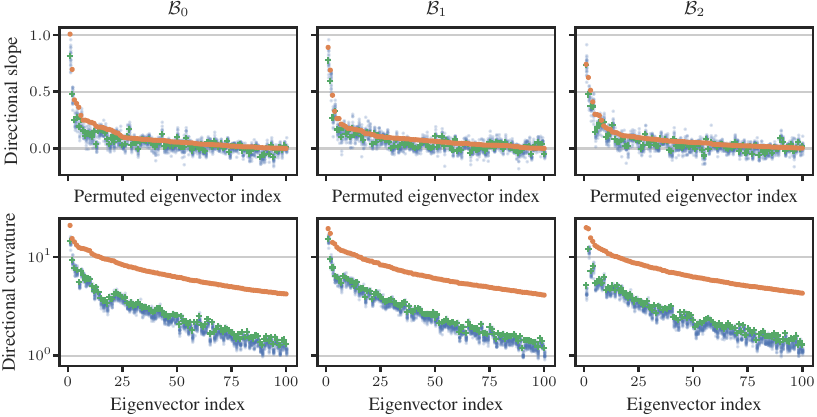}

  \vspace{4ex}

  \cg trajectory and search directions computed on mini-batch\\[0.25ex]
  \includegraphics[scale=0.99]{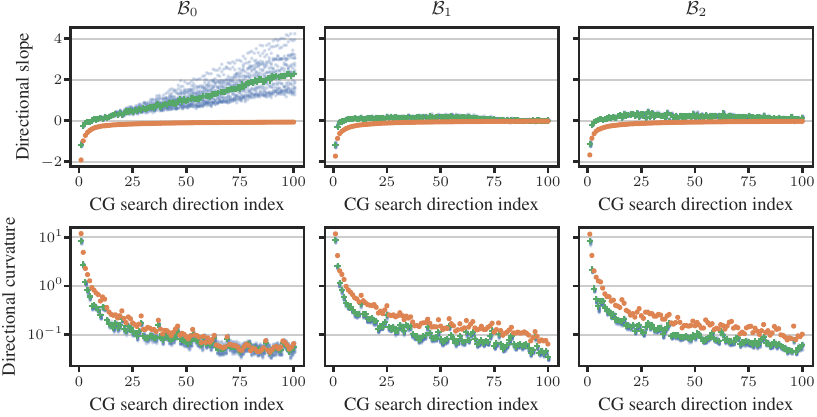}

  \caption{\textbf{Directional slopes and curvatures are biased.}
The experimental setting is similar to \cref{fig:bias}: We use the GGN curvature
    proxy $\hessian_\minibatch \gets \GGN_\minibatch + \beta \mI$ but batch size $2048$.
The upper plot \figtop shows the directional derivatives along the top $100$
    eigenvectors of $\hessian_\minibatch$, the lower plot \figbottom shows the
    directional derivatives along the first $100$ \cg search directions.
For the top panel of the upper plot, we switch the order and sign of the
    eigenvectors such that the orange dots are all above zero and in descending order.
There are strong systematic biases in the directional slopes and curvatures. The
    curvature biases are more pronounced along the eigenvectors, whereas the biases in
    the slope are larger along the \cg directions.}
  \label{fig:bias_ggn_2048}
\end{figure}

\textbf{Results (biases).} The results for setting (i) are shown in
\cref{fig:bias_hessian_512}, for setting (ii) in \cref{fig:bias,fig:bias_ggn_512} and
for (iii) in \cref{fig:bias_ggn_2048}. 
The biases in the directional slopes and curvatures can be observed across all
scenarios. The biases in the slopes are even
more pronounced along the \cg search directions
than along the eigenvectors of the curvature matrix. The opposite holds for the
curvature biases that tend to be larger along the eigenvectors. This is consistent with
our explanations from \cref{sec:biases_theory}.
Both the biases in the slope and curvature decrease with increasing mini-batch size. For
setting (i), \cg encounters a search direction with negative curvature (indicating that,
as expected, the curvature matrix $\hessian_\minibatch$ is indefinite) and thus
terminates already after $4$ iterations.

\begin{figure}[p]
    \centering
    \textbf{Eigenspace overlaps: Additional result for $\hessian_\minibatch \gets
    \nabla^2 \Loss(\paramsmap; \minibatch) + \beta \mI$}

    \vspace{3ex}

    \begin{minipage}{0.47\textwidth}
      \centering
      \textbf{Batch size $512$}\\[2ex]
      \includegraphics[scale=0.95]{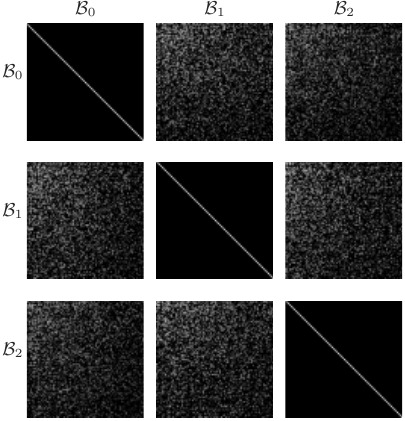}
    \end{minipage}
    \hfill
    \textcolor{white}{.}

    \vspace{5ex}

    \textbf{Eigenspace overlaps: Additional results for $\hessian_\minibatch \gets \GGN_\minibatch + \beta \mI$}

    \vspace{3ex}

    \begin{minipage}{0.47\textwidth}
      \centering
      \textbf{Batch size $512$}\\[2ex]
      \includegraphics[scale=0.95]{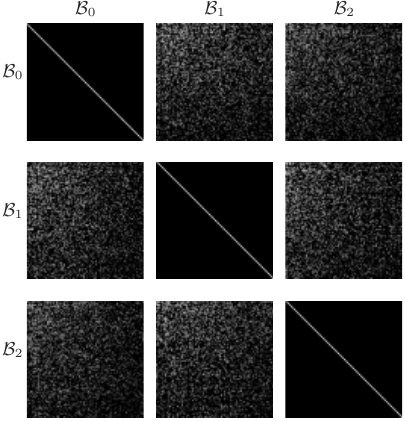}
    \end{minipage}
    \hfill
    \begin{minipage}{0.47\textwidth}
      \centering
      \textbf{Batch size $248$}\\[2ex]
      \includegraphics[scale=0.95]{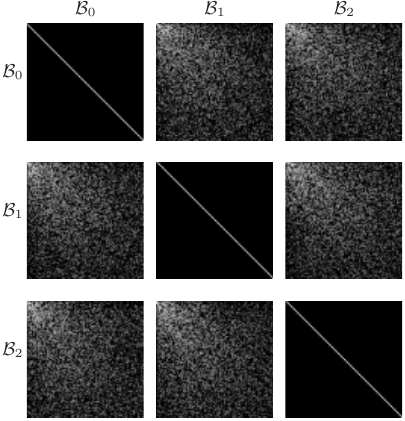}
    \end{minipage}

    \label{fig:additional_overlaps}
    \caption{\textbf{In practice, eigenspaces are misaligned.}
The experimental setting is similar to \cref{fig:overlaps}. The top plot shows
    overlaps between three curvature matrices that use the Hessian of the empirical
    risk, \ie $\hessian_\minibatch \gets \nabla^2 \Loss(\paramsmap; \minibatch) + \beta
    \mI$ at batch size $512$. The bottom plots use the GGN curvature proxy
    $\hessian_\minibatch \gets \GGN_\minibatch + \beta \mI$ at batch size $512$ \figleft
    and $2048$ \figright.
The weights $\mOmega_{i, j}$ are shown as a $100 \times 100$ greyscale image (color
    ranges from black for $\mOmega_{i, j} \leq 10^{-8}$ to white for $\mOmega_{i, j} =
    1$) for $m, m' \in \{0, 1, 2\}$.
Clearly, the eigenspaces for different mini-batches are not perfectly aligned as
    eigenvectors from $\minibatch_m$ overlap with several eigenvectors from
    $\minibatch_{m'}$. At the larger batch size, the eigenspaces are more aligned.}
\end{figure}

\textbf{Experimental procedure (overlaps).} Next, we compute the overlaps between the
eigenspaces of the curvature matrices on different mini-batches $\minibatch$ and
$\otherminibatch$. More specifically, we compute $\mOmega_{i, p} = (\eigvec_i^\top
\tilde{\eigvec}_p)^2$, where $i, p \in \{1, \dots, 100\}$, $\eigvec_i$ is an eigenvector
of $\hessian_{\minibatch}$ and $\tilde{\eigvec}_p$ is an eigenvector of
$\hessian_{\otherminibatch}$. The values $\mOmega_{i, p}$ are bounded between $0$
($\eigvec_i$ and $ \tilde{\eigvec}_p$ are orthogonal) and $1$ ($\eigvec_i$ and $
\tilde{\eigvec}_p$ are identical, up to their sign). We apply a log-transform to
$\mOmega_{i, p}$. In \cref{fig:overlaps,fig:additional_overlaps}, values below $-8$ (\ie
$\mOmega_{i, p} \leq 10^{-8}$) are shown in black, values equal to $0$ (\ie $\mOmega_{i,
p} = 10^0 = 1$) are shown in white.

\textbf{Results (overlaps).} The results are shown in \cref{fig:additional_overlaps}.
They show that the eigenspaces of the curvature matrices are not perfectly aligned:
Eigenvectors from $\minibatch_m$ typically overlap with several eigenvectors from
$\otherminibatch$. The eigenspaces are more aligned at the larger batch size $2048$ than
at batch size $512$. This seems reasonable since random vectors in high-dimensional
spaces tend to be orthogonal to each other ---less stochasticity (due to a larger batch
size) thus leads to better alignment.

\subsection{\cref{sec:exp_cg} (\nameref{sec:exp_cg}): \cref{fig:cg}} 
\label{sec:exp_details_cg}

Here, we describe the experimental details for \cref{fig:cg} from \cref{sec:exp_cg}. For
the derivation and the mathematical details of the standard single-batch \cg method and
the debiased approach, see \cref{sec:details_cg}. The experiment uses the fully trained
\allcnnc model from test problem \ref{tp:allcnnc_cifar100} (see
\cref{sec:exp_details_training}) and the GGN curvature proxy $\hessian_\minibatch \gets
\GGN_\minibatch + \beta \mI$ with $\beta = 0.0005$.

\textbf{Experimental procedure.} The experimental procedure consists of two steps: The
computation of the \cg trajectories and the evaluation of the four performance metrics.
\begin{enumerate}
  \item \textbf{Computation of the \cg trajectories.} For the single-batch \cg approach,
    we use one mini-batch of size $1024$, apply $\rankbound = 30$ \cg iterations and
    store the trajectory $\paramsmap = \params_0, \params_1, \ldots, \params_{30}$. For
    the debiased approach (details in \cref{sec:details_cg}), we use two mini-batches of
    size $512$ to construct the trajectory. As the debiased approach uses a total of
    $60$ GGN-vector products ($30$ for the search directions and $30$ for the update
    magnitudes) at half the computational cost (since the cost for the GGN-vector
    product scales linearly with the mini-batch size), the total cost of both approaches
    are comparable.
    
    For both approaches, the above procedure is repeated for $5$ different
    mini-batches/mini-batch pairs, such that $10$ trajectories are computed in total.

    We compute two additional trajectories: The standard \cg trajectory for the
    full-batch quadratic $\quadratic(\emptyarg; \traindata)$ and the trajectory for a
    debiased full-batch approach, where we use half the training data for the directions
    and the other half for the update magnitudes.

    \item \textbf{Evaluation of performance metrics.} For each of the $12$ trajectories,
    we evaluate four performance metrics: (i)~the regularized loss $\Lossreg$ (see
    \cref{eq:Loss_reg}) on the training set $\traindata$ and (ii)~on the test set
    $\testdata$ as well as the accuracy (\ie the relative number of correctly classified
    samples) on the (iii)~training and (iv)~test \dataset.
\end{enumerate}

\begin{figure}[p]
  \centering
  \includegraphics[width=0.99\textwidth]{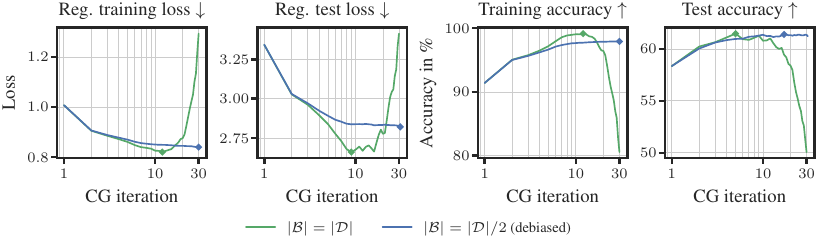}

  \caption{\textbf{Comparison of full-batch \cg approaches.} We use the setting from
  \cref{fig:cg}. The full-batch approach (shown as \colorline{colorexact}) applies
  standard \cg to $\quadratic(\emptyarg; \traindata)$ while the debiased approach
  (shown as \colorline{colorunbiased}) uses one half of the training data for the
  directions and the other half for the update magnitudes. The markers
  \colordiamond{colorexact} and \colordiamond{colorunbiased} are placed at peak
  performance. 
Surprisingly, the full-batch run diverges. The debiased \cg run is stable.}
  \label{fig:cg_full-batch}

  \vspace{5ex}

  \includegraphics[width=0.99\textwidth]{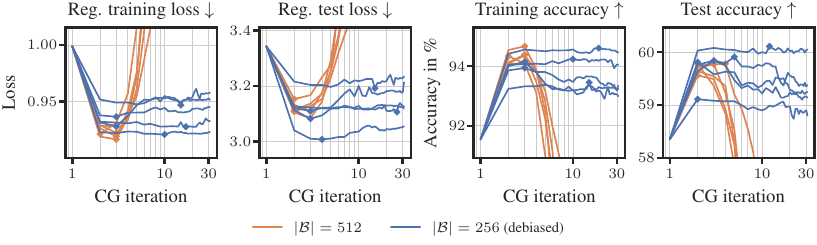}

  \vspace{2ex}

  \includegraphics[width=0.99\textwidth]{exp03/cifar100_allcnnc_wo_dropout_SGD_epoch_349_batch_0_batch_size_1024_curv_opt_ggn_damping_0.000e+00.pdf}

  \vspace{2ex}

  \includegraphics[width=0.99\textwidth]{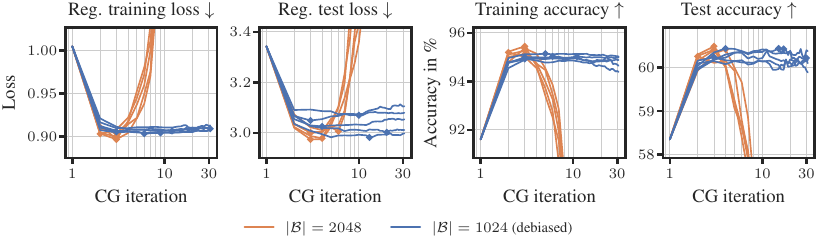}

  \caption{\textbf{Comparison of \cg approaches at different batch sizes.} We use the
  same experimental setting as in \cref{fig:cg} but consider different mini-batch sizes
  $\vert\minibatch\vert \in \{512, 1024, 2048\}$ (from \figtop to \figbottom). The
  debiased approach uses two mini-batches of size $256$, $512$, and $1024$,
  respectively.
Across all mini-batch sizes, the single-batch \cg runs diverge quickly while our
  debiased approach maintains stability.}
  \label{fig:cg_different_batch_sizes}
\end{figure}

\textbf{Results.} The mini-batch results are shown in \cref{fig:cg} and discussed in
\cref{sec:exp_cg}. 
The results for the two full-batch \cg variants are shown in \cref{fig:cg_full-batch}.
Surprisingly, the full-batch \cg approach diverges after roughly $10$ iterations. If we
consider the training data as a (large) sample from the data distribution, the
full-batch \cg approach can be seen as an instance of the mini-batch approach with a
very large mini-batch size. Thus, it might also exhibit the associated biases we
describe in \cref{sec:biases_theory}. Another explanation would be the approximation
error in \cref{eq:fullbatch_quadratic}: $\Lossreg(\emptyarg; \traindata) \; \approx \;
\quadratic(\emptyarg; \traindata)$. If the trajectory moves far away from $\paramsmap$,
it leaves the region where the approximation is valid (whereas the debiased full-batch
version might take smaller steps and thus remains stable). We leave it to future work to
investigate this further.

\textbf{Additional results.} We also provide the results for two other mini-batch sizes
$\vert\minibatch\vert \in \{512, 2048\}$ (the debiased approach uses two mini-batches of
size $256$ and $1024$, respectively). The results are shown in
\cref{fig:cg_different_batch_sizes}.

\textbf{Runtime and memory consumption.} Our current implementation uses the naive
approach that requires \textit{two} matrix-vector products with the GGN for computing
one debiased update magnitude (see \cref{sec:details_cg} for details). As this would
skew the runtime comparison, we refrain from providing concrete numbers here. However,
as we describe in detail in \cref{sec:details_cg}, it is possible to implement the
debiased approach in a more efficient way that only requires one additional
matrix-vector product for each \cg direction to compute the debiased update magnitude.
Using the debiased approach at half the mini-batch size, this brings the computational
costs down to roughly the same level as the standard approach (assuming that the runtime
of a GGN-vector product scales linearly with the mini-batch size). 

As explained in detail in \cref{sec:details_cg}, the memory overhead for the efficiently
implemented debiased \cg  approach is two additional vectors of size $\numparams$. For
the experimental setting used for \cref{fig:cg}, the number of  parameters is
$\numparams = \num{1387108}$, which corresponds to $5.55~\text{MB}$ in single precision.
The computational overhead incurred by the debiased approach is thus about
$11.1~\text{MB}$.

\subsection{\cref{sec:exp_LA} (\nameref{sec:exp_LA}): \cref{fig:la}}
\label{sec:exp_details_la}

\Cref{fig:la} shows a comparison of the vanilla model, the mini-batch \kfac LA, the
debiased version, and the full-batch LA. Here, we describe the experiment in more detail
and present additional results, in particular on OOD data. For the derivation and the
mathematical details of these approaches, see \cref{sec:details_laplace_approximation}.
The experiment uses the fully trained \allcnnc model on \cifarten data from test problem
\ref{tp:allcnnc_cifar10} (see \cref{sec:exp_details_training}).

\textbf{Experimental procedure.} The experimental procedure consists of three steps: The
computation of the \kfac curvature approximations, the evaluation of the corresponding
predictive class probabilities, and the computation of the performance metrics.
\begin{enumerate}
  \item \textbf{\kfac curvature approximations.} We consider a log-equidistant grid of
  $13$ prior precisions $\beta$ between $10^{-4}$ and $10^0$ and add $\beta = 10$.
  
  For each of those $14$ values, we compute the \kfac curvature approximation via
  \backpack \citep{dangel2020backpack} (see \cref{sec:exp_details_curvature_matrices})
  using (i) a single batch of size $64$, $256$, and $1024$, (ii) the two-batch
  debiased LA version (details in \cref{sec:details_debiased_kfac}) at batch sizes
  $32$, $128$, and $512$, and (iii) the full-batch LA. For the latter, we accumulate
  mini-batch \kfac approximations over the entire training set. As an additional
  baseline, we consider (iv) the vanilla model (without LA), which is independent of the
  prior precision $\beta$. For approaches (i) and (ii), we repeat the experiment for $5$
  different mini-batches/mini-batch pairs. 
  
  \item \textbf{Evaluation of predictive class probabilities.} The first step in the
  evaluation of the performance metrics is the computation of the predictive
  uncertainty. For a given test \dataset of size $\numtestdata$, these can be
  represented as a matrix $\mP \in \R^{\numtestdata \times \numclasses}$, where $\mP_{n,
  c}$ is the probability that the $n$-th sample belongs to class $c$, \ie rows of $\mP$
  sum to $1$. In all experiments, we draw $S = 40$ MC
  samples $\{\params^{(s)}\}_{s=1}^S$ from the weight posterior following the procedure in
  \cref{sec:details_laplace_approximation_sampling}. The predictive class probabilities
  are then obtained via \cref{eq:laplace_monte_carlo}. In the case of the vanilla model
  without LA, the sum in \cref{eq:laplace_monte_carlo} collapses to a single term
  corresponding to the MAP model.
  
  The evaluation procedure above can be applied to arbitrary test \datasets. We consider
  the training data, test data and \cifartenc (\ie OOD \datasets at severity $1$ to
  $5$).

  \item \textbf{Performance metrics.} We consider the following performance metrics:
  \begin{itemize}
    \item \textbf{Accuracy.} The predictive classes are obtained from $\mP$ by
    extracting the class with the highest probability. The accuracy is the relative
    number of correctly classified samples.

    \item \textbf{Negative log-likelihood (NLL).} For each datum in the test set, we
    compute the negative log-probability of the true class. We then average over all
    samples in the test \dataset. This coincides with the empirical risk from
    \cref{eq:Loss_reg}, evaluated on the test \dataset.

    \item \textbf{Expected calibration error (ECE).} The ECE is a measure of the
    calibration of the model's predictive probabilities. It groups the classification
    confidences (\ie the maximum entry in each of $\mP$'s rows) into bins, and within
    these bins, compares the average confidence with the actual accuracy. We use
    \texttt{MulticlassCalibrationError} from \texttt{torchmetrics}
    \citep{Detlefsen2022TorchMetrics}.

    \item \textbf{AUROC.} For the OOD \datasets, we provide the Area Under the Receiver
    Operating Characteristic curve (AUROC). Our goal is to distinguish in-distribution
    (ID) and out-of-distribution (OOD) samples using the model's uncertainty. We use the
    entropy of the predictive distribution $p(\vy_\testsymbol \mid \vx_\testsymbol,
    \trainset)$ as our uncertainty estimate $u(\vx_\testsymbol) \in \R$ for input
    $\vx_\testsymbol$. The ground-truth binary labels are the ID/OOD indicators of the
    test inputs. Several thresholds $\xi$ could be established to turn the scalar
    uncertainty estimates into binary predictions by using
    $\mathbf{1}_{\{u(\vx_\testsymbol) > \xi\}}$. Instead of choosing a particular
    threshold and evaluating its accuracy, the AUROC metric directly evaluates
    $u(\emptyarg)$ by measuring the area under the plot of the true positive rate (TPR)
    against the false positive rate (FPR) for all possible thresholds.
  \end{itemize}
\end{enumerate}

\textbf{Additional experimental details.} 
\begin{itemize}
  \item \textbf{Uncertainty over weights but not biases.} As the prior acts only on the
  weights of the network but not its biases (see \cref{sec:exp_details_training}), we
  only consider the uncertainty over the weights in the LA. This slightly reduces the
  size of the covariance matrix as it excludes the bias parameters.

  \item \textbf{Single vs. double precision.} Although the \kfac factors are positive
  semi-definite by construction, their numerical eigenvalues can be negative due to
  numerical inaccuracies. To balance precision and computational cost, we use double
  precision for all computations until drawing the MC samples and single precision afterwards.
\end{itemize}
 
\begin{figure}[p]
  \centering
  \textbf{Performance on the training set}\\[1ex]
  \includegraphics[width=0.99\textwidth]{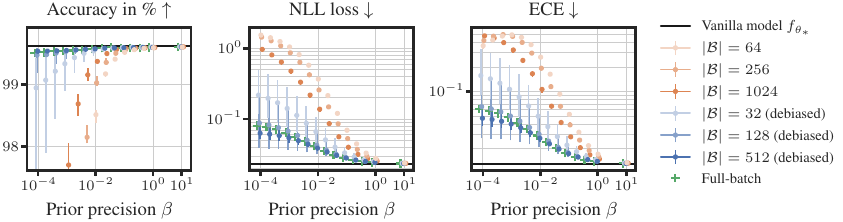}

  \vspace{3ex}

  \textbf{Performance on \cifartenc (severity level 1)}\\[1ex]
  \includegraphics[width=0.99\textwidth]{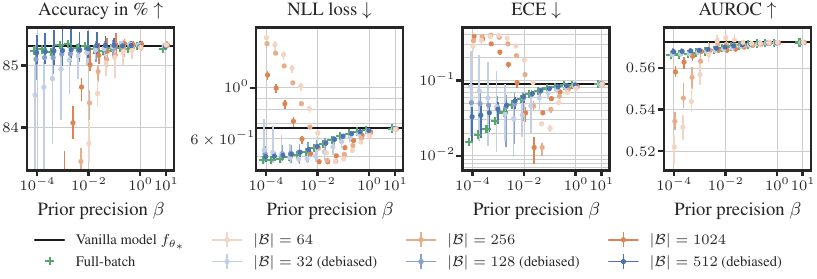}

  \vspace{3ex}

  \textbf{Performance on \cifartenc (severity level 2)}\\[1ex]
  \includegraphics[width=0.99\textwidth]{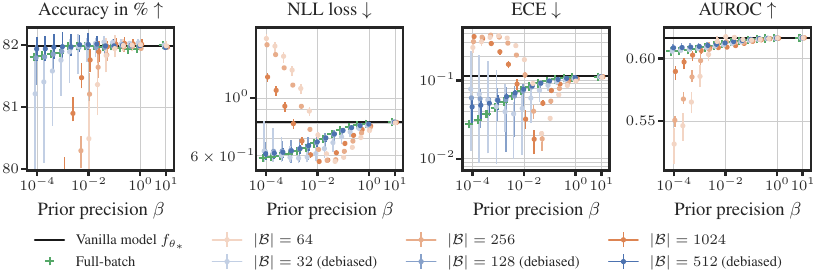}

  \caption{\textbf{Debiased LA mimics the full-batch LA very well.} 
  The experimental setting is the same as in \cref{fig:la}, but we report results on
  additional \datasets (training set and OOD \datasets at severity level $1$ and $2$).
  For the OOD \datasets, we report the AUROC metric in addition to the accuracy, NLL and
  ECE.
In contrast to the single-batch approach, the debiased version mimics the behavior of
  the full-batch approach very well over the entire range of prior precisions and across
  \datasets. }
  \label{fig:additional_la_results_1}
\end{figure}

\begin{figure}[p]
  \centering

  \textbf{Performance on \cifartenc (severity level 3)}\\[1ex]
  \includegraphics[width=0.99\textwidth]{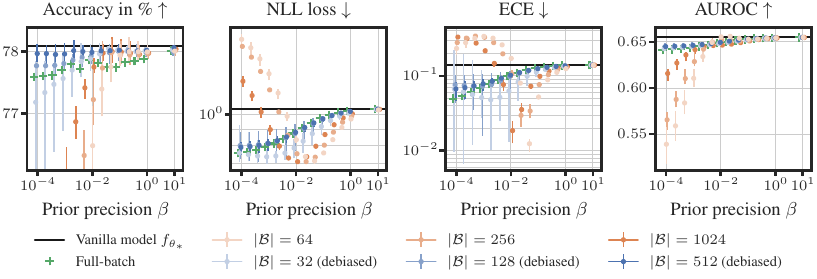}

  \vspace{3ex}

  \textbf{Performance on \cifartenc (severity level 4)}\\[1ex]
  \includegraphics[width=0.99\textwidth]{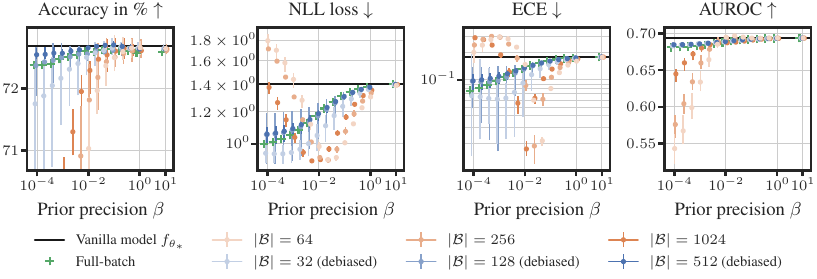}

  \vspace{3ex}
  
  \textbf{Performance on \cifartenc (severity level 5)}\\[1ex]
  \includegraphics[width=0.99\textwidth]{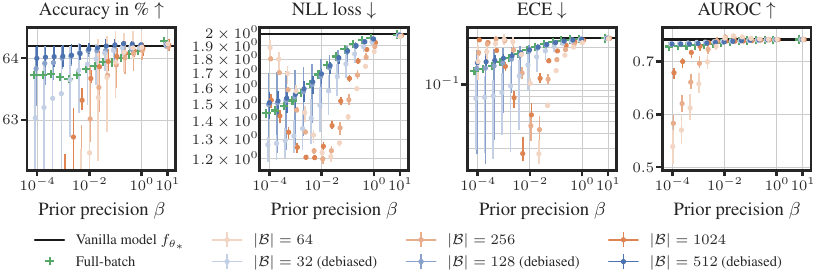}

  \caption{\textbf{Debiased LA mimics the full-batch LA very well.} 
  The experimental setting is the same as in \cref{fig:la}, but we report results on
  additional \datasets (OOD \datasets at severity level $3$, $4$, and $5$). We report
  the AUROC metric in addition to the accuracy, NLL, and ECE.
In contrast to the single-batch approach, the debiased version mimics the behavior of
  the full-batch approach very well over the entire range of prior precisions and across
  \datasets. }
  \label{fig:additional_la_results_2}
\end{figure}

\textbf{Results.} The results are presented in
\cref{fig:la,fig:additional_la_results_1,fig:additional_la_results_2}. For the
single-batch and debiased approach (where we use $5$ mini-batches/mini-batch pairs
each), we report the average performance as a dot and the min/max as a vertical line.
The performance of the vanilla model is shown as a horizontal line as its performance
does not depend on $\beta$, as explained above.

Across all \datasets and performance metrics, the debiased LA mimics the behavior of the
full-batch approach much better than the single-batch LA (although both approaches use a
comparable amount of computational resources). In particular, for small prior
precisions, where the covariance matrix relies almost exclusively on the \kfac curvature
information, the debiased LA maintains a good performance in contrast to the
single-batch approach. 

\begin{figure}[thb]
  \centering
  \textbf{Performance on the test set}\\[1ex]
  \includegraphics[width=0.99\textwidth]{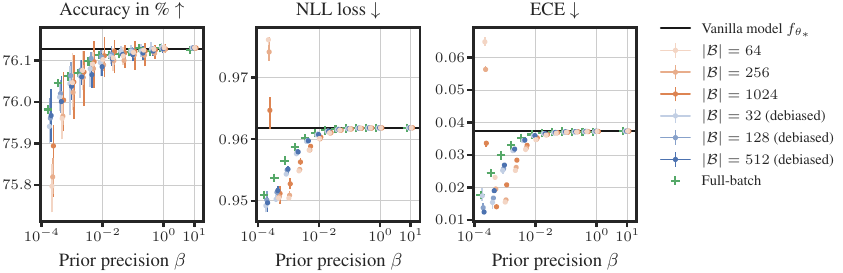}

  \caption{\textbf{Our approach scales to \resnetfifty on \imagenet.} The experimental
  setting is the same as in \cref{fig:la}, but we use test problem
  \ref{tp:resnet50_imagenet} (see \cref{sec:exp_details_training}) (\resnetfifty on
  \imagenet). The results are consistent with the findings on \cifarhun: The debiased
  approach behaves similarly to the full-batch LA and maintains stability even for small
  prior precisions. }
  \label{fig:additional_la_imagenet}
\end{figure}

\begin{figure}[thb]
  \centering
  \textbf{Performance on the test set}\\[1ex]
  \includegraphics[width=0.99\textwidth]{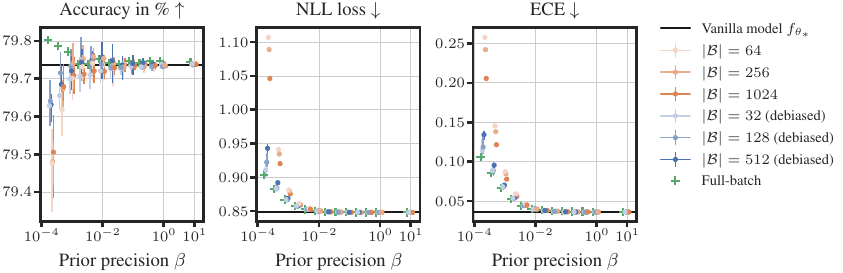}

  \caption{\textbf{Curvature biases are also present for the \vitlittle architecture on \imagenet.}
  The experimental setting is the same as in \cref{fig:la} and
  \cref{fig:additional_la_imagenet}, but we use test problem
  \ref{tp:vit_little_imagenet} (see \cref{sec:exp_details_training}) (\vitlittle on
  \imagenet). Again, the debiased approach behaves similarly to the full-batch LA and
  maintains stability even for small prior precisions (where the covariance is almost
  exclusively based on the \kfac curvature information).} 
  \label{fig:additional_la_imagenet_vit}
\end{figure}

\textbf{Additional results.} 
\begin{itemize}
  \item \textbf{\resnetfifty on \imagenet.} To showcase the
  scalability of our debiased LA approach, we repeat the experiment for a \resnetfifty
  model on the \imagenet \dataset (test problem \ref{tp:resnet50_imagenet} in
  \cref{sec:exp_details_training}). The results are shown in
  \cref{fig:additional_la_imagenet}. Again, the debiased approach behaves similarly to
  the full-batch LA and maintains stability over the entire spectrum of prior precisions.
  
  \item \textbf{\vitlittle on \imagenet.} Similar to the previous experiment, we also
  apply our LA debiasing approach to a \vitlittle model on the \imagenet \dataset (test
  problem \ref{tp:vit_little_imagenet} in \cref{sec:exp_details_training}) to ensure
  that it is beneficial for other architectures as well. The results are shown in
  \cref{fig:additional_la_imagenet_vit}. The debiased approach mimics the calibration
  behavior of the full-batch LA remarkably well and remains stable for lower prior
  precision values as well where the single-batch approach behaves erratically.
\end{itemize}

\textbf{Runtime analysis (computational overhead of debiasing).} We claim in
\cref{sec:exp_LA} that, by using the debiased approach at half the mini-batch size that
is used by the single-batch approach, we obtain a fair comparison (more details in
\cref{sec:computational_cost}). Here, we substantiate this claim by providing a detailed
runtime comparison of the biased, debiased, and full-batch LA approaches on \cifarten.

\begin{figure}[thb!]
  \centering
  \includegraphics[width=0.99\textwidth]{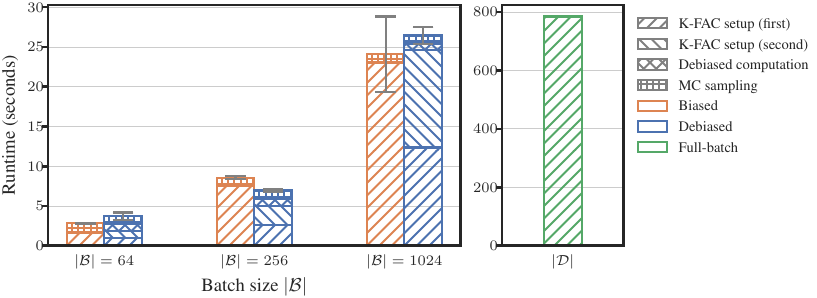}
  \caption{\textbf{The computational overhead of debiasing is negligible for Laplace approximations.}
    We follow the experimental setup of \cref{sec:exp_LA} and provide a detailed runtime
    comparison of the biased (\colorrectangle{colorbiased}), debiased
    (\colorrectangle{colorunbiased}), and full-batch (\colorrectangle{colorexact}) \kfac
    Laplace approximation schemes on \cifarten.
The total runtime includes the construction of \kfac{}'s Kronecker factors (on one
    or two mini-batches for the biased and debiased approach, respectively), the
    debiasing computations (see \cref{sec:details_debiased_kfac}), and the Monte Carlo
    sampling of $S = 40$ weights. 
All reported runtimes are averages over $5$ runs. The error bars for the total
    runtime cover one standard deviation. The results show that the debiased approach
    introduces a negligible overhead. Compared to the full-batch variant, both
    mini-batch approaches are orders of magnitude faster. 
Note that the evaluation (i.e., the computation of the predictive uncertainty) is
    not included in the comparison, as the time requirements are the same for all
    approaches; however, this step requires the most computational resources.}
  \label{fig:computational_overhead_LA}
\end{figure}

\textbf{Results (runtimes).} We report the runtimes of the biased
(\colorrectangle{colorbiased}), debiased (\colorrectangle{colorunbiased}), and
full-batch (\colorrectangle{colorexact}) \kfac Laplace approximation schemes in
\cref{fig:computational_overhead_LA}. We find that (i) both biased and debiased
approaches are orders of magnitude cheaper than the full-batch LA, and (ii) the debiased
approach introduces only negligible overhead and can, sometimes, even improve upon the
runtime of the biased approach.

\textbf{Memory consumption.} The \allcnnc model we use for the LA experiment in
\cref{sec:exp_LA} has $\numparams = \num{1368480}$ trainable parameters (this amounts to
$10.95~\text{MB}$ of memory in double precision). The \kfac approximation requires
storing \num{11483965} ($91.87~\text{MB}$) numbers to represent all Kronecker factors.
The memory of a \kfac approximation is thus roughly equivalent to that of $8.4$ models.
As the debiased approach is based on \textit{two} such approximations in the case of a
naive implementation, the overhead of debiasing is another $91.87~\text{MB}$ of memory.
However, this can be significantly improved by building up the \kfac approximations
block by block: Having two corresponding blocks available (based on two different
mini-batches) already suffices to compute the respective debiased block (see
\cref{sec:details_debiased_kfac}). This way, we only have two blocks (represented by
their Kronecker factors) in memory at the same time (instead of two entire \kfac
approximations) which greatly reduces the memory overhead of debiasing, down to the
largest block. For the \allcnnc model, this is two Kronecker factors of sizes $192
\times 192$ and $1728 \times 1728$ amounting to \num{3022848} numbers in total, \ie the
equivalent of $2.20$ models or $24.18~\text{MB}$ in double precision. The statement that
the debiased approach roughly doubles the memory consumption is thus a worst-case
scenario.

\subsection{Additional experiment: Biases for a \textsc{WideResNet 40-4} model
architecture}

\begin{figure}[thb!]
  \centering
  Eigenvectors computed on mini-batch \\[0.25ex]
  \includegraphics[width=0.99\textwidth]{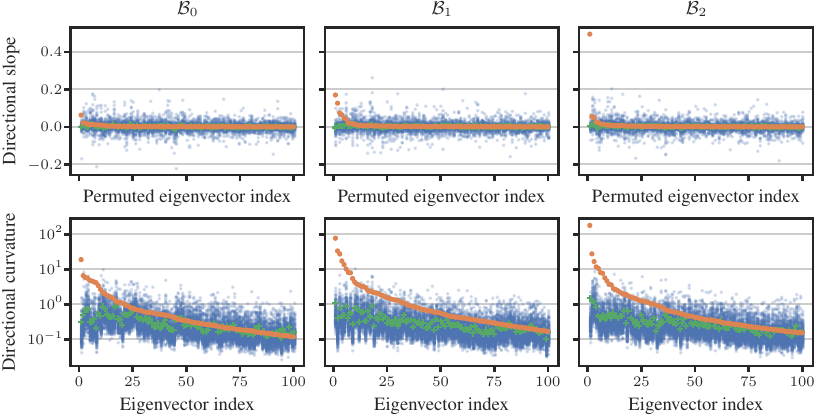}
  \caption{\textbf{Directional slopes and curvatures are biased.}
The experimental setting is similar to \cref{fig:bias}, but we use test problem
    \ref{tp:wideresnet_cifar100} (\wideresnet on \cifarhun) with the GGN curvature proxy
    $\hessian_\minibatch \gets \GGN_\minibatch + \beta \mI$ at batch size $256$.
For the top panel, we switch the order and sign of the eigenvectors such that the
    orange dots are all above zero and in descending order.
There is a strong, systematic bias, particularly in the curvature: Computing the
    eigenvectors and directional curvatures on the same data results in over-estimation
    that is very pronounced along the first few eigenvectors but then decays more
    quickly than in test problem \ref{tp:allcnnc_cifar100}.}
  \label{fig:biases_resnet}
\end{figure}

\textbf{Experimental setting and results.} Here, we extend our analysis from
\cref{sec:empirical_study_dd} (details in \cref{sec:exp_details_biases}) to test problem
\ref{tp:wideresnet_cifar100} (see \cref{sec:exp_details_training}): A \wideresnet model
on \cifarhun. We use the GGN curvature proxy $\hessian_\minibatch \gets \GGN_\minibatch
+ \beta \mI$ at batch size $256$ and compute the directional slopes and curvatures along
the top $100$ eigenvectors of $\hessian_\minibatch$. The results are shown in
\cref{fig:biases_resnet}. Along the first few top-curvature directions, the curvature
biases are even larger for this test problem than for the \allcnnc model on \cifarhun,
but then they decay more quickly. 

\subsection{Additional experiment: \kfac and the dependence of the biases on mini-batch size}
\label{sec:exp_kfac}

\textbf{Experimental setting.} Here, we extend our analysis of
test problem
\ref{tp:allcnnc_cifar100} (see \cref{sec:exp_details_training}) from
\cref{sec:empirical_study_dd} to \kfac and investigate the impact of the mini-batch size
on the curvature biases. 

\begin{figure}[thb!]
  \centering
  Eigenvectors computed on mini-batch $\minibatch_0$ with batch size\\[0.25ex]
  \includegraphics[width=0.99\textwidth]{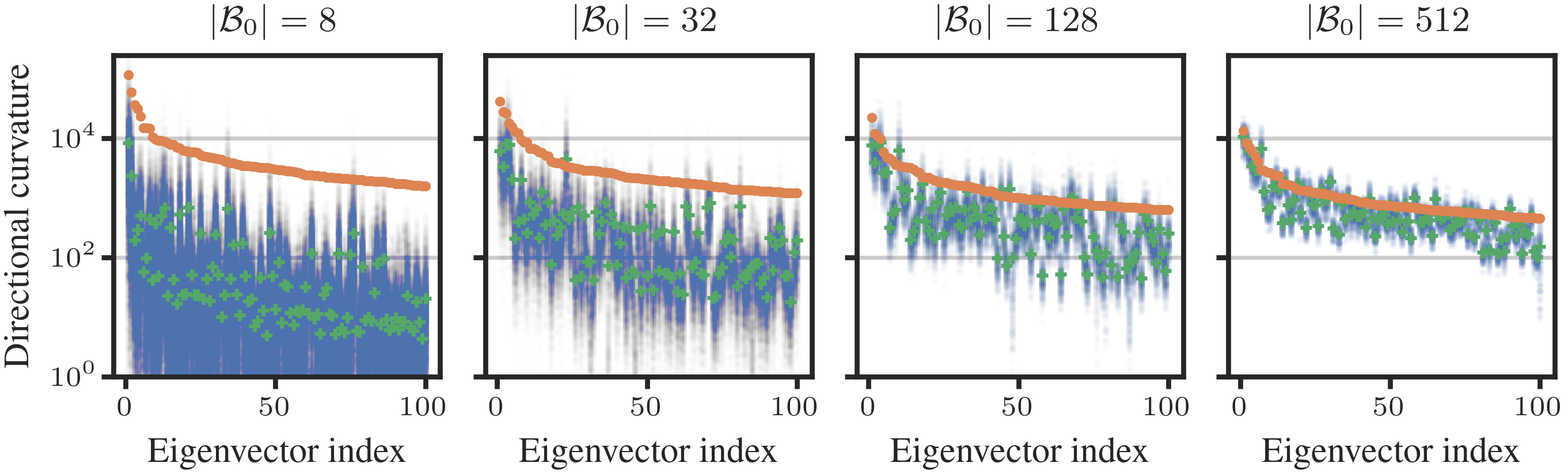}
  \caption{\textbf{Directional curvatures with \kfac.} 
  We use the \cifarhun dataset with the fully trained \allcnnc model. For each
  mini-batch size $\in \{8, 32, 128, 512\}$, we draw one mini-batch and compute the top
  $100$ eigenvectors $\eigvec_1, \ldots, \eigvec_{100}$ of the \kfac matrix
  $\KFAC_{\minibatch_0}$. We show the directional curvatures $\eigvec_p^\top
  \KFAC_{\minibatch_m} \eigvec_p$, $p \in \{1, \ldots, 100\}$ on the same mini-batch ($m =
  0$) as \colordot{colorbiased}, all other mini-batches in the training set ($m \in \{1,
  2, \ldots\}$) as \colordot{colorunbiased} and their average as
  \colorcross{colorexact}.
Similar to \cref{fig:bias}, we observe a systematic bias in the curvature. The bias
  decreases with increasing mini-batch size. }
  \label{fig:kfac}
\end{figure}

\textbf{Results \& discussion.}
\cref{fig:kfac} shows the directional curvatures of the \kfac approximation for four
different mini-batch sizes. When we compute the eigenvectors and directional curvatures
on the same mini-batch, we observe a systematic curvature bias that decreases with
increasing mini-batch size. 
There are two phenomena at play: With increasing mini-batch size, (i) the green crosses
move upwards and (ii) the orange dots move downwards. Our intuition for this is as
follows: With increasing mini-batch size, the eigenvectors become more meaningful such
that, on other data, they also exhibit large curvature---this explains (i). Similarly,
with increasing mini-batch size, it gets harder to find directions of \textit{extreme}
curvature as these directions have to exhibit large curvature on \textit{all} data
points within the mini-batch---this explains (ii).

\subsection{Additional experiment: Development of the biases over the course of training}
\label{sec:exp_biases_over_time}

\begin{figure}[thb!]
  \centering
  \includegraphics[width=0.99\textwidth]{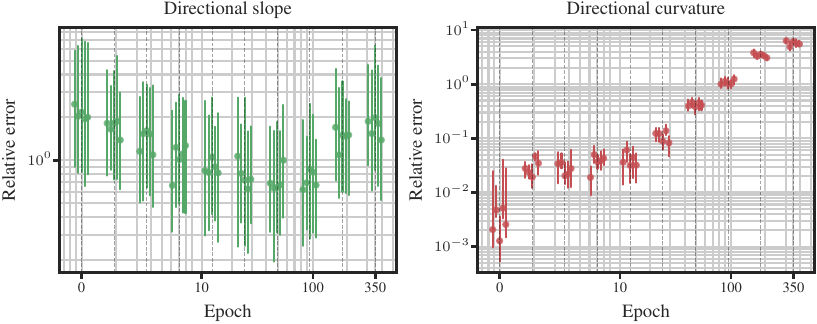}
  \caption{\textbf{Curvature biases increase over the course of training.} We evaluate
    the relative slope and curvature biases at different checkpoints during training for
    the \allcnnc model on \cifarhun. $5$ mini-batches are drawn per checkpoint. For each
    mini-batch, the relative biases are represented as a dot \colordot{black} (at the
    mean relative bias) and a vertical line \colorline{black} ranging from the $25\%$ to
    the $75\%$ percentile. While the biases in the slope remain relatively stable over
    the course of training, the biases in the curvature increase (over more than $3$
    orders of magnitude).}
  \label{fig:biases_over_time}
\end{figure}

\textbf{Experimental setting and results.} Here, we investigate how the biases in the
slope and curvature develop over the course of training for test problem
\ref{tp:allcnnc_cifar100} (see \cref{sec:exp_details_training}). We use the same
procedure as in \cref{sec:exp_details_biases} but evaluate the biases at $10$ different
checkpoints during training (spread log-equidistantly between the first and last epoch).
At each checkpoint, we draw $5$ mini-batches, compute the top $100$ eigenvectors of the
corresponding GGN-based quadratics, and finally evaluate the \textit{relative} errors in
the directional slopes and curvatures (where the ground truth is the full-batch
quadratic's slope/curvature). For each mini-batch, this distribution of $100$ relative
curvature and slope biases is represented as a dot (at the mean relative bias) and a
vertical line ranging from the $25\%$ to the $75\%$ percentile in
\cref{fig:biases_over_time}. 
While the biases in the slope remain relatively stable over the course of
training, the
biases in the curvature increase over more than $3$ orders of magnitude up to a relative
error of order $10$ in epoch $349$ (which is consistent with the results in
\cref{fig:bias}). This suggests that the eigenspaces of different curvature matrices
become more and more misaligned as training progresses steadily increasing the need for
effective debiasing strategies.

\subsection{Additional experiment: The curvature bias increases with $\numparams$}
\label{sec:exp_numparams}

In high-dimensional spaces, it becomes increasingly unlikely that random vectors are
aligned.
While the eigenspaces of the curvature matrices are not completely random, they
are subject to noise. It is thus conceivable that their overlap decreases as the number
of parameters $\numparams$ increases. If this hypothesis is true, the curvature biases
should become more pronounced in large models.

\begin{figure}[thb!]
  \begin{minipage}[t]{0.45\textwidth}
    \centering
    \vspace{0pt}  \includegraphics[width=0.99\textwidth]{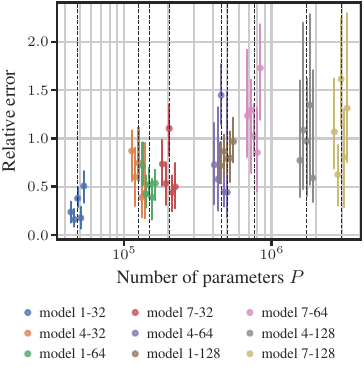}
  \end{minipage}
  \hfill
  \begin{minipage}[t]{0.52\textwidth}
    \vspace{0pt}  \caption{\textbf{Increasing curvature biases with $\numparams$.} We train $9$
      convolutional neural networks with different widths and depths for $100$ epochs on
      the \cifarten \dataset using \adam with standard hyperparameters. For the
      fully trained model, we evaluate the relative error 
      $
        \vert 
        \dcurv{\eigvec_p} \quadratic(\paramsmap; \minibatch)
        - 
        \dcurv{\eigvec_p} \quadratic(\paramsmap; \traindata)
        \vert
        \cdot
        \vert
        \dcurv{\eigvec_p} \quadratic(\paramsmap; \traindata)
        \vert^{-1}
      $
      for $\GGN_{\minibatch}$'s top $100$ eigenvectors at batch size
      $\vert\minibatch\vert = 128$. $5$ different mini-batches are used per model
      resulting in a total of $45$ error distributions (each consisting of $100$
      numbers). These distributions are represented by their median (as a dot
      \colordot{colorunbiased}) and the $25\%$ and $75\%$ percentiles (as a line segment
      \colorline{colorunbiased}) $\numparams$. The $5$ distributions for each model are
      slightly spread along the $x$-axis for better visibility. The experiment confirms
      our hypothesis: The relative errors tend to increase with the number of parameters
      $\numparams$.}
      \label{fig:bias_numparams}
  \end{minipage}
\end{figure}

\textbf{Experimental procedure \& results.} To test this hypothesis, we use test problem
\ref{tp:convnet_cifar10} that implements a simple convolutional neural network with
variable width and depth (for details, see \cref{sec:exp_details_training}). For each
fully trained model, we evaluate the \textit{relative} error between $\dcurv{\eigvec_p}
\quadratic(\paramsmap; \minibatch)$ and $\dcurv{\eigvec_p} \quadratic(\paramsmap;
\traindata)$ for the $\GGN_{\minibatch}$'s top $100$ eigenvectors at batch size $128$.
This procedure is repeated for $5$ different mini-batches for each of the $9$ models,
resulting in a total of $45$ error distributions (see \cref{fig:bias_numparams}). The
results confirm our hypothesis: The relative errors tend to increase with the number of
parameters $\numparams$. In the massively overparameterized regime, the biases might
thus become even more relevant and effective debiasing strategies are needed.

\stopcontents[sections]   

\end{document}